\documentclass[runningheads]{llncs}

\usepackage[mobile]{eccv}

\usepackage{eccvabbrv}

\usepackage{graphicx}
\usepackage{booktabs}
\usepackage{algorithm}
\usepackage{algpseudocode}
\usepackage{amsmath, amssymb}
\usepackage{color}
\usepackage{makecell}

\usepackage{tabularx}
\usepackage{array}
\usepackage{enumitem}

\usepackage[accsupp]{axessibility}  

\usepackage{hyperref}

\usepackage{orcidlink}

\begin{document}

\title{Improving Image-to-Image Translation via \\a Rectified Flow Reformulation} 


\author{Satoshi Iizuka\thanks{These authors contributed equally to this work.} \and
Shun Okamoto\protect\footnotemark[1] \and
Kazuhiro Fukui}

\authorrunning{Iizuka et al.}

\institute{University of Tsukuba}

\maketitle

\begin{abstract}
In this work, we propose \textit{Image-to-Image Rectified Flow Reformulation (I2I-RFR)}, a practical plug-in reformulation that recasts standard I2I regression networks as continuous-time transport models.
While pixel-wise I2I regression is simple, stable, and easy to adapt across tasks, it often over-smooths ill-posed and multimodal targets, whereas generative alternatives often require additional components, task-specific tuning, and more complex training and inference pipelines. Our method augments the backbone input by channel-wise concatenation with a noise-corrupted version of the ground-truth target and optimizes a simple $t$-reweighted pixel loss. 
This objective admits a rectified-flow interpretation via an induced velocity field, enabling ODE-based progressive refinement at inference time while largely preserving the standard supervised
training pipeline. In most cases, adopting I2I-RFR requires only expanding the input channels, and inference can be performed with a few explicit solver steps (e.g., 3 steps) without distillation. Extensive experiments across multiple image-to-image translation and video restoration tasks show that I2I-RFR generally improves performance across a wide range of tasks and backbones, with particularly clear gains in perceptual quality and detail preservation. 
Overall, I2I-RFR provides a lightweight way to incorporate continuous-time refinement into conventional I2I models without requiring a heavy generative pipeline.
\end{abstract}
\vspace{-5mm}

\newcommand{\repfigzoom}[2]{%
  \includegraphics[
    width=0.16\linewidth,
    trim=#1,clip
  ]{figs/results/Comp_L1_RFR/#2}%
}
\newcommand{\repfig}[1]{\includegraphics[width=0.16\linewidth]{figs/results/Comp_L1_RFR/#1}}
\begin{figure*}[t]
   \centering
   \setlength{\tabcolsep}{1pt}
   \begin{tabular}{ccc ccc}
   \multicolumn{3}{c}{Low-Light Blur Enhancement} &
   \multicolumn{3}{c}{Underwater Enhancement} \\
   \repfigzoom{85mm 0mm 85mm 0mm}{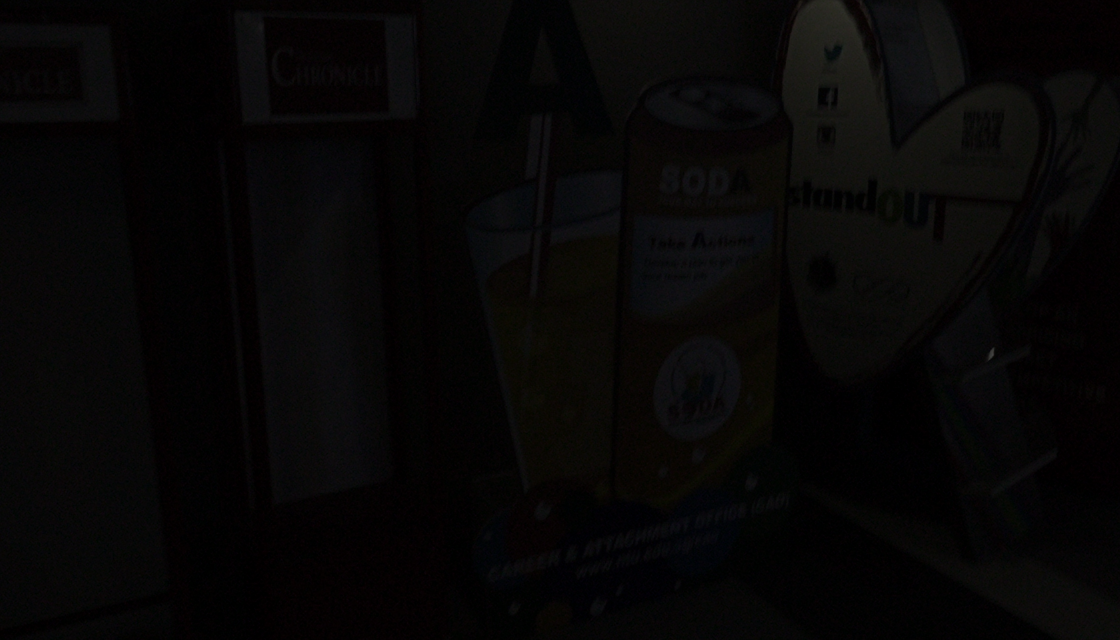} &
   \repfigzoom{85mm 0mm 85mm 0mm}{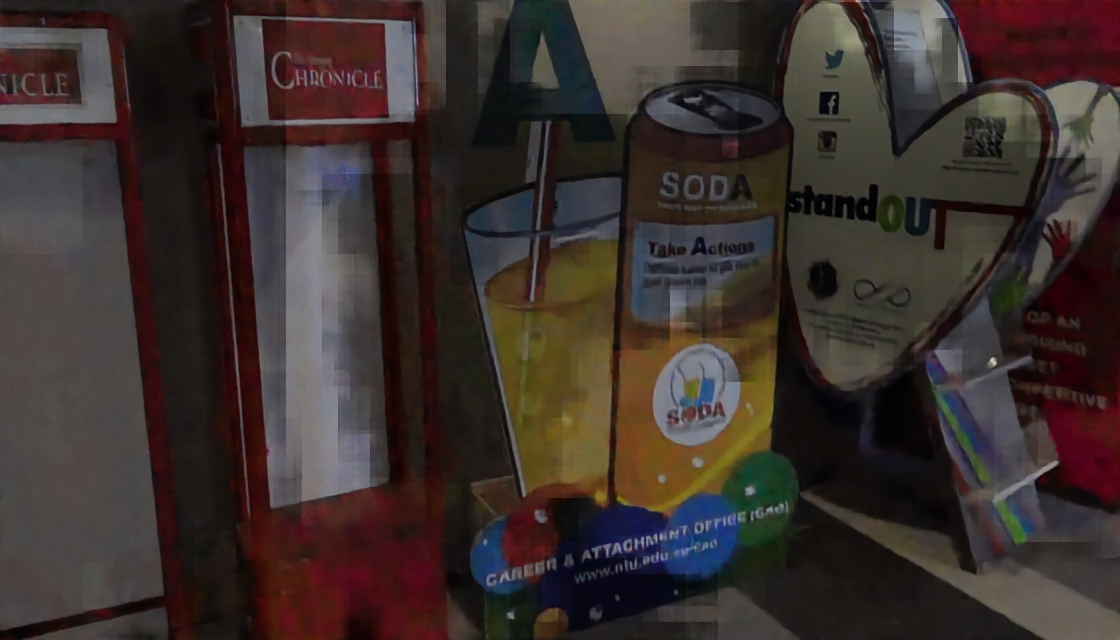} &
   \repfigzoom{85mm 0mm 85mm 0mm}{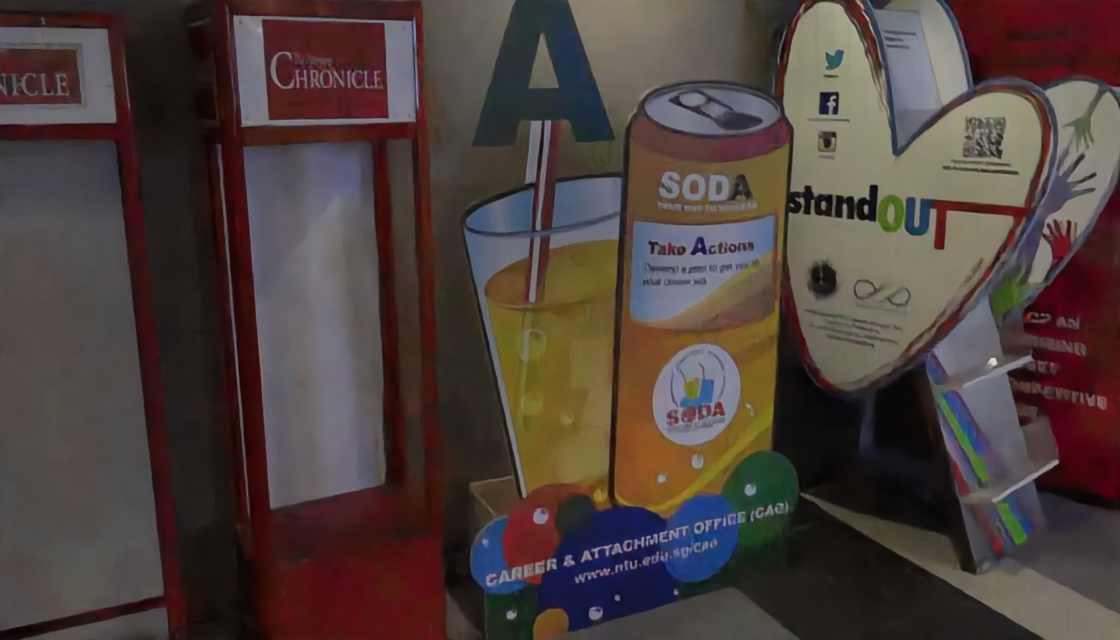} &
   \repfig{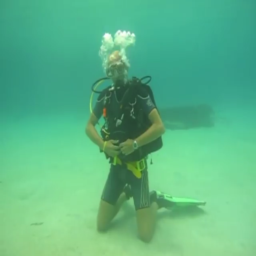} &
   \repfig{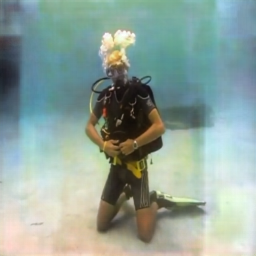} &
   \repfig{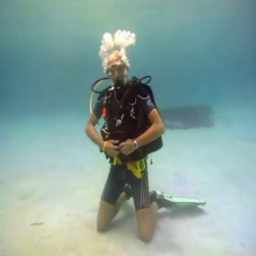} \\
   Input & $\ell_1$ Regression & $\ell_1$ RFR & Input & $\ell_1$ Regression & $\ell_1$ RFR \\
   \end{tabular}
   \setlength{\tabcolsep}{6pt}
   \vspace{-4mm}
    \caption{Comparison between conventional $\ell_1$ regression and our $\ell_1$ Rectified Flow Reformulation (RFR). Both use the same image-to-image network; the only difference is the number of input channels. RFR reduces regression artifacts and improves perceptual fidelity.}
   \label{fig:comp_reg_rfr}
   \vspace{-6mm}
\end{figure*}

\section{Introduction}
Image-to-Image (I2I) translation is a fundamental problem in computer vision, spanning image super-resolution, restoration, low-light enhancement, and many others.
A dominant I2I paradigm is formulated as regression: given an input image $x$, a neural network predicts the target image $y$ and is trained with pixel-wise losses such as $\ell_1$ or mean squared error (MSE).
This formulation is stable and practical, and a rich set of architectural advances, including U-Nets~\cite{ronneberger2015unet}, residual designs~\cite{he2016resnet}, attention modules~\cite{vaswani2017attention,wang2018nonlocal}, and multi-scale processing~\cite{lin2017fpn}, has been accumulated and successfully deployed in real applications.
However, many I2I problems are intrinsically ill-posed: severe degradations, missing regions, or strong information loss often make the conditional distribution $p(y \mid x)$ multimodal.
In such cases, pixel-wise regression tends to collapse toward a single point estimate (e.g., the conditional mean or median), often yielding perceptually suboptimal outputs such as detail loss and task-dependent artifacts, as shown in Fig.~\ref{fig:comp_reg_rfr}.
A widely used remedy is adversarial training, introduced with Generative Adversarial Networks (GANs)~\cite{GoodfellowPMXWOCB14}, where a discriminator encourages outputs that look realistic.
Although adversarial training is effective in many scenarios, adversarial objectives introduce additional discriminator networks and hyperparameters that are sensitive to the balance between the generator and discriminator, and may lead to training instability or artifacts, making it expensive to deploy and tune for each task.

Diffusion models and related probabilistic generative models have recently achieved impressive image quality~\cite{ho2020ddpm} and have also been extended to conditional generation for I2I tasks, including super-resolution~\cite{saharia2021sr3}, general image-to-image translation~\cite{saharia2022palette}, and inpainting~\cite{lugmayr2022repaint}.
Although these models are well-suited to multimodal targets, they also bring diffusion-based training and inference pipelines that differ substantially from standard I2I regression.
In particular, diffusion pipelines typically rely on specific parameterizations and conditioning mechanisms, noise schedules, and iterative sampling with tens to hundreds of denoising steps~\cite{ho2020ddpm,song2020ddim}.
When applied to task-specific I2I settings, this often increases implementation complexity and training cost, and can limit the extent to which established I2I architectures and loss designs can be reused.
Latent diffusion~\cite{rombach2022ldm} further requires an encoder--decoder, e.g., a variational autoencoder~\cite{kingma2014vae}, to define and learn a latent space, and relying on large pretrained models can restrict architectural flexibility and raise domain-gap concerns.
As a result, there remains a practical gap between (i) standard regression-based I2I models that are easy to train and customize, and (ii) generative frameworks that better handle multimodality but require substantial additional design and computational overhead.

In this paper, we address this gap by reformulating supervised I2I regression through the lens of Rectified Flow (RF), a continuous-time transport framework that learns a velocity field and generates samples by solving an ordinary differential equation (ODE)~\cite{liu2022rectifiedflow,lipman2022flowmatching}.
RF offers an appealing alternative to diffusion: it retains a principled transport view while making the sampling dynamics explicit as an ODE, naturally supporting the use of off-the-shelf ODE solvers and flexible speed--quality trade-offs.
Existing RF formulations have primarily been studied for unconditional generation or broad conditional generation (e.g., text-to-image)~\cite{esser2024scalingrf}, where models typically rely on generic time-conditioned backbones for synthesis, rather than exploiting the rich body of task-specific I2I architectures and modules.
For many I2I tasks, however, strong performance is often driven by task-specialized architectural priors. For example, in low-light image restoration, DarkIR~\cite{feijoo2025darkir} uses an asymmetric encoder--decoder that combines Fourier-domain processing for enhancement with dilated spatial attention for deblurring, achieving strong results on low-light deblurring benchmarks. This motivates the following question: \emph{can we obtain the benefits of continuous-time transport models---in particular, progressive refinement and reduced averaging artifacts---without abandoning the simplicity, stability, and strong inductive biases of existing I2I regressors?}

We answer this question with \textit{Image-to-Image Rectified Flow Reformulation} (I2I-RFR), a plug-in training and inference scheme that recasts a wide range of standard I2I networks within an RF-inspired continuous-time formulation while preserving their regression-oriented design. Given an input image $x$ and a ground-truth image $y$, we construct a noisy mixture $y_t=(1-t)y+t\varepsilon$ with $\varepsilon\sim\mathcal{N}(0,I)$ and $t\in(0,1]$, and feed the concatenation $[x; y_t]$ to an existing I2I backbone. The network is trained to reconstruct $y$ from $[x; y_t]$ with a $t$-reweighted pixel-wise loss such as $\ell_1$, preserving the spirit of standard pixel-wise learning while admitting an RF interpretation via an induced velocity field. In most cases, adopting I2I-RFR requires only expanding the input channels from $C_x$ to $C_x+C_y$ (often $2C$), allowing existing backbones and training recipes to be reused with minimal changes. This yields a practical connection between supervised regression and continuous-time transport, without redesigning the backbone as a generic velocity or noise predictor.

At inference time, we solve the corresponding ODE to progressively refine the prediction. Because $x$ provides informative spatial cues, the state $y_t$ acts as a refinement variable rather than a fully unconstrained generative latent, and we find that a small number of explicit Euler steps already yields strong performance in practice. Moreover, since the concatenated input $[x; y_t]$ implicitly carries information about the corruption level, we observe that explicit time embeddings, which are common in diffusion and RF models~\cite{ho2020ddpm,liu2022rectifiedflow}, are not necessary in our formulation under the evaluated settings. To improve training efficiency, we sample $t$ non-uniformly using inverse-CDF Beta sampling, which places slightly more mass on the high-noise regime ($t\approx1$). In our experiments, this serves as a simple and effective default.

I2I-RFR is designed to be practical: it does \emph{not} require adding a GAN discriminator, redesigning the backbone for velocity or noise prediction, introducing a latent autoencoder, or relying on large pretrained models. Instead, it builds directly on the accumulated engineering and architectural knowledge of supervised I2I models and equips them with few-step ODE-based refinement. 
We validate I2I-RFR across multiple I2I tasks, where it generally improves performance over strong regression baselines. Because the reformulation is broadly applicable across architectures, it can also be extended to video-to-video translation models in the same manner, yielding gains on video restoration tasks.

\section{Related Work}

\subsection{Supervised Image-to-Image Translation}
Many computer vision problems can be cast as image-to-image (I2I) translation, including super-resolution~\cite{DongPAMI2016,liang2021swinir}, image restoration~\cite{chen2022simple,mao2024adarevd}, low-light enhancement~\cite{zhou2022lednet,feijoo2025darkir}, underwater image enhancement~\cite{tang2023underwater,zhao2024toward}, and video restoration~\cite{MaggioniTIP2012,KimECCV2018}. A large body of work in these areas has been developed under paired supervision and relies on regression-style training with pixel-level objectives. This setting has produced a rich ecosystem of backbone architectures and modules tailored to different degradations and application domains.

Classic encoder--decoder designs such as U-Net~\cite{ronneberger2015unet} and residual learning~\cite{he2016resnet} remain widely used building blocks, while recent task-oriented backbones achieve strong performance through specialized attention, multi-scale processing, and frequency-domain priors. For example, SwinIR~\cite{liang2021swinir} targets image super-resolution with Transformer components, and Restormer~\cite{zamir2022restormer} develops an efficient attention architecture for general-purpose image restoration. In domain-specific settings, task-specialized designs can be particularly important; LEDNet~\cite{zhou2022lednet} and DarkIR~\cite{feijoo2025darkir} are representative examples for low-light enhancement and low-light deblurring. These supervised I2I architectures constitute valuable engineering assets, motivating approaches that can improve perceptual quality while preserving the standard regression interface and enabling direct reuse of existing backbones.

\subsection{Perceptual and Adversarial Objectives}
To enhance perceptual fidelity, prior work frequently augments pixel losses with feature-space objectives.
Perceptual losses compare deep features extracted from pretrained recognition networks such as VGG~\cite{SimonyanICLR2015}, and have been popularized for image translation tasks, including style transfer and super-resolution~\cite{johnson2016perceptual}. 
While effective, perceptual losses often only partially mitigate multimodality and may be insufficient under severe degradations or strong information loss.
They also increase training cost due to feature extraction and require task-dependent choices of the feature network, layers, and loss weights.

Another widely used approach is adversarial training. GANs~\cite{GoodfellowPMXWOCB14} introduce a discriminator to encourage outputs that match the target distribution, and conditional GAN formulations have been applied to I2I translation~\cite{IsolaCVPR2017}. GAN-based super-resolution, exemplified by SRGAN~\cite{ledig2017srgan}, can improve realism and recover high-frequency details.
However, adversarial objectives add extra networks and hyperparameters, and can be sensitive to the generator--discriminator balance, potentially leading to instability or artifacts and increasing engineering overhead when adapting to new tasks or backbones.

In contrast to prior approaches that mainly improve regression outputs through additional loss terms, our work reformulates the supervised I2I pipeline itself into a few-step continuous-time refinement procedure. Importantly, the core formulation works with a standard pixel loss alone, yet remains compatible with existing task-specific composite loss designs.

\subsection{Diffusion Models for Conditional I2I Translation}
Diffusion models provide a probabilistic alternative for multimodal conditional targets. DDPM~\cite{ho2020ddpm} trains a denoiser across noise levels, and DDIM~\cite{song2020ddim} enables faster sampling via deterministic trajectories. Conditional diffusion models have been adapted to I2I tasks, including super-resolution (SR3)~\cite{saharia2021sr3} and general I2I translation (Palette)~\cite{saharia2022palette}. These approaches often yield strong perceptual quality, but they typically rely on diffusion-specific pipelines, such as noise schedules, specialized conditioning, and iterative sampling with many denoising steps.
Latent diffusion further introduces an encoder--decoder, often a VAE~\cite{kingma2014vae}, to define and learn a latent space~\cite{rombach2022ldm}, adding extra components and potentially constraining architectural flexibility. Distillation-based acceleration can reduce step counts, but usually requires an additional training stage and careful tuning~\cite{salimans2022progressivedistillation}. Overall, diffusion-based I2I methods can be powerful, yet their pipeline complexity and multi-step inference can make drop-in reuse of task-specialized regression backbones less straightforward and can be costly for large backbones, interactive applications, and video processing.

\subsection{Rectified Flow}
Rectified Flow~\cite{liu2022rectifiedflow} is closely related to Flow Matching~\cite{lipman2022flowmatching} and learns a time-dependent velocity field that transports samples between two endpoint distributions.
Let $x_0 \sim \pi_0$ and $x_1 \sim \pi_1$ denote samples drawn from the source and target distributions, respectively. Following~\cite{liu2022rectifiedflow}, the linear interpolation path is given by
\begin{equation}
x_t = (1-t)x_0 + t x_1,\qquad t\in[0,1].
\label{eq:rf_path}
\end{equation}
The velocity field is parameterized by a neural network $v_\theta(x_t,t)$ and trained by minimizing
\begin{equation}
\min_{\theta}\; \mathbb{E}\big[\,\|(x_1-x_0) - v_\theta(x_t,t)\|\,\big],
\label{eq:rf_obj}
\end{equation}
where the expectation is taken over empirical draws of $(x_0,x_1)$ and the sampling of $t$.
For conditional generation, the same formulation applies by conditioning the velocity field on an external signal $c$ and using $v_\theta(x_t,c,t)$, as in recent rectified-flow models for text-to-image generation~\cite{esser2024scalingrf}. After training, samples are generated by solving the ODE $d x_t = v_\theta(x_t, c, t)\,dt$, which transports $\pi_0$ to $\pi_1$, or vice versa via time reversal.

Most existing rectified-flow formulations are developed for unconditional or weakly conditioned synthesis
(e.g., text-to-image)~\cite{esser2024scalingrf}. 
They rely on generic time-conditioned backbones and operate in settings where the evolving state is treated as a general generative latent. 
In contrast, our setting is paired image-to-image translation with pixel-aligned conditioning. I2I-RFR introduces a supervised I2I reformulation that preserves regression backbones and interprets the ODE state as a refinement variable guided by the input image, rather than as an unconstrained generative
latent.

\section{Proposed Method}
We propose \textit{Image-to-Image Rectified Flow Reformulation} (I2I-RFR), a practical plug-in reformulation that places standard supervised I2I regression within a rectified-flow-inspired continuous-time refinement framework, without redesigning the backbone architecture.
The core idea is to realize rectified-flow transport using a conventional image regression network fed with a pair consisting of the conditioning input and a noise-corrupted target state. This plug-in reformulation requires only minimal changes to existing I2I models, yet provides a continuous-time refinement view that can substantially improve detail preservation across diverse I2I tasks.

\subsection{Image-to-Image Rectified Flow Reformulation}
Let $f_{\theta}$ be a generic I2I translation network that maps an input image
$x\in\mathbb{R}^{C_{\mathrm{in}}\times H\times W}$ to an output image in
$\mathbb{R}^{C_{\mathrm{out}}\times H\times W}$. Given paired training data $(x,y)$ with
$y\in\mathbb{R}^{C_{\mathrm{out}}\times H\times W}$, conventional supervised training minimizes a pixel-wise regression objective:
\begin{equation}
\min_{\theta}\; \mathbb{E}\big[\,\|y - f_\theta(x)\|^{p}_{p}\,\big],
\end{equation}
where $\|\cdot\|_p$ denotes the $\ell_p$ norm and
$\|e\|_p^p = \sum_i |e_i|^p$.
The parameter $p=1$ corresponds to the $\ell_1$ loss and $p=2$
corresponds to the squared $\ell_2$ loss.
In the following, we set $p=1$ throughout and denote the $\ell_1$ norm by $\|\cdot\|$ for simplicity.

Starting from a standard I2I regressor, I2I-RFR introduces a reformulated supervised learning problem in which a generic I2I network $f_{\theta'}$ receives the pair of the input image and a noise-corrupted target state, and is trained to reconstruct the target image. Under this formulation, standard supervised I2I regression can be interpreted through the lens of rectified flow
as a continuous-time transport process, while preserving the original regression-oriented backbone interface. The resulting objective is given by
\begin{equation}
\min_{\theta'}\; \mathbb{E}\bigg[\bigg\lVert \frac{y - f_{\theta'}([x; y_t])}{t}\bigg\rVert\bigg].
\end{equation}
Here, $[a;b]$ denotes channel-wise concatenation, and $y_t$ is a noisy target obtained by mixing the
ground-truth image $y$ with Gaussian noise $\varepsilon\sim\mathcal{N}(0,I)$ using an interpolation
parameter $t\in(0,1]$:
\begin{equation}
y_t = (1-t)\,y + t\,\varepsilon.
\end{equation}
In this formulation, the image translation network $f_{\theta'}$ is identical to the original network
$f_{\theta}$ except for the number of input channels to accommodate the concatenated input:
\begin{align*}
f_{\theta}   &\colon \mathbb{R}^{C_{\mathrm{in}}\times H\times W}\to \mathbb{R}^{C_{\mathrm{out}}\times H\times W},\\
f_{\theta'}  &\colon \mathbb{R}^{(C_{\mathrm{in}}+C_{\mathrm{out}})\times H\times W}\to \mathbb{R}^{C_{\mathrm{out}}\times H\times W}.
\end{align*}
Consequently, I2I-RFR can be applied to existing regression models by (i) expanding the input-channel dimension and (ii) feeding the noise-corrupted targets alongside the inputs during training. This makes the reformulation compatible with a broad range of carefully engineered I2I architectures for tasks such as image super-resolution, restoration, and enhancement. Furthermore, I2I-RFR applies to video-to-video translation models in the same manner and yields gains on video restoration tasks.
The training procedure is summarized in Algorithm~\ref{alg:rfr_train}.

\begin{algorithm}[t]
\caption{Training of image-to-image rectified flow reformulation}
\label{alg:rfr_train}
\begin{algorithmic}[1]
\Require dataset $\mathcal{D}$, 
         I2I network $f_{\theta'}$,
         learning rate $\eta$,
         lower bound $t_{\min} > 0$ for $t$
\Ensure $\theta'$: updated model parameters

\Repeat
    \State Sample $(x, y)$ from $\mathcal{D}$
    \State $u \sim \mathcal{U}(0, 1)$
    \State $\varepsilon \sim \mathcal{N}(0, I)$
    \State $t \gets \max\big(u^{1/2},\, t_{\min}\big)$ \Comment{Inverse-CDF Beta(2, 1) sampling}
    \State $y_t \gets (1 - t)\, y + t\, \varepsilon$
    \State $\theta' \gets \theta' - \eta \nabla_{\theta'} \big\lVert \tfrac{y - f_{\theta'}([x; y_t])}{t} \big\rVert$    \Comment{$[x;y_t]: \rm{Concat}(x, y_t)$}
\Until{convergence}
\end{algorithmic}
\end{algorithm}

\par\smallskip \smallskip
\noindent
\textbf{Parameterization and induced velocity interpretation.}
Given paired data $(x,y)\sim\mathcal D$, we sample $t\in(0,1]$ and $\varepsilon\sim\mathcal N(0,I)$.
To match the rectified-flow notation, we set $c:=x$, $x_0:=y$ and $x_1:=\varepsilon$, and form the linear path $x_t=(1-t)x_0+t x_1$. Along this path, the ground-truth velocity is constant, $v^\star=x_1-x_0$, and we also have the identity $x_t-x_0=t(x_1-x_0)$.
The rectified-flow dynamics are then obtained \emph{implicitly} by defining an induced velocity field $v_{\theta'}(x_t,c,t)=\frac{x_t-f_{\theta'}([c;x_t])}{t}$.
Using $x_t-x_0=t(x_1-x_0)$, we immediately obtain $(x_1-x_0)-v_{\theta'}(x_t,c,t)=\frac{f_{\theta'}([c;x_t]) - x_0}{t}$.
Therefore, minimizing the $t$-reweighted regression loss $\mathbb E\!\left[\left\|\frac{x_0-f_{\theta'}([c;x_t])}{t}\right\|\right]$ is equivalent to minimizing the corresponding induced velocity regression residual $\mathbb E\!\left[\|(x_1-x_0)-v_{\theta'}(x_t,c,t)\|\right]$.
In this sense, the proposed objective admits a rectified-flow interpretation while preserving the interface of a conventional I2I regressor.
Note that the endpoint labeling is purely a convention: swapping $(x_0,x_1)$ corresponds to the time reversal $t\mapsto 1-t$ and flips the sign of the target velocity ($v\mapsto -v$), yielding an equivalent optimization problem. In this paper, we choose $x_0$ as the data endpoint and
$x_1$ as the noise endpoint mainly for notational simplicity. In practice, at inference time, we initialize at the prior endpoint and integrate backward in time. For numerical stability during training, we clamp $t\ge t_{\min}$ to avoid the singularity as $t\to 0$.

\par\smallskip
\noindent
\textbf{Time conditioning in I2I-RFR.}
Diffusion models and many rectified-flow implementations often condition the network on the noise level by feeding $t$ through a time-embedding module. In I2I-RFR, however, the input is the channel-wise concatenation $[x; y_t]$: the input image $x$ provides strong spatial priors about the target $y$, while the corruption level is directly reflected in $y_t$. As a result, explicit time embeddings are often unnecessary in our supervised I2I setting. Empirically, adding time embeddings provides no consistent improvement across our tasks and backbones, and can slightly degrade performance or increase optimization variance. We therefore omit explicit time embeddings in our default model, which simplifies implementation by avoiding an additional embedding module and facilitates the reuse of existing I2I backbones with minimal changes.

\par\smallskip
\noindent
\textbf{Sampling of the interpolation parameter.}
We construct noisy inputs via $y_t=(1-t)y+t\varepsilon$ and train the regression network $f_{\theta'}([x,y_t])$ with a loss that contains an explicit factor of $t^{-p}$, e.g., $\big\|\tfrac{y-f_{\theta'}([x;y_t])}{t}\big\|_p^p$.
If $t$ were sampled uniformly on $(0,1]$, small $t$ values would receive disproportionately large weight through $t^{-p}$ and could induce high gradient variance. To stabilize optimization, we adopt a
non-uniform sampling strategy based on a Beta distribution:
\begin{equation}
t \sim \mathrm{Beta}(p+1,\,1), \qquad t \leftarrow \max(t,t_{\min}),
\end{equation}
whose density is proportional to $t^p$ on $[0,1]$ and thus counterbalances the $t^{-p}$ factor in the loss, making the effective weighting approximately constant. We implement this sampling via inverse-CDF: $u\sim\mathcal{U}(0,1)$ and $t=u^{1/(p+1)}$, followed by clamping to $t_{\min}$ to avoid numerical issues near $t=0$. As a practical side effect, this sampling provides sufficient coverage of the high-noise regime ($t\approx 1$), which is important for robust refinement.

\begin{figure*}[t]
    \centering
    \begin{tabular}{cc}
        \includegraphics[width=0.49\linewidth]{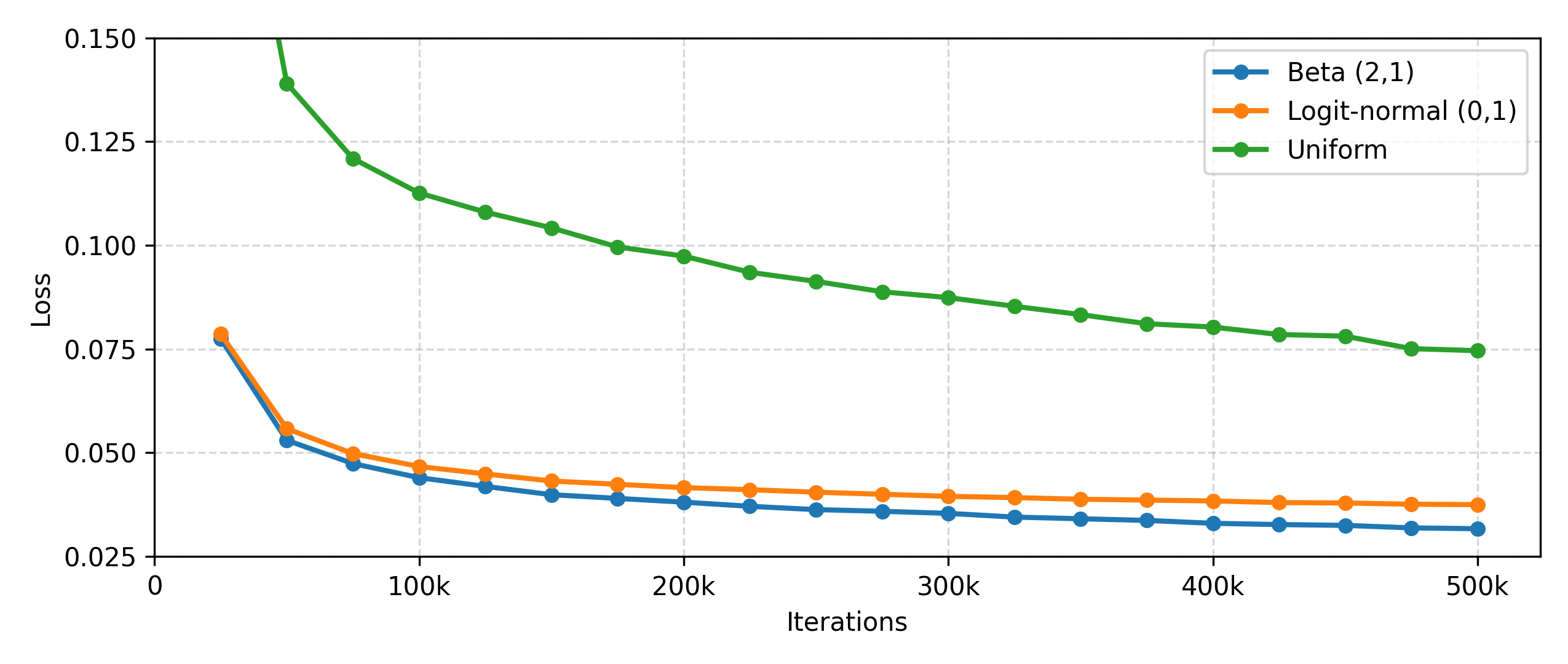} &
        \includegraphics[width=0.49\linewidth]{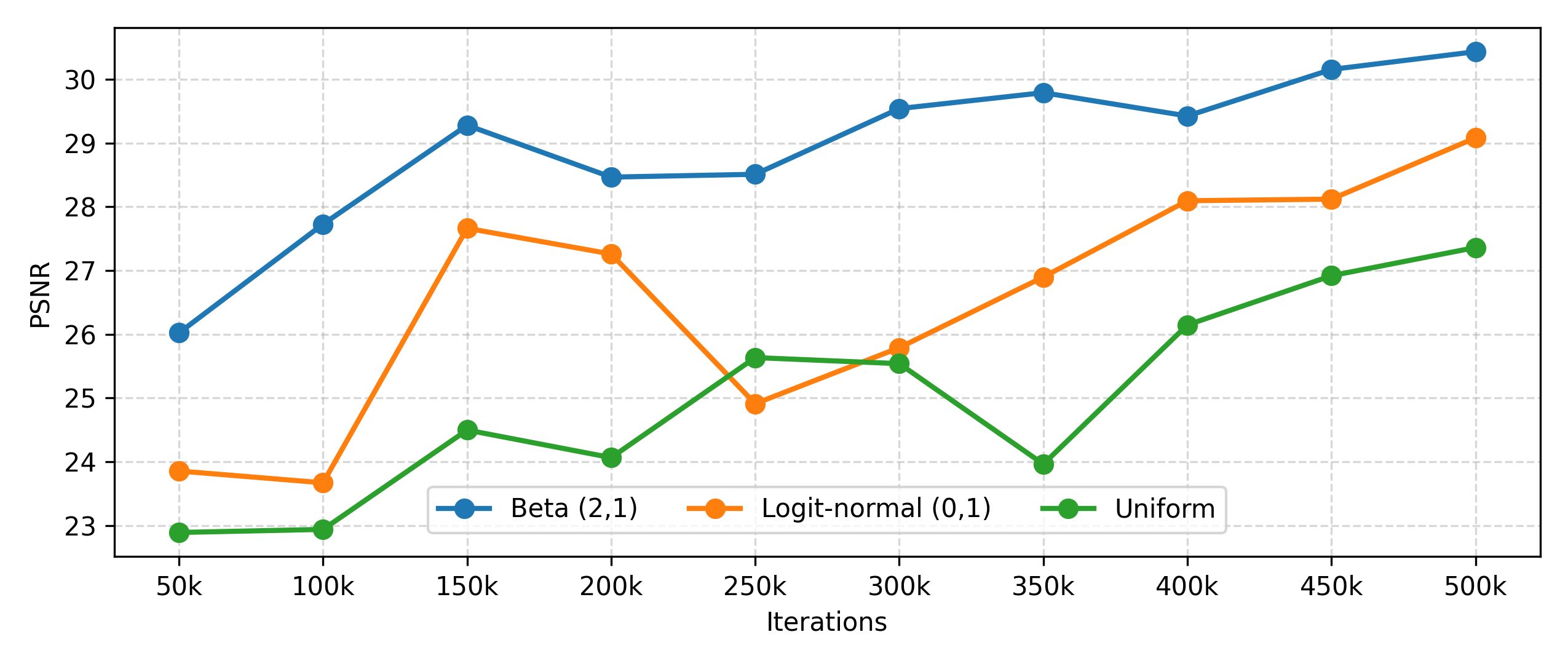} \\
    \end{tabular}
    \vspace{-4mm}
    \caption{Comparison of sampling strategies for training (left) and validation (right).}
    \vspace{-5mm}
    \label{fig:sampling_strategy}
\end{figure*}

Figure~\ref{fig:sampling_strategy} provides an illustrative comparison of $t$-sampling strategies on low-light enhancement with LEDNet~\cite{zhou2022lednet}.
The plot shows the training loss and validation PSNR as functions of training iterations for $\mathrm{Beta}(2,1)$, logit-normal$(0,1)$, and uniform sampling under the same training budget. In this setting, $\mathrm{Beta}(2,1)$ sampling shows smoother convergence and achieves the best validation PSNR among the compared strategies. 
We do not claim that $\mathrm{Beta}(2,1)$ sampling is universally optimal; rather, it serves as an effective default in our experiments.

\par\smallskip
\noindent
\textbf{Relationship to $x_0$-prediction in rectified flow and diffusion models.}
Predicting the clean image, i.e., $x_0$-prediction, is a recognized alternative parameterization in diffusion and flow-based models, but it has been less commonly used as the primary training target in modern large-scale generative settings. For example, DDPM reports that directly predicting $x_0$ led to worse sample quality in early experiments and adopts noise prediction as the default parameterization~\cite{ho2020ddpm}; many subsequent works follow this choice~\cite{cao2022survey}.
Alternative targets such as $v$-prediction have also been proposed to improve stability when sampling with few steps~\cite{salimans2022progressivedistillation}. More broadly, flow-based frameworks such as Flow Matching and Rectified Flow are typically trained by regressing a time-dependent vector field~\cite{lipman2022flowmatching,liu2022rectifiedflow}, including recent large-scale rectified-flow Transformers for text-to-image generation~\cite{esser2024scalingrf}.
In contrast, our work targets supervised image-to-image translation with a pixel-space condition $x$ that provides informative, spatially aligned cues. Under this setting, the ODE state $y_t$ behaves as a refinement variable rather than a fully unconstrained generative latent, making an $x_0$-predictor particularly effective in practice. Thus, our contribution is not to advocate $x_0$-prediction universally, as discussed in recent revisits for unconditional and class-conditional generation~\cite{li2025backtobasics}, but to propose an image-to-image rectified flow reformulation that upgrades standard regression backbones with minimal architectural changes. Moreover, our formulation does not require explicit time embeddings and introduces a Beta-based $t$-sampling strategy tailored to the reformulated objective, which helps stabilize training.

\subsection{Inference}
Given an input image $x$, we generate the output by solving the induced rectified-flow ODE with a
fixed-step explicit Euler integrator, as summarized in Algorithm~\ref{alg:rf_inference}.
We initialize the state at the prior endpoint ($t=1$) by sampling $y_{t}\sim\mathcal N(0,I)$, and integrate
backward in time from $t=1$ toward $t=0$ using $N$ uniform steps with step size $\Delta t = 1/N$.
At step $n$, we set $t_n = 1-\frac{n}{N}$ and evaluate the induced velocity field as
$v=\frac{y_{t_n}-f_{\theta'}([x; y_{t_n}])}{t_n}$.
We then update $y_{t_{n+1}} \leftarrow y_{t_n} - \Delta t\, v$, which corresponds to an explicit Euler
discretization of the backward-time dynamics.
Note that this update moves $y_{t_n}$ toward the network prediction $f_{\theta'}([x;y_{t_n}])$, since
$y_{t_n}-\Delta t\,\frac{y_{t_n}-f_{\theta'}([x;y_{t_n}])}{t_n}
= y_{t_n}+\Delta t\,\frac{f_{\theta'}([x;y_{t_n}]) - y_{t_n}}{t_n}$.
After $N$ steps, the final state (at $t\approx 0$) is taken as the output $\hat y$.

\begin{algorithm}[t]
\caption{Inference of Image-to-Image Rectified Flow}
\label{alg:rf_inference}
\begin{algorithmic}[1]
\Require input image $x$,
         trained I2I network $f_{\theta'}$,
         number of time steps $N$
\Ensure output image $\hat{y}$

\State Sample $y_{t} \sim \mathcal{N}(0, I)$ \Comment{initialization}

\For{$n = 0$ to $N-1$}
    \State $t \gets 1 - \frac{n}{N}$
    \State $v \gets \tfrac{y_t - f_{\theta'}([x; y_t])}{t}$
    \State $y_t \gets y_t - \frac{1}{N} \, v$
\EndFor

\State $\hat{y} \gets y_t$
\end{algorithmic}
\end{algorithm}

\par\smallskip
\noindent
\textbf{Efficiency.}
In standard rectified flow, inference typically requires tens of ODE solver steps
(e.g., 25 steps). Although this is fewer than the 50--250 steps commonly used by
diffusion models with DDIM sampling, it can still be costly for large networks,
interactive applications, and video processing. Distillation-based acceleration can
reduce the number of steps, but it introduces an additional training stage and
requires careful tuning.

In contrast, our formulation works well with only a small number of explicit Euler
steps in practice. We attribute this efficiency to the informative spatial cues in the
input image $x$, which make the ODE state $y_t$ behave as a refinement variable
rather than a fully unconstrained generative latent. By providing $x$ alongside the
current state via channel-wise concatenation, a few refinement steps are often
sufficient. Throughout the paper, we use $N=3$ as the default setting.

\section{Results}
I2I-RFR is evaluated on a diverse set of image-to-image (I2I) and video-to-video translation tasks, including \textit{image super-resolution}, \textit{deblurring}, \textit{low-light enhancement}, \textit{underwater enhancement}, and \textit{video restoration}. For each task, I2I-RFR is applied to strong recent backbones, including task-specialized architectures based on convolutional networks and Transformers. To assess robustness across model complexity and training pipelines, comparisons also include (i) a canonical U-Net~\cite{ronneberger2015unet} as a simple regression baseline and (ii) Palette~\cite{saharia2022palette}, a representative diffusion-based framework for conditional I2I translation that unifies a wide range of I2I tasks within a single diffusion formulation. 
These baselines help isolate the benefit of the reformulation from backbone capacity and highlight practical differences in pipeline design, as diffusion-based I2I typically relies on diffusion-specific components such as noise schedules, explicit noise-level conditioning, and many-step denoising trajectories.

All methods are trained from scratch unless official checkpoints are used as reference baselines (noted per task). For fair comparison, models that require additional inputs beyond the benchmark setting (e.g., text prompts) are excluded, and pretrained models that rely on large-scale external data are not used. The goal is not to exhaustively compete with every task-specific state-of-the-art, but to demonstrate that I2I-RFR can provide gains across diverse tasks and backbones under simple and stable supervised training. In addition, while I2I-RFR performs strongly with an $\ell_1$ loss alone, a variant with commonly used composite losses is also reported to show that the reformulation is compatible with additional objectives such as perceptual losses.

\noindent
\textbf{Implementation details.}
Unless otherwise specified, all models are optimized with Adam using an initial learning rate of $10^{-4}$ and a cosine learning-rate schedule down to $10^{-6}$. Task-dependent settings such as input resolution, batch size, data augmentation, and training iterations follow standard practice for each benchmark and backbone. For I2I-RFR, $t$ is sampled as $t\sim\mathrm{Beta}(2,1)$ with $t_{\min}=10^{-3}$, and the default loss is the $t^{-1}$-reweighted $\ell_1$ distance. 
At inference time, the induced ODE is solved with a fixed-step explicit Euler integrator. Unless stated otherwise, we use $N=3$ steps, which serves as a practical compromise between speed and refinement quality; additional analyses of runtime and inference depth are provided in the appendix. 
For quantitative evaluation, we use a fixed random seed and generate a single sample per input.
Classifier-free guidance is not used, and explicit time embedding modules are omitted to keep the architecture changes minimal and preserve direct reuse of existing regression backbones.

\begin{table*}[t]
\centering
\caption{Super-resolution results on standard benchmarks.
Best values are in \textcolor{red}{red}, and second-best values are in \textcolor{blue}{blue}.
Values in parentheses denote the relative changes compared to the original model.
}
\vspace{-3mm}
\label{tab:sr_results}
\resizebox{\textwidth}{!}{
\begin{tabular}{l|ccc|ccc|ccc|ccc}
    \toprule
    & \multicolumn{3}{c}{Set5} & \multicolumn{3}{c}{Set14} & \multicolumn{3}{c}{BSD100} & \multicolumn{3}{c}{Urban100} \\
    \midrule
    Model & PSNR$\uparrow$ & SSIM$\uparrow$ & LPIPS$\downarrow$ & PSNR$\uparrow$ & SSIM$\uparrow$ & LPIPS$\downarrow$ & PSNR$\uparrow$ & SSIM$\uparrow$ & LPIPS$\downarrow$ & PSNR$\uparrow$ & SSIM$\uparrow$ & LPIPS$\downarrow$ \\
    \midrule
    NLSN~\cite{mei2021image} & \textcolor{red}{32.84} & 0.9020 & 0.1694 & \textcolor{red}{29.05} & \textcolor{blue}{0.7920} & 0.2715 & \textcolor{blue}{27.87} & 0.7459 & 0.3595 & \textcolor{blue}{27.22} & 0.8176 & 0.1945 \\
    NLSN + I2I-RFR & \makecell{32.58 \\ (-0.26)} & \makecell{0.8998 \\ (-0.0022)} & \makecell{\textcolor{blue}{0.1676} \\ \textbf{(-0.0018)}} & \makecell{28.88 \\ (-0.17)} & \makecell{0.7888 \\ (-0.0032)} & \makecell{0.2662 \\ \textbf{(-0.0053)}} & \makecell{27.75 \\ (-0.12)} & \makecell{0.7430 \\ (-0.0029)} & \makecell{0.3498 \\ \textbf{(-0.0097)}} & \makecell{26.83 \\ (-0.39)} & \makecell{0.8079 \\ (-0.0097)} & \makecell{0.1996 \\ (+0.0051)} \\
    \midrule
    SwinIR~\cite{liang2021swinir} & \textcolor{blue}{32.80} & \textcolor{blue}{0.9025} & 0.1687 & 29.00 & \textcolor{red}{0.7922} & 0.2703 & \textcolor{red}{27.89} & \textcolor{red}{0.7469} & 0.3578 & \textcolor{red}{27.24} & \textcolor{red}{0.8199} & 0.1948 \\
    SwinIR + I2I-RFR & \makecell{\textcolor{red}{32.84} \\ \textbf{(+0.04)}} & \makecell{\textcolor{red}{0.9028} \\ \textbf{(+0.0003)}} & \makecell{\textcolor{red}{0.1665} \\ \textbf{(-0.0022)}} & \makecell{\textcolor{blue}{29.01} \\ \textbf{(+0.01)}} & \makecell{0.7916 \\ (-0.0006)} & \makecell{0.2621 \\ \textbf{(-0.0082)}} & \makecell{27.85 \\ (-0.04)} & \makecell{\textcolor{blue}{0.7461} \\ (-0.0008)} & \makecell{0.3497 \\ \textbf{(-0.0081)}} & \makecell{27.19 \\ (-0.05)} & \makecell{\textcolor{blue}{0.8180} \\ (-0.0019)} & \makecell{\textcolor{red}{0.1897} \\ \textbf{(-0.0051)}} \\
    \midrule
    U-Net~\cite{ronneberger2015unet} & 32.09 & 0.8934 & 0.1767 & 28.57 & 0.7806 & 0.2891 & 27.53 & 0.7347 & 0.3795 & 25.97 & 0.7810 & 0.2414 \\
    U-Net + I2I-RFR & \makecell{31.63 \\ (-0.46)} & \makecell{0.8884 \\ (-0.0050)} & \makecell{\textcolor{red}{0.1665} \\ \textbf{(-0.0102)}} & \makecell{28.16 \\ (-0.41)} & \makecell{0.7704 \\ (-0.0102)} & \makecell{\textcolor{blue}{0.2606} \\ \textbf{(-0.0285)}} & \makecell{27.18 \\ (-0.35)} & \makecell{0.7244 \\ (-0.0103)} & \makecell{\textcolor{blue}{0.3404} \\ \textbf{(-0.0391)}} & \makecell{25.55 \\ (-0.42)} & \makecell{0.7707 \\ (-0.0103)} & \makecell{0.2218 \\ \textbf{(-0.0196)}} \\
    \midrule
    Palette~\cite{saharia2022palette} & 22.28 & 0.6582 & 0.2079 & 21.49 & 0.5139 & 0.3059 & 24.72 & 0.6063 & 0.3967 & 19.58 & 0.5193 & 0.3035 \\
    Palette + I2I-RFR & \makecell{32.09 \\ \textbf{(+9.81)}} & \makecell{0.8959 \\ \textbf{(+0.2377)}} & \makecell{0.1695 \\ \textbf{(-0.0384)}} & \makecell{28.47 \\ \textbf{(+6.98)}} & \makecell{0.7788 \\ \textbf{(+0.2649)}} & \makecell{\textcolor{red}{0.2459} \\ \textbf{(-0.0600)}} & \makecell{27.40 \\ \textbf{(+2.68)}} & \makecell{0.7339 \\ \textbf{(+0.1276)}} & \makecell{\textcolor{red}{0.3299} \\ \textbf{(-0.0668)}} & \makecell{26.40 \\ \textbf{(+6.82)}} & \makecell{0.7973 \\ \textbf{(+0.2780)}} & \makecell{\textcolor{blue}{0.1914} \\ \textbf{(-0.1121)}} \\
    \bottomrule
\end{tabular}
}
\vspace{-6mm}
\end{table*}

\par\smallskip
\noindent
\textbf{Image Super-resolution.}
Image super-resolution aims to recover high-resolution (HR) images from low-resolution (LR) inputs. 
Representative regression-based SR models, NLSN~\cite{mei2021image} and SwinIR~\cite{liang2021swinir}, are selected as baselines, where the former is CNN-based and the latter is Transformer-based.
I2I-RFR is applied to both models by expanding the input layer to accept a 6-channel input.
Both the original models and their I2I-RFR variants are trained on DF2K~\cite{agustsson2017ntire}.
During training, HR patches are randomly cropped to $256\times256$ pixels, and the corresponding LR inputs are generated by $\times4$ bicubic downsampling. For I2I-RFR, the interpolation state is constructed in the LR domain to match the model input shape: the network is fed with $[x^{\mathrm{LR}}; y_t^{\mathrm{LR}}]$, while the prediction target and the loss are defined in the HR domain, i.e., the model predicts the clean HR image $y^{\mathrm{HR}}$ from the LR input state.

Table~\ref{tab:sr_results} shows that I2I-RFR generally improves LPIPS across backbones, while PSNR and SSIM remain similar or can slightly decrease for some models, reflecting the perception--distortion trade-off in ill-posed SR~\cite{blau2018perception}. In particular, SwinIR with I2I-RFR provides the best overall balance among the compared methods, improving LPIPS on all four benchmarks while largely preserving distortion metrics. Notably, applying I2I-RFR to the Palette denoising backbone substantially improves SR quality, suggesting that the regression-preserving transport formulation can be effective even when starting from a diffusion-oriented U-Net.

\par\smallskip
\noindent
\textbf{Image Deblurring.}
We evaluate I2I-RFR on three representative deblurring backbones, NAFNet~\cite{chen2022simple},
Restormer~\cite{zamir2022restormer}, and FFTFormer~\cite{kong2023efficient}. Training follows the GoPro
protocol~\cite{Nah_2017_CVPR}, and evaluation is performed on RealBlur-J~\cite{RealBlur}, providing a cross-dataset test
relative to GoPro training. I2I-RFR is applied by expanding the input channels from 3 to 6 and
replacing the pixel loss with its $t$-reweighted version, without modifying the backbone structure.
Metrics include PSNR, SSIM, and LPIPS.
As shown in Table~\ref{tab:deblur_results}, I2I-RFR consistently improves perceptual quality over the baseline models.

\begin{table*}[t]
\centering
\caption{Image deblurring results on RealBlur-J benchmark~\cite{RealBlur}.
Best values are in \textcolor{red}{red}, and second-best values are in \textcolor{blue}{blue}.
Values in parentheses denote the relative changes compared to the original model.
}
\vspace{-3mm}
\label{tab:deblur_results}
\setlength{\tabcolsep}{6pt} 
    \begin{tabular}{llll}
        \toprule
        Model & PSNR$\uparrow$ & SSIM$\uparrow$ & LPIPS$\downarrow$ \\
        \midrule
        NAFNet~\cite{chen2022simple} & 28.62 & 0.8683 & 0.1557 \\
        NAFNet + I2I-RFR & 28.74 \textbf{(+0.12)} & 0.8737 \textbf{(+0.0054)} & 0.1449 \textbf{(-0.0108)} \\
        \midrule
        Restormer~\cite{zamir2022restormer} & 28.73 & 0.8677 & 0.1522 \\
        Restormer + I2I-RFR & \textcolor{blue}{29.04} \textbf{(+0.31)} & \textcolor{blue}{0.8812} \textbf{(+0.0135)} & \textcolor{red}{0.1408} \textbf{(-0.0114)} \\
        \midrule
        FFTFormer~\cite{kong2023efficient} & 28.19 & 0.8535 & 0.1691 \\
        FFTFormer + I2I-RFR & 28.58 \textbf{(+0.39)} & 0.8685 \textbf{(+0.0150)} & 0.1487 \textbf{(-0.0204)} \\
        \midrule
        U-Net~\cite{ronneberger2015unet} & 28.03 & 0.8440 & 0.1848 \\
        U-Net + I2I-RFR & 27.93 (-0.10) & 0.8476 \textbf{(+0.0036)} & 0.1753 \textbf{(-0.0095)} \\
        \midrule
        Palette~\cite{saharia2022palette} & 26.29 & 0.6313 & 0.2826 \\
        Palette + I2I-RFR & \textcolor{red}{29.05} \textbf{(+2.76)} & \textcolor{red}{0.8831} \textbf{(+0.2518)} & \textcolor{blue}{0.1436} \textbf{(-0.1390)} \\
        \bottomrule
    \end{tabular}
\vspace{-3mm}
\end{table*}

\begin{table*}[t]
\centering
\caption{Low-light blur enhancement results on LOLBlur test set.
Best values are in \textcolor{red}{red}, and second-best values are in \textcolor{blue}{blue}.
Values in parentheses denote the relative changes compared to the original model.
}
\vspace{-3mm}
\label{tab:lowlight_results}
\setlength{\tabcolsep}{6pt}
    \begin{tabular}{llll}
        \toprule
        Model & PSNR$\uparrow$ & SSIM$\uparrow$ & LPIPS$\downarrow$ \\
        \midrule
        LEDNet~\cite{zhou2022lednet} & 26.04 & 0.845 & 0.159 \\
        LEDNet + I2I-RFR & \textcolor{blue}{27.42} \textbf{(+1.38)} & 0.874 \textbf{(+0.029)} & 0.138 \textbf{(-0.021)} \\
        LEDNet + I2I-RFR-CL & 26.56 \textbf{(+0.52)} & 0.869 \textbf{(+0.024)} & 0.140 \textbf{(-0.019)} \\
        \midrule
        DarkIR~\cite{feijoo2025darkir} & 27.31 & 0.896 & 0.111 \\
        DarkIR + I2I-RFR & 27.19 (-0.12) & \textcolor{blue}{0.897} \textbf{(+0.001)} & \textcolor{blue}{0.106} \textbf{(-0.005)} \\
        DarkIR + I2I-RFR-CL & \textcolor{red}{27.72} \textbf{(+0.41)} & \textcolor{red}{0.900} \textbf{(+0.004)} & \textcolor{red}{0.103} \textbf{(-0.008)} \\
        \midrule
        U-Net~\cite{ronneberger2015unet} & 20.45 & 0.764 & 0.267 \\
        U-Net + I2I-RFR & 22.48 \textbf{(+2.03)} & 0.826 \textbf{(+0.062)} & 0.184 \textbf{(-0.083)} \\
        \midrule
        Palette~\cite{saharia2022palette} & 8.03 & 0.007 & 0.936 \\
        Palette + I2I-RFR & 26.88 \textbf{(+18.85)} & 0.875 \textbf{(+0.868)} & 0.141 \textbf{(-0.795)} \\
        \bottomrule
    \end{tabular}
\vspace{-3mm}
\end{table*}

\par\smallskip
\noindent
\textbf{Low-Light Blur Enhancement.}
Low-light blur enhancement targets images that are simultaneously under-exposed and blurred (e.g., due to motion or defocus). The goal is to recover a clean, well-lit image by correcting illumination and removing blur to restore sharp edges and fine textures. Because both degradations destroy information and amplify noise, the task is highly ill-posed and requires balancing detail recovery against artifact suppression.

We evaluate I2I-RFR on two strong low-light deblurring models, LEDNet~\cite{zhou2022lednet} and
DarkIR~\cite{feijoo2025darkir}, following the LOLBlur training and evaluation protocol~\cite{zhou2022lednet}.
Official checkpoints are used as reference baselines, and the corresponding I2I-RFR variants are trained on the same LOLBlur training split. For the composite-loss setting, the original loss design is kept, and only the pixel term is replaced by its $t$-reweighted counterpart, denoted as I2I-RFR-CL. Evaluation is performed on the LOLBlur test set using PSNR, SSIM, and LPIPS.

Table~\ref{tab:lowlight_results} shows that I2I-RFR substantially improves LEDNet, while for DarkIR the plain I2I-RFR variant mainly improves perceptual quality while keeping distortion metrics comparable to the original model.
Notably, DarkIR is already a state-of-the-art baseline for this benchmark, and combining it with the composite-loss version of I2I-RFR further improves all three metrics, thereby pushing the state of the art further within this setting. This suggests that the proposed reformulation remains effective even on a strong task-specialized model, and that its benefit becomes clearest when the original loss design is preserved. A plausible reason is that DarkIR is tightly coupled with its original task-specific objective design, so reweighting only the pixel term while keeping the remaining composite losses intact leads to a better match with the backbone's inductive bias and training dynamics.

\par\smallskip
\noindent
\textbf{Underwater Image Enhancement.}
We evaluate I2I-RFR on the LSUI benchmark~\cite{peng2023ushape} using two strong recent baselines: the
diffusion-based method by Tang \textit{et al.}~\cite{tang2023underwater} and SFGNet~\cite{zhao2024toward}.
All models are trained on the LSUI training split and evaluated on the LSUI test set; in addition, we
report results on the UIEB test set to assess cross-dataset generalization. For Tang \textit{et al.}, the
original DDPM training is replaced with I2I-RFR training using the same denoising backbone, enabling a
direct architectural comparison. For SFGNet, I2I-RFR augments the input with the noisy target state while
keeping the backbone unchanged.

\begin{table}[t]
\centering
\caption{Underwater image enhancement results on LSUI test set and UIEB test set.
Best values are in \textcolor{red}{red}, and second-best values are in \textcolor{blue}{blue}.
Values in parentheses denote the relative changes compared to the original model.
}
\vspace{-3mm}
\label{tab:underwater_results}
\setlength{\tabcolsep}{10pt}
\resizebox{\textwidth}{!}{
    \begin{tabular}{l|ccc|ccc}
        \toprule
        & \multicolumn{3}{c}{LSUI Test Set} & \multicolumn{3}{c}{UIEB Test Set} \\
        \toprule
        Model & PSNR$\uparrow$ & SSIM$\uparrow$ & LPIPS$\downarrow$ & PSNR$\uparrow$ & SSIM$\uparrow$ & LPIPS$\downarrow$ \\
        \midrule
        Tang et al.~\cite{tang2023underwater} & 18.34 & 0.8320 & 0.1413 & 14.63 & 0.7110 & 0.2339 \\
        Tang et al. + I2I-RFR & \makecell{\textcolor{red}{29.27} \\ \textbf{(+10.93)}} & \makecell{\textcolor{red}{0.9381} \\ \textbf{(+0.1061)}} & \makecell{\textcolor{blue}{0.0838} \\ \textbf{(-0.0575)}} & \makecell{19.56 \\ \textbf{(+4.93)}} & \makecell{0.8462 \\ \textbf{(+0.1352)}} & \makecell{\textcolor{blue}{0.1637} \\ \textbf{(-0.0702)}} \\
        \midrule
        SFGNet~\cite{zhao2024toward} & 27.54 & 0.9293 & \textcolor{red}{0.0834} & 18.82 & 0.8315 & 0.1738 \\
        SFGNet + I2I-RFR & \makecell{26.09 \\ (-1.45)} & \makecell{0.9190 \\ (-0.0103)} & \makecell{0.1164 \\ (+0.0330)} & \makecell{\textcolor{blue}{19.69} \\ \textbf{(+0.87)}} & \makecell{\textcolor{red}{0.8515} \\ \textbf{(+0.0200)}} & \makecell{0.1640 \\ \textbf{(-0.0098)}} \\
        SFGNet + I2I-RFR-CL & \makecell{26.40 \\ (-1.14)} & \makecell{0.9203 \\ (-0.0090)} & \makecell{0.0907 \\ (+0.0073)} & \makecell{19.63 \\ \textbf{(+0.81)}} & \makecell{\textcolor{blue}{0.8470} \\ \textbf{(+0.0155)}} & \makecell{0.1674 \\ \textbf{(-0.0064)}} \\
        \midrule
        U-Net~\cite{ronneberger2015unet} & 27.42 & 0.9244 & 0.1110 & 17.20 & 0.7957 & 0.2352 \\
        U-Net + I2I-RFR & \makecell{28.34 \\ \textbf{(+0.92)}} & \makecell{0.9322 \\ \textbf{(+0.0078)}} & \makecell{0.0893 \\ \textbf{(-0.0217)}} & \makecell{17.69 \\ \textbf{(+0.49)}} & \makecell{0.8092 \\ \textbf{(+0.0135)}} & \makecell{0.2203 \\ \textbf{(-0.0149)}} \\
        \midrule
        Palette~\cite{saharia2022palette} & 23.36 & 0.7424 & 0.1838 & 17.73 & 0.6542 & 0.3113 \\
        Palette + I2I-RFR & \makecell{\textcolor{blue}{29.26} \\ \textbf{(+5.90)}} & \makecell{\textcolor{blue}{0.9376} \\ \textbf{(+0.1952)}} & \makecell{0.0866 \\ \textbf{(-0.0972)}} & \makecell{\textcolor{red}{20.00} \\ \textbf{(+2.27)}} & \makecell{0.8469 \\ \textbf{(+0.1927)}} & \makecell{\textcolor{red}{0.1632} \\ \textbf{(-0.1481)}} \\
        \bottomrule
    \end{tabular}
}
\vspace{-3mm}
\end{table}

As shown in Table~\ref{tab:underwater_results}, I2I-RFR yields large gains for the diffusion backbone retrained under the proposed formulation, and also improves the U-Net and Palette baselines on both LSUI and UIEB. For SFGNet, however, the effect is mixed: while the UIEB results improve, the LSUI in-domain metrics degrade under the plain I2I-RFR formulation. This suggests that the benefit of the reformulation can depend on how strongly the original backbone and training objective are tuned to the task. In particular, SFGNet appears to be more sensitive to replacing its original optimization recipe with the plain reformulated objective, while the composite-loss variant better preserves the original design and recovers much of the lost performance.

\par\smallskip
\noindent
\textbf{Video Restoration.}
\label{sec:video_restoration}
We evaluate I2I-RFR on old film restoration using RTN~\cite{wan2022bringing} and the more recent RRTN~\cite{lin2024restoring}.
Training follows the RRTN protocol on REDS~\cite{Nah_2019_CVPR_Workshops} with the same synthetic
degradation pipeline, and evaluation is performed on DAVIS with matched degradations. I2I-RFR is applied by expanding the input channels and replacing the pixel loss with its $t$-reweighted version. 
For the composite-loss setting, the original loss design is kept, and only the pixel term is reweighted, denoted as I2I-RFR-CL. Metrics include PSNR, SSIM, and LPIPS.
As shown in Table~\ref{tab:video_restoration_results}, I2I-RFR improves the quality of the baselines in most cases and can even match or surpass GAN-trained variants without adversarial training.

\begin{table}[t]
\centering
\caption{Video restoration results on DAVIS test set.
Best values are in \textcolor{red}{red}, and second-best values are in \textcolor{blue}{blue}.
Values in parentheses denote the relative changes compared to the original model.
}
\vspace{-3mm}
\label{tab:video_restoration_results}
\setlength{\tabcolsep}{10pt}
\resizebox{\textwidth}{!}{
    \begin{tabular}{llll}
        \toprule
        Model & PSNR$\uparrow$ & SSIM$\uparrow$ & LPIPS$\downarrow$ \\
        \midrule
        RTN~\cite{wan2022bringing} & 24.54 & 0.8294 & \textcolor{blue}{0.1562} \\
        RTN + I2I-RFR & 25.05 \textbf{(+0.51)} & 0.8330 \textbf{(+0.0036)} & 0.2100 (+0.0538) \\
        RTN + I2I-RFR-CL & 25.52 \textbf{(+0.98)} & 0.8395 \textbf{(+0.0101)} & 0.1728 (+0.0166) \\
        \midrule
        RRTN~\cite{lin2024restoring} & 22.66 & 0.7453 & 0.2379 \\
        RRTN + I2I-RFR & \textcolor{red}{25.87} \textbf{(+3.21)} & \textcolor{red}{0.8659} \textbf{(+0.1206)} & 0.1699 \textbf{(-0.0680)} \\
        RRTN + I2I-RFR-CL & \textcolor{blue}{25.80} \textbf{(+3.14)} & \textcolor{blue}{0.8505} \textbf{(+0.1052)} & \textcolor{red}{0.1456} \textbf{(-0.0923)} \\
        \bottomrule
    \end{tabular}
}
\vspace{-2mm}
\end{table}

\begin{table}[t]
\centering
\caption{Parameterization effect.
Best values are in \textcolor{red}{red}, and second-best are in \textcolor{blue}{blue}.
}
\vspace{-2mm}
\label{tab:parameterization_results_main}
\setlength{\tabcolsep}{20pt}
\resizebox{\textwidth}{!}{
    \begin{tabular}{llll}
        \toprule
        & PSNR$\uparrow$ & SSIM$\uparrow$ &  LPIPS$\downarrow$ \\
        \midrule
        \multicolumn{4}{l}{Low-light image enhancement results on LOLBlur test set.} \\
        \midrule
        LEDNet + I2I-RFR & \textcolor{red}{27.42} & \textcolor{blue}{0.874} & \textcolor{blue}{0.138} \\
        LEDNet + I2I-RFR $v$-pred & 11.57 & 0.060 & 0.803 \\
        \midrule
        DarkIR + I2I-RFR & \textcolor{blue}{27.19} & \textcolor{red}{0.897}  & \textcolor{red}{0.106} \\
        DarkIR + I2I-RFR $v$-pred & 16.73 & 0.431 & 0.512 \\
        \midrule
        \multicolumn{4}{l}{Old film restoration results on DAVIS test set.} \\
        \midrule
        RTN + I2I-RFR & \textcolor{blue}{25.05} & 0.8330 & 0.2100 \\
        RTN + I2I-RFR $v$-pred & 20.76 & 0.7127 & 0.3025 \\
        \midrule
        RRTN + I2I-RFR & \textcolor{red}{25.87} & \textcolor{red}{0.8659} & \textcolor{blue}{0.1699} \\
        RRTN + I2I-RFR $v$-pred & 24.72 & \textcolor{blue}{0.8566} & \textcolor{red}{0.1565} \\
        \bottomrule
    \end{tabular}
}
\vspace{-2mm}
\end{table}

\newcommand{\repfigvx}[1]{\includegraphics[width=0.325\linewidth]{figs/results/Comp_v_x/#1}}
\begin{figure*}[t]
   \centering
   \setlength{\tabcolsep}{1pt}
   \begin{tabular}{ccc}
   \repfigvx{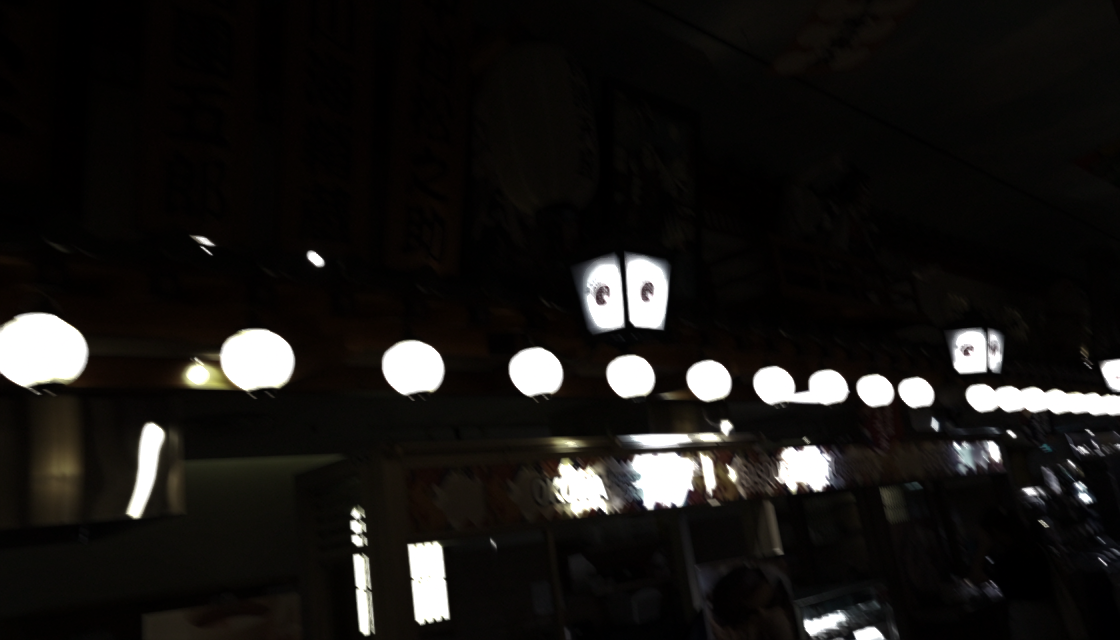} &
   \repfigvx{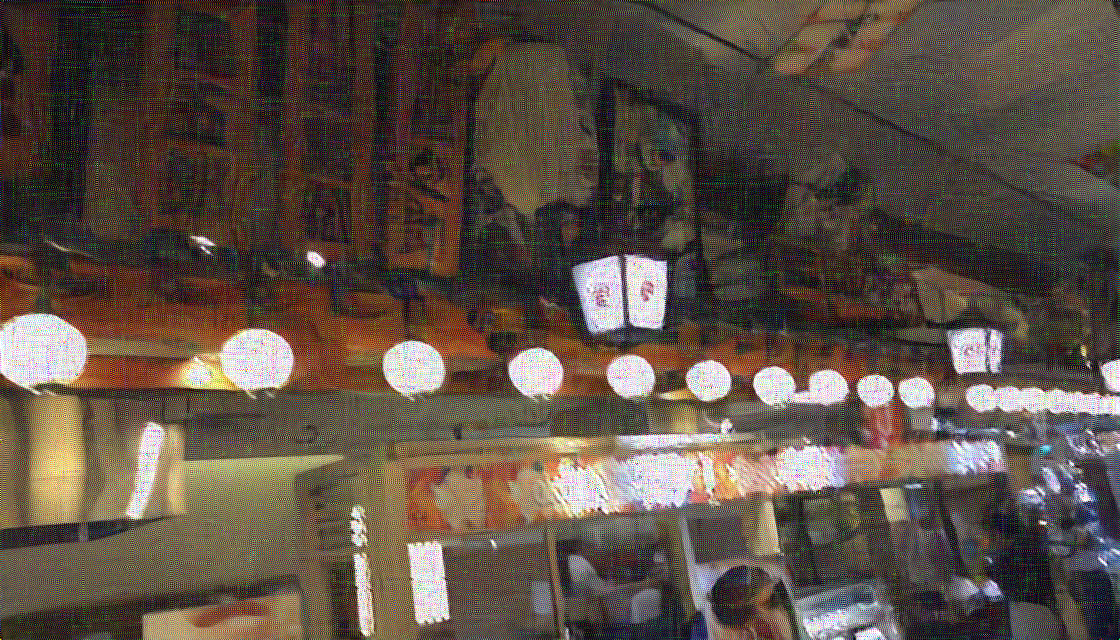} &
   \repfigvx{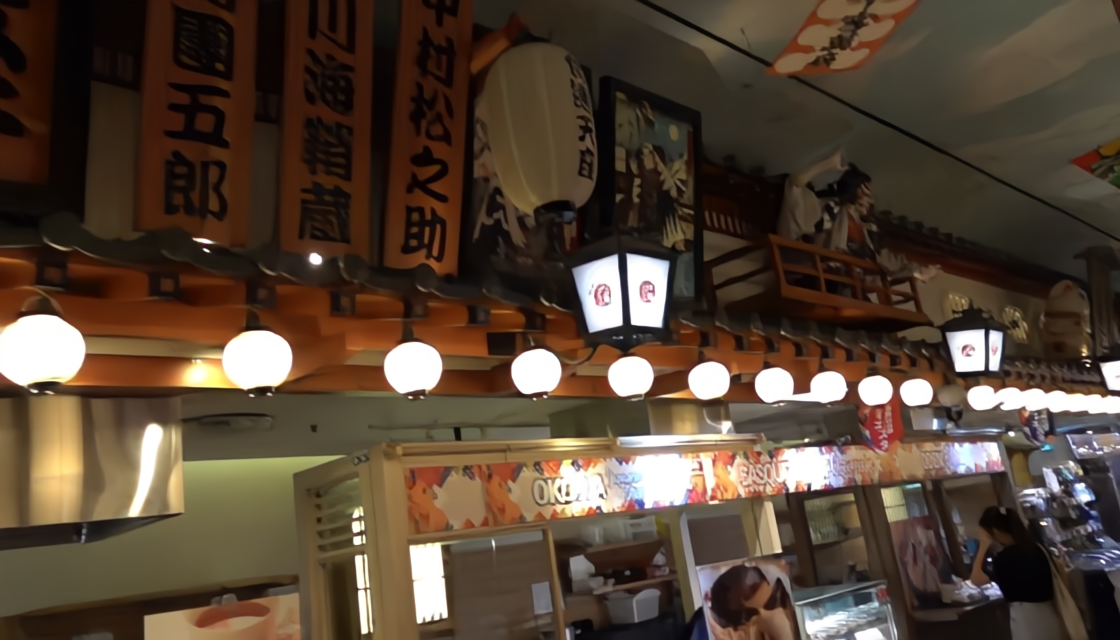} \\
   \repfigvx{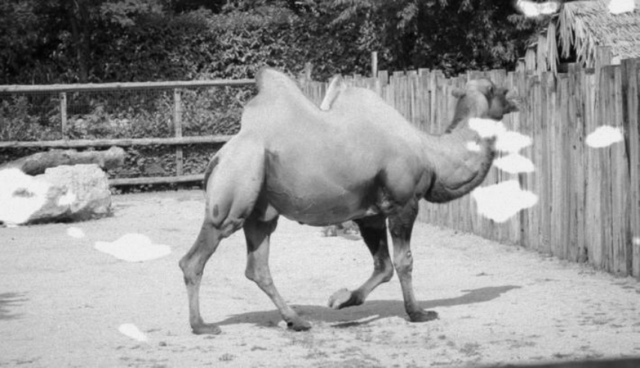} &
   \repfigvx{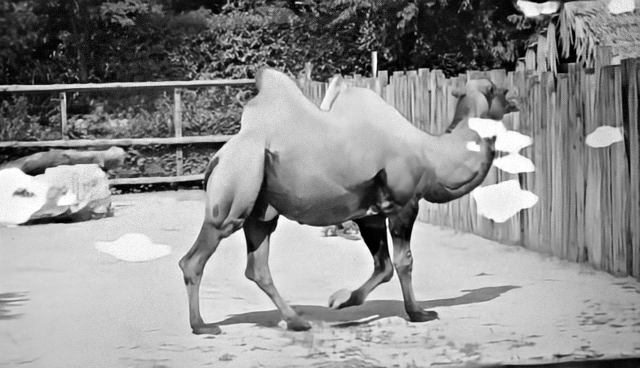} &
   \repfigvx{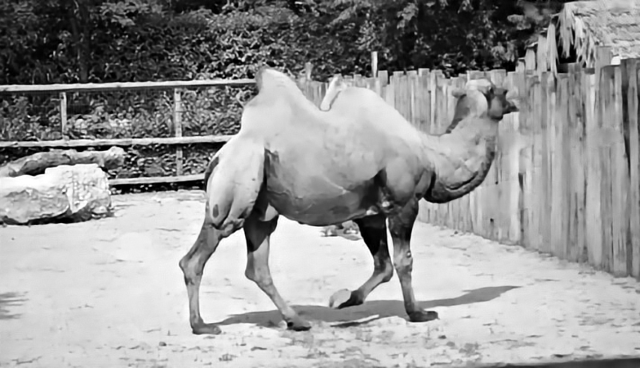} \\
   Input & I2I-RFR $v$-pred & I2I-RFR \\
   \end{tabular}
   \setlength{\tabcolsep}{6pt}
   \vspace{-3mm}
    \caption{Comparison between $v$-prediction and our direct target prediction in I2I-RFR for DarkIR~\cite{feijoo2025darkir} on low-light blur enhancement (top) and RRTN~\cite{lin2024restoring} on video restoration (bottom). The $v$-prediction variant produces severe artifacts, whereas direct target prediction yields more plausible results. \textbf{Zoom in for best view.}}
   \label{fig:comp_v_x_main}
   \vspace{-3mm}
\end{figure*}

\par\smallskip \smallskip
\noindent
\textbf{Parameterization Study.}
I2I-RFR is formulated as a direct prediction of the clean target image. For comparison, the same backbones can also be trained in a velocity-prediction form by regressing the rectified-flow velocity, since the backbone interface is unchanged up to the choice of training target.

The comparison between direct target prediction and $v$-prediction is conducted on low-light blur enhancement and video restoration. These tasks are representative ill-posed I2I settings in which perceptual fidelity and optimization stability are critical, and they jointly cover both image and video translation. The training data and optimization setup are kept identical for both variants; the only difference lies in the prediction target and the corresponding loss parameterization.
As shown in Table~\ref{tab:parameterization_results_main} and Fig.~\ref{fig:comp_v_x_main}, $v$-prediction leads to a substantial performance drop in most cases. A plausible explanation is that the evaluated backbones, including LEDNet and DarkIR, are carefully engineered for direct I2I mapping with task-specific priors and modules, e.g., frequency-domain processing. These inductive biases are more naturally exploited when predicting the clean target image than when regressing a velocity field.
Note that the LPIPS result for RRTN with an I2I-RFR $v$-prediction variant is the only case that improves, suggesting that the $v$-prediction form is not uniformly inferior to direct target prediction. Nevertheless, direct target prediction is generally more suitable for I2I backbones, as they are originally designed for direct image prediction.

\par\smallskip
\noindent
\textbf{Effect of time embeddings.}
Table~\ref{tab:time_embedding_results} presents a case study on low-light blur enhancement with DarkIR and on video restoration with RRTN, in which we add sinusoidal time embeddings to the intermediate features, as is commonly done in diffusion and rectified-flow models. Time embeddings degrade performance across all metrics in these settings. A likely reason is that I2I-RFR already exposes the corruption level through the concatenated input $[x; y_t]$, making explicit time conditioning redundant and, in some cases, less stable. We therefore omit time embeddings by default.

\begin{table*}[t]
\centering
\caption{Effect of time embeddings.}
\vspace{-3mm}
\label{tab:time_embedding_results}
\setlength{\tabcolsep}{16pt}
\resizebox{\textwidth}{!}{
    \begin{tabular}{l|ccc}
        \toprule
        & PSNR$\uparrow$ & SSIM$\uparrow$ & LPIPS$\downarrow$ \\
        \midrule
        \multicolumn{4}{l}{Low-light blur enhancement results on LOLBlur test set.} \\
        \midrule
        DarkIR + I2I-RFR & \textcolor{red}{27.19} & \textcolor{red}{0.897} & \textcolor{red}{0.106} \\
        DarkIR + I2I-RFR + time embedding & 26.80 & 0.877 & 0.127 \\
        \midrule
        \multicolumn{4}{l}{Old film restoration results on DAVIS test set.} \\
        \midrule
        RRTN + I2I-RFR & \textcolor{red}{25.87} & \textcolor{red}{0.8659} & \textcolor{red}{0.1699} \\
        RRTN + I2I-RFR + time embedding & 23.18 & 0.7922 & 0.2530 \\
        \bottomrule
    \end{tabular}
}
\vspace{-3mm}
\end{table*}

\section{Conclusion}
This work introduced \textit{Image-to-Image Rectified Flow Reformulation} (I2I-RFR), a practical plug-in reformulation that equips supervised image-to-image and video-to-video regression models with few-step ODE-based refinement while largely preserving the original backbone interface. By feeding a noise-corrupted target state alongside the input and training with a simple $t$-reweighted pixel loss, I2I-RFR reuses the inductive biases of task-specialized backbones and improves perceptual quality and detail preservation across diverse benchmarks in a practical few-step inference regime. Future work includes extending the evaluation to broader domains and tasks, such as illustration and other non-photographic data, and further studying how the reformulation interacts with task-specific modules, composite loss designs, and inference depth.

\bibliographystyle{splncs04}
\bibliography{references}

\newpage
\appendix

\section{Computational Overhead and Runtime}
\label{sec:runtime}
Table~\ref{tab:runtime} summarizes the parameter count, inference time, and peak memory usage of the original backbones and their I2I-RFR variants. These measurements were obtained by averaging over up to the first 100 evaluation samples of each benchmark on a system with an NVIDIA A6000 GPU and two AMD EPYC 7543 CPUs, using a batch size of 1. For the generic U-Net and Palette baselines, we report measurements at $1120\times640$ as a representative RGB I2I resolution. The parameter overhead of I2I-RFR is minimal because the backbone is reused almost unchanged, and the only architectural modification is the expansion of the input layer to accept the noisy target state. For regression backbones, the runtime increase mainly stems from iterative inference: with the default choice of $N=3$, I2I-RFR requires three forward passes instead of one. 
For diffusion-based baselines, however, runtime is substantially reduced because the original multi-step denoising process is replaced by only a few explicit Euler steps. Overall, the memory overhead is generally modest, while the measured latency remains practical relative to the substantially higher cost of diffusion-style sampling.

\begin{table*}[h]
\centering
\caption{\textbf{Computational overhead and runtime.}
Runtime is measured on the same GPU with a batch size of 1. All I2I-RFR variants use $N=3$ inference steps. For the generic U-Net and Palette baselines, we report measurements at
$1120\times640$ as a representative RGB I2I resolution.
For each backbone, we report the transition from the original model to its I2I-RFR counterpart.}
\label{tab:runtime}
\vspace{-3mm}
\setlength{\tabcolsep}{7pt}
\renewcommand{\arraystretch}{1.12}
\resizebox{\textwidth}{!}{
\begin{tabular}{lllccc}
\toprule
Task & Image Size & Backbone & Params (M) & Time (s) & VRAM (GB) \\
\midrule

Generic RGB I2I & 1120$\times$640 & U-Net
& \makecell[l]{31.0319 $\rightarrow$ 31.0336 \\ \scriptsize($+0.0017$)}
& \makecell[l]{0.0375 $\rightarrow$ 0.0775 \\ \scriptsize($+0.0400$)}
& \makecell[l]{1.32 $\rightarrow$ 1.58 \\ \scriptsize($+0.26$)} \\

& & \makecell[l]{Palette \\ (Diffusion)}
& \makecell[l]{62.6413 $\rightarrow$ 62.6413 \\ \scriptsize($+0.0000$)}
& \makecell[l]{13.5106 $\rightarrow$ 0.8149 \\ \scriptsize($-12.6957$)}
& \makecell[l]{20.93 $\rightarrow$ 20.93 \\ \scriptsize($+0.00$)} \\
\midrule

Super-Res & 480$\times$320 & NLSN
& \makecell[l]{44.1565 $\rightarrow$ 44.1634 \\ \scriptsize($+0.0069$)}
& \makecell[l]{0.0516 $\rightarrow$ 0.1194 \\ \scriptsize($+0.0678$)}
& \makecell[l]{1.39 $\rightarrow$ 1.39 \\ \scriptsize($+0.00$)} \\

& & SwinIR
& \makecell[l]{11.8998 $\rightarrow$ 11.9047 \\ \scriptsize($+0.0049$)}
& \makecell[l]{0.0785 $\rightarrow$ 0.1467 \\ \scriptsize($+0.0682$)}
& \makecell[l]{0.64 $\rightarrow$ 0.62 \\ \scriptsize($-0.02$)} \\
\midrule

Deblur & 670$\times$763 & NAFNet
& \makecell[l]{67.8888 $\rightarrow$ 67.8906 \\ \scriptsize($+0.0017$)}
& \makecell[l]{0.1170 $\rightarrow$ 0.3027 \\ \scriptsize($+0.1857$)}
& \makecell[l]{2.11 $\rightarrow$ 2.25 \\ \scriptsize($+0.14$)} \\

& & Restormer
& \makecell[l]{26.1266 $\rightarrow$ 26.1279 \\ \scriptsize($+0.0013$)}
& \makecell[l]{0.3543 $\rightarrow$ 1.0436 \\ \scriptsize($+0.6893$)}
& \makecell[l]{4.37 $\rightarrow$ 4.39 \\ \scriptsize($+0.02$)} \\
\midrule

Low-light & 1120$\times$640 & LEDNet
& \makecell[l]{7.4089 $\rightarrow$ 7.4098 \\ \scriptsize($+0.0009$)}
& \makecell[l]{0.0683 $\rightarrow$ 0.1928 \\ \scriptsize($+0.1245$)}
& \makecell[l]{3.21 $\rightarrow$ 3.25 \\ \scriptsize($+0.04$)} \\

& & DarkIR
& \makecell[l]{12.9573 $\rightarrow$ 12.9590 \\ \scriptsize($+0.0017$)}
& \makecell[l]{0.1723 $\rightarrow$ 0.4722 \\ \scriptsize($+0.2999$)}
& \makecell[l]{2.37 $\rightarrow$ 2.40 \\ \scriptsize($+0.03$)} \\
\midrule

Underwater & 256$\times$256 & \makecell[l]{Tang et al. \\ (Diffusion)}
& \makecell[l]{10.7056 $\rightarrow$ 10.7056 \\ \scriptsize($+0.0000$)}
& \makecell[l]{0.1703 $\rightarrow$ 0.0603 \\ \scriptsize($-0.1100$)}
& \makecell[l]{0.87 $\rightarrow$ 0.87 \\ \scriptsize($+0.00$)} \\

& & SFGNet
& \makecell[l]{1.2977 $\rightarrow$ 1.2979 \\ \scriptsize($+0.0002$)}
& \makecell[l]{0.0329 $\rightarrow$ 0.0654 \\ \scriptsize($+0.0325$)}
& \makecell[l]{0.61 $\rightarrow$ 0.62 \\ \scriptsize($+0.01$)} \\
\midrule

Video-Res & 30$\times$688$\times$368 & RTN
& \makecell[l]{6.1779 $\rightarrow$ 6.1782 \\ \scriptsize($+0.0003$)}
& \makecell[l]{0.3721 $\rightarrow$ 0.8136 \\ \scriptsize($+0.4415$)}
& \makecell[l]{8.29 $\rightarrow$ 9.13 \\ \scriptsize($+0.84$)} \\

& & RRTN
& \makecell[l]{7.4933 $\rightarrow$ 7.4935 \\ \scriptsize($+0.0001$)}
& \makecell[l]{0.6408 $\rightarrow$ 1.3750 \\ \scriptsize($+0.7342$)}
& \makecell[l]{9.40 $\rightarrow$ 9.47 \\ \scriptsize($+0.07$)} \\
\bottomrule
\end{tabular}
}
\end{table*}

\begin{table}[t]
\centering
\caption{\textbf{Ablation study on the number of inference steps ($N$).}
All models are evaluated using the proposed I2I-RFR with the $\ell_1$ loss.
The results show that the optimal inference depth is backbone-dependent: increasing $N$ often improves LPIPS, while PSNR and SSIM may saturate or decrease.
The best and second-best values are highlighted in \textcolor{red}{red} and \textcolor{blue}{blue}, respectively.}
\label{tab:inference_steps}
\vspace{-3mm}
\setlength{\tabcolsep}{12pt}
\resizebox{\textwidth}{!}{
    \begin{tabular}{ll|ccccc}
        \toprule
        & & \multicolumn{5}{c}{Number of Inference Steps ($N$)} \\
        \cmidrule{3-7}
        Model & Metric & 1 & 3 & 5 & 7 & 10 \\
        \midrule
        \multicolumn{7}{l}{Image super-resolution results on Set5 benchmark.} \\
        \midrule
        SwinIR~\cite{liang2021swinir}
        & PSNR$\uparrow$    & \textcolor{red}{32.85} & \textcolor{blue}{32.84} & 32.78 & 32.63 & 32.58 \\
        & SSIM$\uparrow$    & \textcolor{red}{0.9028} & \textcolor{red}{0.9028} & \textcolor{blue}{0.9019} & 0.9007 & 0.8994 \\
        & LPIPS$\downarrow$ & 0.1657 & 0.1665 & \textcolor{blue}{0.1640} & 0.1646 & \textcolor{red}{0.1607} \\
        \midrule
        \multicolumn{7}{l}{Image deblurring results on RealBlur-J benchmark.} \\
        \midrule
        Restormer~\cite{zamir2022restormer} 
        & PSNR$\uparrow$    & \textcolor{red}{29.04} & \textcolor{red}{29.04} & \textcolor{blue}{29.01} & 28.99 & 28.96 \\
        & SSIM$\uparrow$    & 0.8805 & \textcolor{red}{0.8812} & \textcolor{blue}{0.8808} & 0.8802 & 0.8792 \\
        & LPIPS$\downarrow$ & 0.1422 & 0.1408 & 0.1400 & \textcolor{blue}{0.1394} & \textcolor{red}{0.1387} \\
        \midrule
        \multicolumn{7}{l}{Low-light blur enhancement results on LOLBlur test set.} \\
        \midrule
        DarkIR~\cite{feijoo2025darkir} 
        & PSNR$\uparrow$    & \textcolor{red}{27.35} & \textcolor{blue}{27.19} & 27.05 & 26.96 & 26.86 \\
        & SSIM$\uparrow$    & \textcolor{red}{0.901} & \textcolor{blue}{0.897} & 0.892 & 0.889 & 0.884 \\
        & LPIPS$\downarrow$ & 0.108 & 0.106 & 0.105 & \textcolor{blue}{0.104} & \textcolor{red}{0.103} \\
        \midrule
        \multicolumn{7}{l}{Underwater image enhancement results on LSUI test set.} \\
        \midrule
        Tang et al.~\cite{tang2023underwater}
        & PSNR$\uparrow$    & \textcolor{blue}{29.02} & \textcolor{red}{29.27} & 28.78 & 28.89 & 28.59 \\
        & SSIM$\uparrow$    & \textcolor{blue}{0.9356} & \textcolor{red}{0.9381} & 0.9353 & 0.9335 & 0.9295 \\
        & LPIPS$\downarrow$ & 0.0865 & 0.0838 & 0.0834 & \textcolor{blue}{0.0799} & \textcolor{red}{0.0789} \\
        \midrule
        \multicolumn{7}{l}{Video restoration results on DAVIS test set.} \\
        \midrule
        RRTN~\cite{lin2024restoring}
        & PSNR$\uparrow$    & \textcolor{blue}{25.85} & \textcolor{red}{25.87} & 25.57 & 25.39 & 25.08 \\
        & SSIM$\uparrow$    & \textcolor{red}{0.8682} & \textcolor{blue}{0.8659} & 0.8574 & 0.8503 & 0.8424 \\
        & LPIPS$\downarrow$ & 0.1888 & \textcolor{red}{0.1699} & \textcolor{blue}{0.1723} & 0.1761 & 0.1815 \\
        \bottomrule
    \end{tabular}
}
\end{table}

\section{Effect of the Number of Inference Steps}
\label{sec:steps}
As mentioned in the main paper, I2I-RFR can be used with a small number of inference steps, and we use $N=3$ by default in all main experiments.
To study the role of inference depth, we vary the number of explicit Euler steps $N$ at test time while keeping the same trained checkpoint for each backbone.
Thus, the results at $N=3$ are identical to those reported in the main paper.

Table~\ref{tab:inference_steps} shows that the optimal step count is backbone-dependent, but a clear perception--distortion trade-off~\cite{blau2018perception} appears in many cases: larger $N$ often improves LPIPS, while PSNR and SSIM tend to saturate or decrease.
This suggests that the first few steps already capture most of the distortion-relevant information, and additional rollout mainly acts as perceptual refinement.

Importantly, the $N=1$ case is not equivalent to ordinary one-shot $\ell_1$ regression.
Even with a single inference step, the model is still trained under the I2I-RFR formulation with noisy target states and the $t$-reweighted objective.
Therefore, $N=1$ should be understood as the shallowest transport-style inference regime of I2I-RFR, rather than as the original regression baseline.
This also helps explain why the $N=1$ regime can already be competitive with, or even outperform, the corresponding standard regression baselines reported in the main paper.

Although $N=3$ is not universally optimal for every backbone and every metric, it provides a strong practical compromise between speed, distortion preservation, and perceptual refinement.
Qualitatively, as shown in Fig.~\ref{fig:vis_steps}, we also observe that the dominant visual improvement is typically achieved within the first few steps, and the differences beyond $N=3$ are often minor to the human eye.
We therefore use $N=3$ as the default operating point in the paper.

These observations also highlight a practical difference from diffusion-based methods such as DDIM~\cite{song2020ddim}, which often require substantially more denoising steps to achieve strong performance. In contrast, I2I-RFR learns refinement trajectories that already work well in a few-step regime, which is important for practical image and video restoration applications.

\newcommand{\ifigzoom}[2]{%
  \includegraphics[
    width=0.16\linewidth,
    trim=#1,clip
  ]{figs/results/Comp_step/#2}%
}
\newcommand{\ifig}[1]{\includegraphics[width=0.16\linewidth]{figs/results/Comp_step/#1}}
\begin{figure}[t]
   \centering
   \setlength{\tabcolsep}{1pt}
   \begin{tabular}{cccccc}
   \ifig{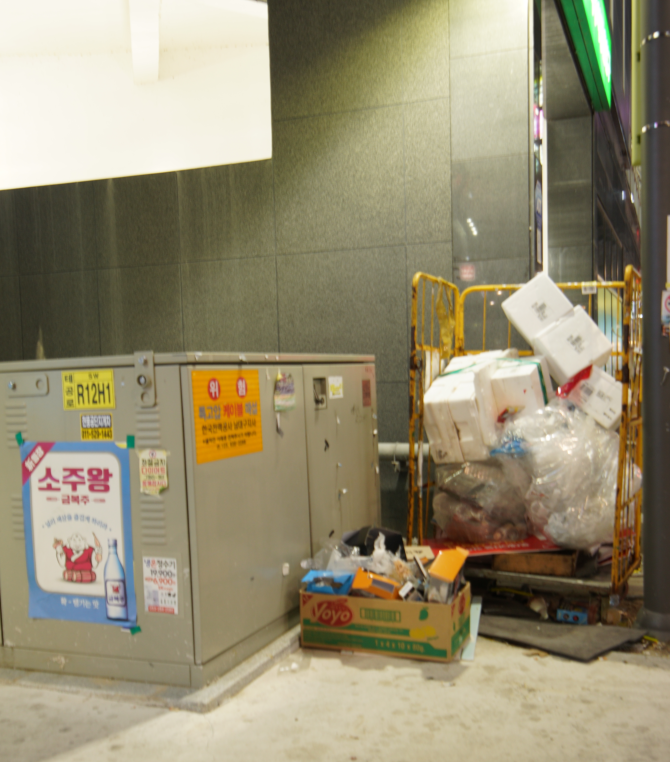} &
   \ifig{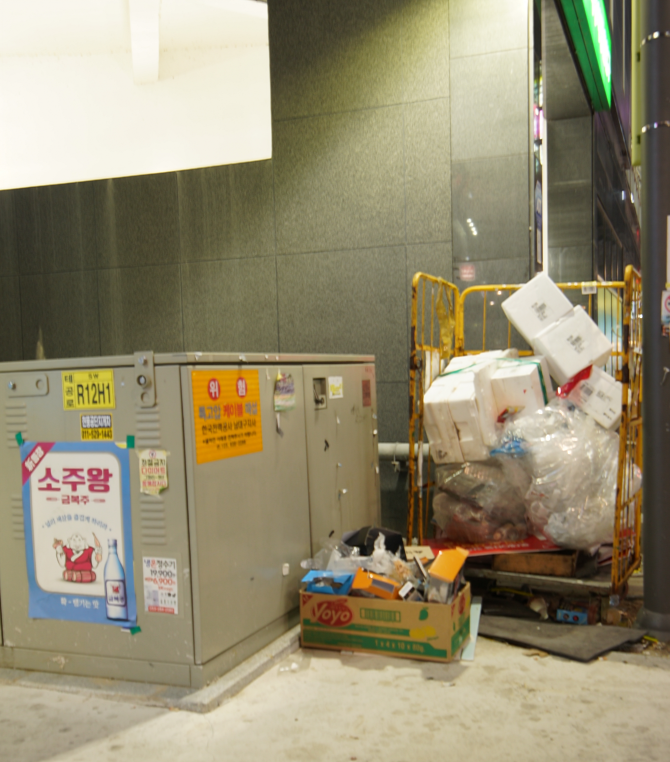} &
   \ifig{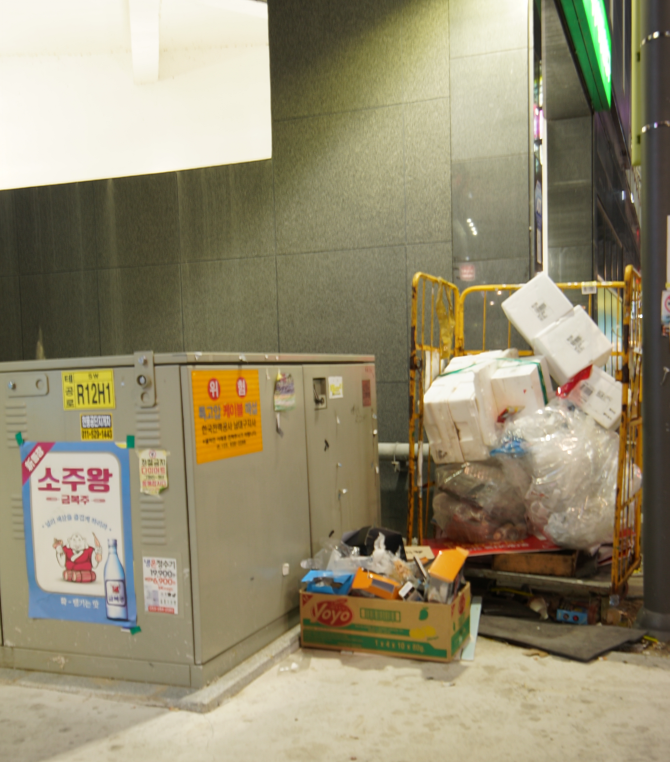} &
   \ifig{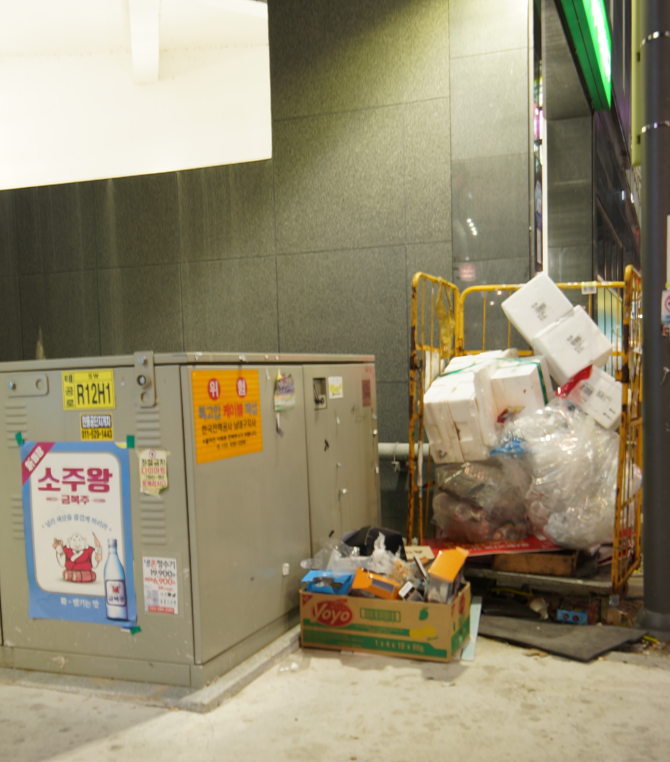} &   
   \ifig{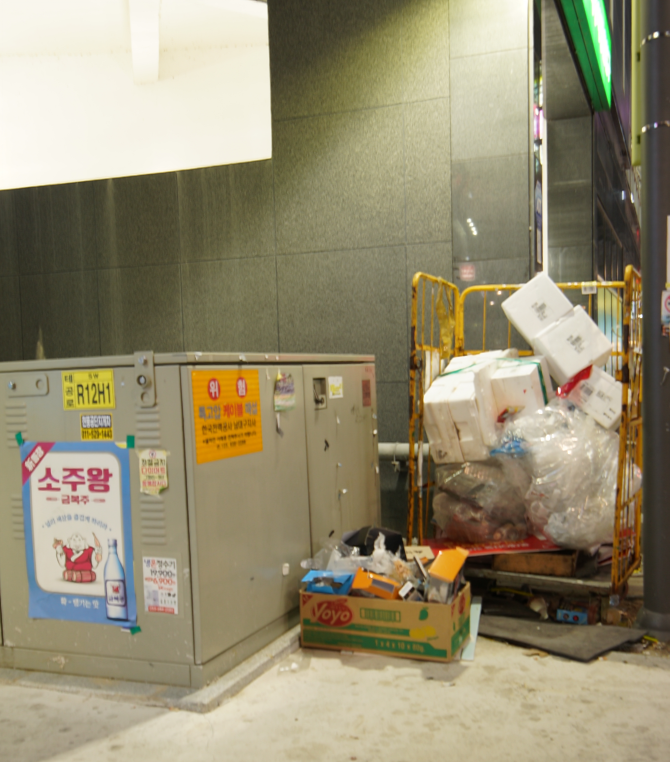} &
   \ifig{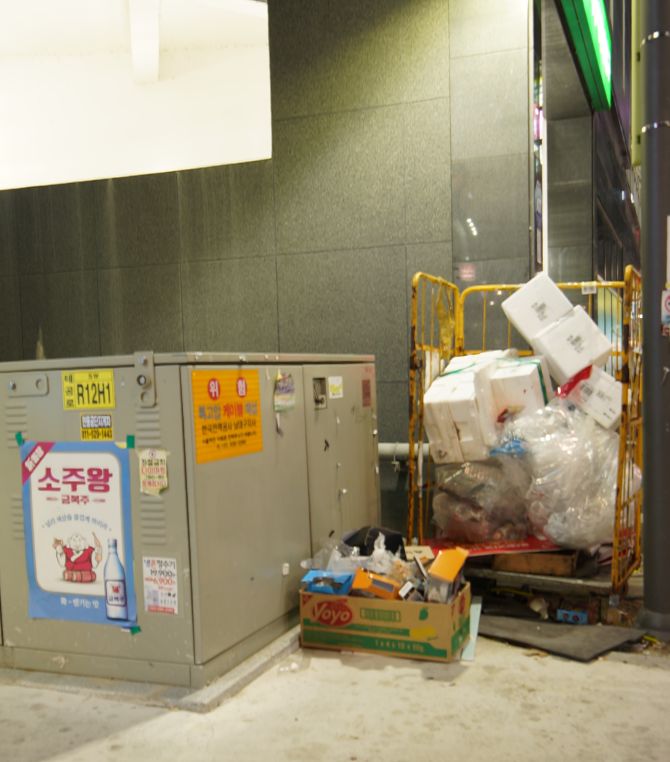} \\
   \ifigzoom{85mm 0mm 85mm 0mm}{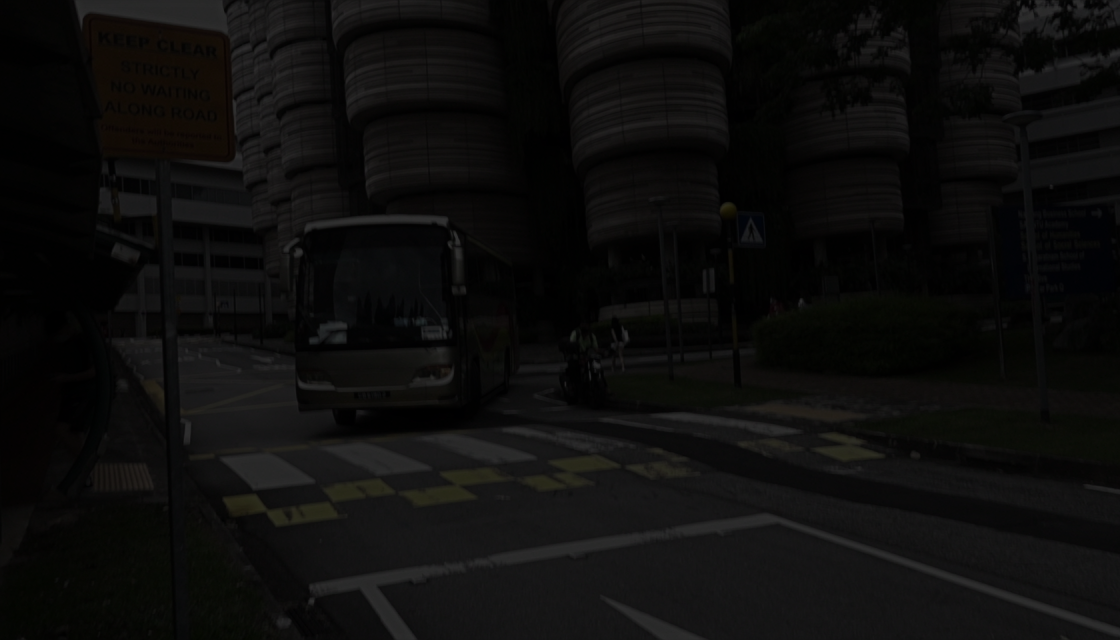} &
   \ifigzoom{85mm 0mm 85mm 0mm}{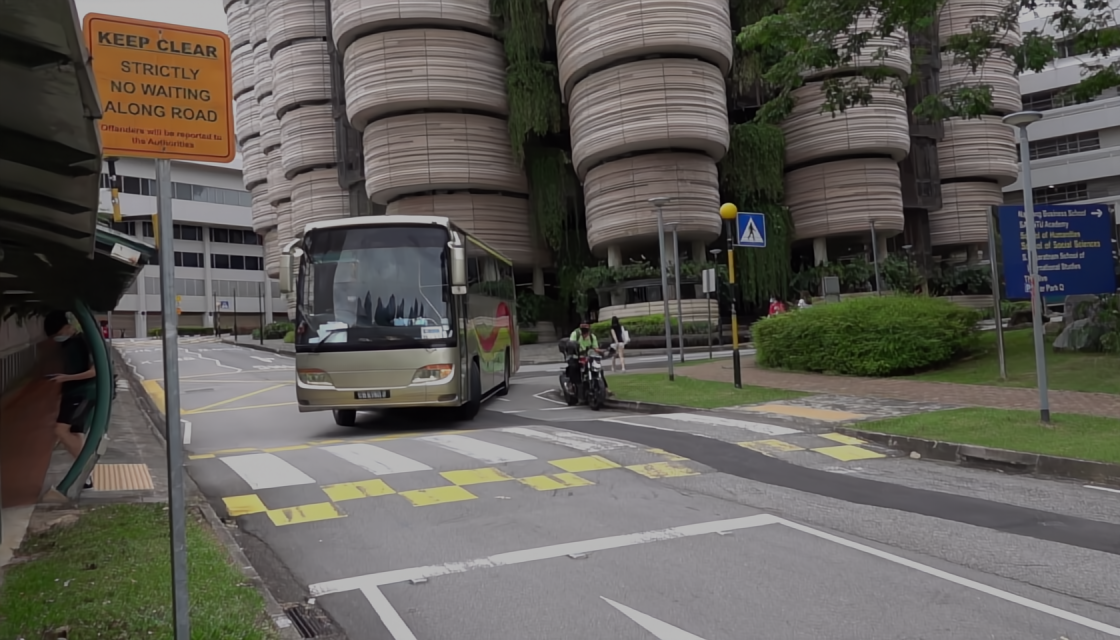} &
   \ifigzoom{85mm 0mm 85mm 0mm}{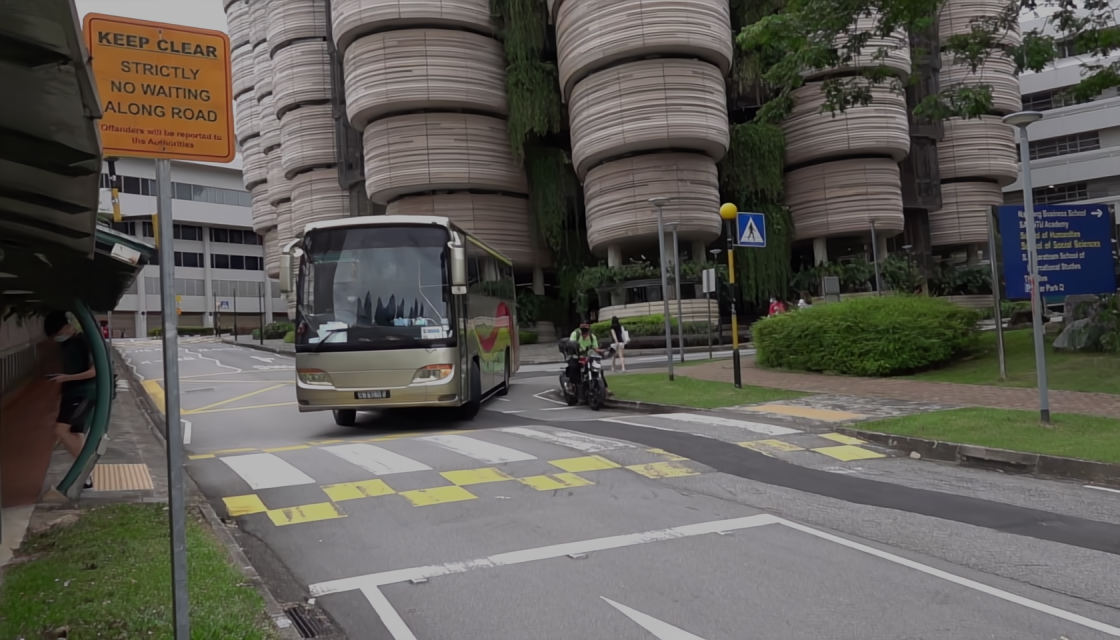} &
   \ifigzoom{85mm 0mm 85mm 0mm}{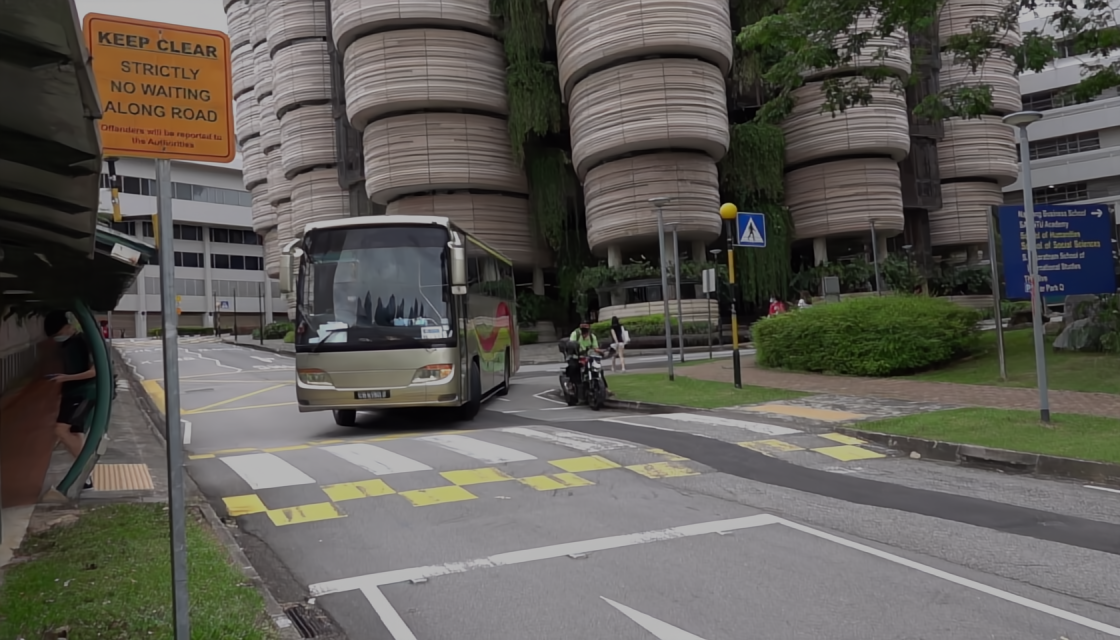} &   
   \ifigzoom{85mm 0mm 85mm 0mm}{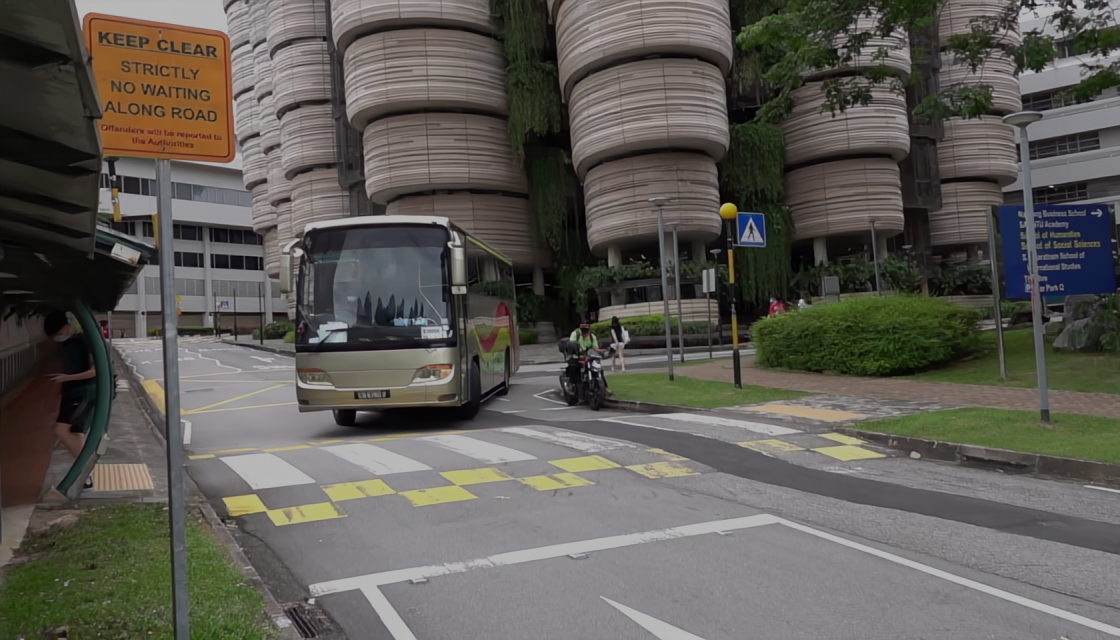} &
   \ifigzoom{85mm 0mm 85mm 0mm}{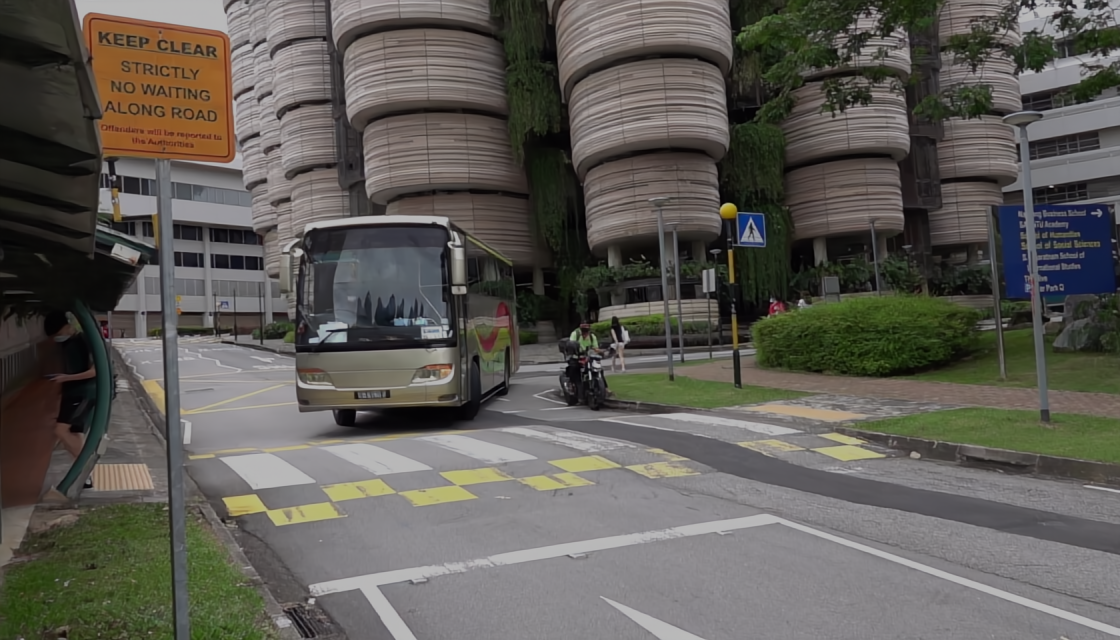} \\
   \ifig{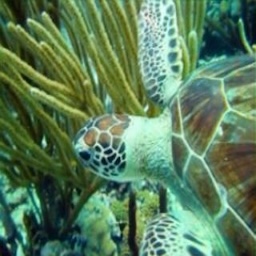} &
   \ifig{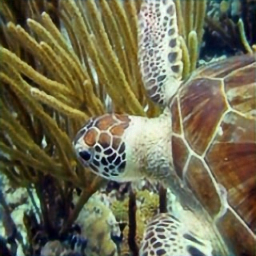} &
   \ifig{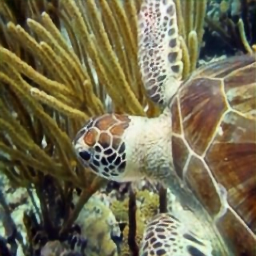} &
   \ifig{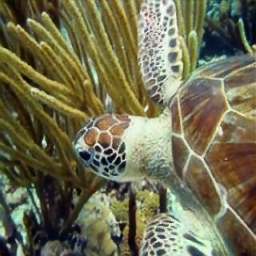} &   
   \ifig{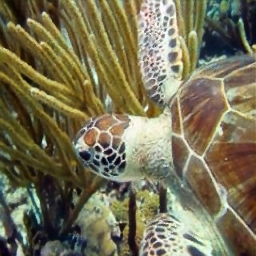} &
   \ifig{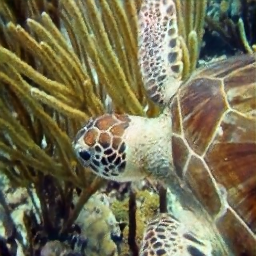} \\   
   Input & $N=1$ & $N=3$ & $N=5$ & $N=7$ & $N=10$
   \end{tabular}
   \setlength{\tabcolsep}{6pt}
   \vspace{-3mm}
    \caption{\textbf{Qualitative effect of the number of inference steps.}
    We compare representative outputs generated with different numbers of explicit Euler steps.
    From top to bottom, the three rows correspond to image deblurring, low-light enhancement, and underwater image enhancement, respectively.
    While larger step counts can continue to improve LPIPS in some cases, the dominant visual refinement is typically achieved within the first few steps, and the differences beyond $N=3$ are often minor to the human eye.}
   \label{fig:vis_steps}
   \vspace{-3mm}
\end{figure}

\section{Full Implementation Details}
\label{sec:impl}
We provide the implementation details of the I2I-RFR variants of the baseline models used in the main paper, which could not be fully included there due to space constraints.
Unless otherwise specified, all models are optimized with Adam using an initial learning rate of $10^{-4}$ and a cosine learning-rate schedule down to $10^{-6}$. For I2I-RFR, $t$ is sampled as $t\sim\mathrm{Beta}(2,1)$ with $t_{\min}=10^{-3}$, and the default loss is the $t^{-1}$-reweighted $\ell_1$ distance. At inference time, the induced ODE is solved with a fixed-step explicit Euler integrator using $N=3$ steps for all tasks unless stated otherwise. 
Classifier-free guidance and explicit time embedding modules are not used.
All images are normalized to [0,1] before being mixed with Gaussian noise, and
the final predictions are clipped back to the valid image range for evaluation.
For quantitative evaluation, we use a fixed random seed and generate a single
sample per input.
For all models, we use the checkpoints with the lowest validation loss as the final models.
The final checkpoints were selected at training iterations ranging from 200K to 4M, depending on the model.
Training was performed on NVIDIA RTX 3090, RTX A6000, and H100 GPUs, depending on availability. Because video restoration models were more computationally demanding, we used H100 GPUs for them whenever possible.
For vanilla Palette~\cite{saharia2022palette}, we use 50 DDIM~\cite{song2020ddim} steps for inference and adopt the default noise schedule for all tasks during training and inference, namely a linear schedule with $\beta_{\min}=10^{-6}$ and $\beta_{\max}=10^{-2}$ over 2000 diffusion timesteps.
We use AlexNet features to compute LPIPS.
Further details for each task are as follows:
\begin{itemize}
    \item \textbf{Image Super-resolution.} 
    We train all models on the DF2K dataset~\cite{agustsson2017ntire}, which consists of 3,450 training images, with batch sizes ranging from 8 to 32 depending on GPU memory constraints.
    During training, HR patches are randomly cropped to $256\times256$ pixels, and the corresponding LR inputs are generated by $\times4$ bicubic downsampling. 
    We apply random horizontal and vertical flips and random 90-degree rotations for data augmentation.
    For I2I-RFR, the conditioning state is constructed in the LR domain to match the model input shape: the network is fed with $[x^{\mathrm{LR}}; y_t^{\mathrm{LR}}]$, while the prediction target and the loss are defined in the HR domain.
    We evaluate the models on the standard benchmarks Set5, Set14, BSD100, and Urban100 at their original resolutions.
    In terms of evaluation metrics, we report PSNR and SSIM on the Y channel of the YCbCr color space as is standard practice.
    \item \textbf{Image Deblurring.}
    We train all models on the GoPro dataset~\cite{Nah_2017_CVPR}, which consists of 2,103 training image pairs, with batch sizes ranging from 8 to 32 depending on GPU memory constraints.
    During training, input images are randomly cropped to $256\times256$ pixels, and we apply random horizontal and vertical flips for data augmentation.
    We evaluate the models on the RealBlur-J test set~\cite{RealBlur}, which consists of 980 image pairs at their original resolutions.
    \item \textbf{Low-Light Blur Enhancement.}
    We train all models on the LOLBlur dataset~\cite{zhou2022lednet} which consists of 10,200 training image pairs, with batch sizes ranging from 8 to 32 depending on GPU memory constraints.
    During training, input images are randomly cropped to $256\times256$ pixels, and we apply random horizontal flips and random 90-degree rotations for data augmentation.
    We use the pretrained checkpoints of LEDNet and DarkIR provided by the original authors as the baselines, as retraining these models ourselves did not yield performance comparable to that of the released checkpoints.
    The corresponding I2I-RFR variants are trained from scratch on the LOLBlur training split unless stated otherwise.
    We evaluate the models on the LOLBlur test set~\cite{zhou2022lednet}, which consists of 1,800 image pairs at their original resolutions.
    \item \textbf{Underwater Image Enhancement.}
    We train all models on the LSUI benchmark~\cite{peng2023ushape}, which consists of 3,879 training image pairs in the updated version, with batch sizes ranging from 8 to 16 depending on GPU memory constraints.
    Following standard practice~\cite{tang2023underwater}, training and evaluation are performed on 256$\times$256 images: during training, image patches are randomly cropped from the original images, whereas during evaluation, the images are resized to 256$\times$256 pixels.
    We apply random horizontal flips and random 90-degree rotations for data augmentation.
    We evaluate the models on the LSUI test set~\cite{peng2023ushape}, which consists of 400 image pairs, and additionally report results on the UIEB test set~\cite{li2019underwater}, which contains 90 image pairs, to assess generalization beyond the synthetic LSUI distribution.
    \item \textbf{Video Restoration.}
    We train all models on the REDS dataset~\cite{Nah_2019_CVPR_Workshops}, which consists of 240 training video sequences, each containing 100 frames, with a batch size of 4.
    Following RRTN~\cite{lin2024restoring}, the synthetic degradation pipeline is applied to the original videos on-the-fly during training, 
    which replicates the old-film degradation process with grayscale conversion, random cropping to $128\times128$ pixels, random horizontal and vertical flips, and random 90-degree rotations.
    We use 7 consecutive frames as each input-output sequence pair.
    This task requires not only spatial modeling but also temporal modeling, and thus we do not provide U-Net and Palette baselines for this task.
    Following standard practice, we evaluate the models on the DAVIS test set~\cite{ponttuset2017davis} which consists of 30 video sequences. 
    We apply the same synthetic degradation pipeline to the original videos in the test set for evaluation, and we resize the original videos in the test set such that the shorter side is 384 pixels. 
    During inference, we process 30 consecutive frames at a time. We use the first 15 frames as reference frames for restoring the next 15 frames, and then slide this window forward by 15 frames until the end of the video sequence, following RRTN~\cite{lin2024restoring}.
\end{itemize}

\section{Extended Ablations of the I2I-RFR Recipe}
\label{sec:recipe}
As discussed in the main paper, the default I2I-RFR recipe uses Beta-based $t$-sampling, no explicit time embeddings, and no adversarial training.
Moreover, the performance gain is not solely due to the rectified-flow-style reformulation itself, but also to the choice of parameterization, namely predicting the clean image $y$ rather than the velocity $v$. To further verify the role of each design choice, we provide additional analyses and ablation studies for the main components of the I2I-RFR recipe.

\subsection{Visualization of the Beta-based $t$-Sampling Strategy}

\begin{figure}[t]
    \centering
    \setlength{\tabcolsep}{2pt}
    \begin{tabular}{cc}
        \includegraphics[width=0.48\linewidth]{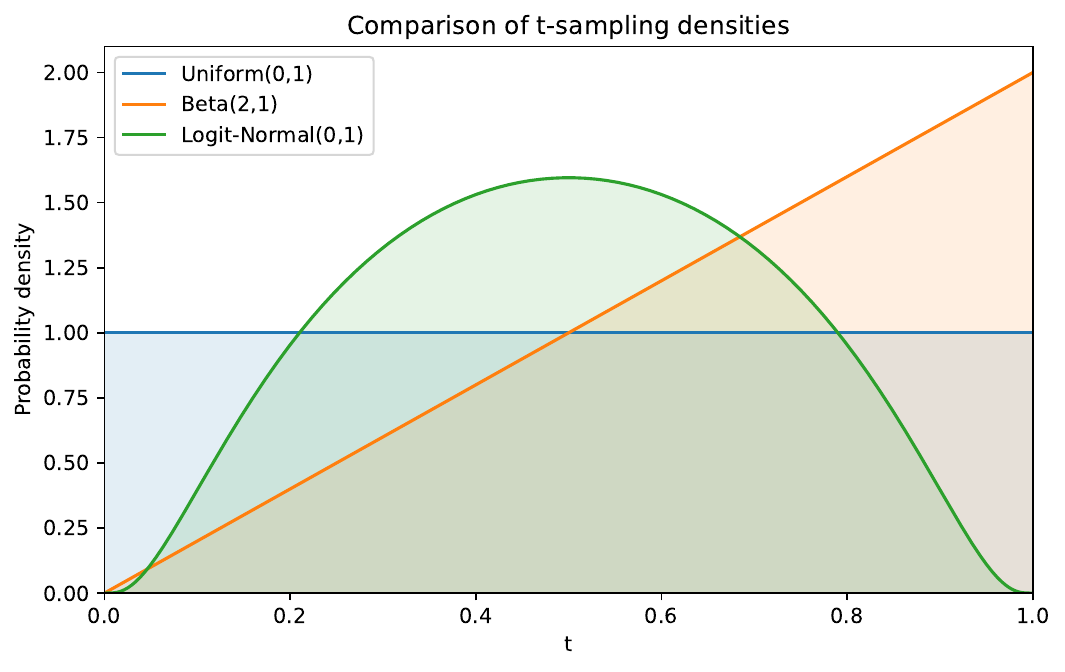} &
        \includegraphics[width=0.48\linewidth]{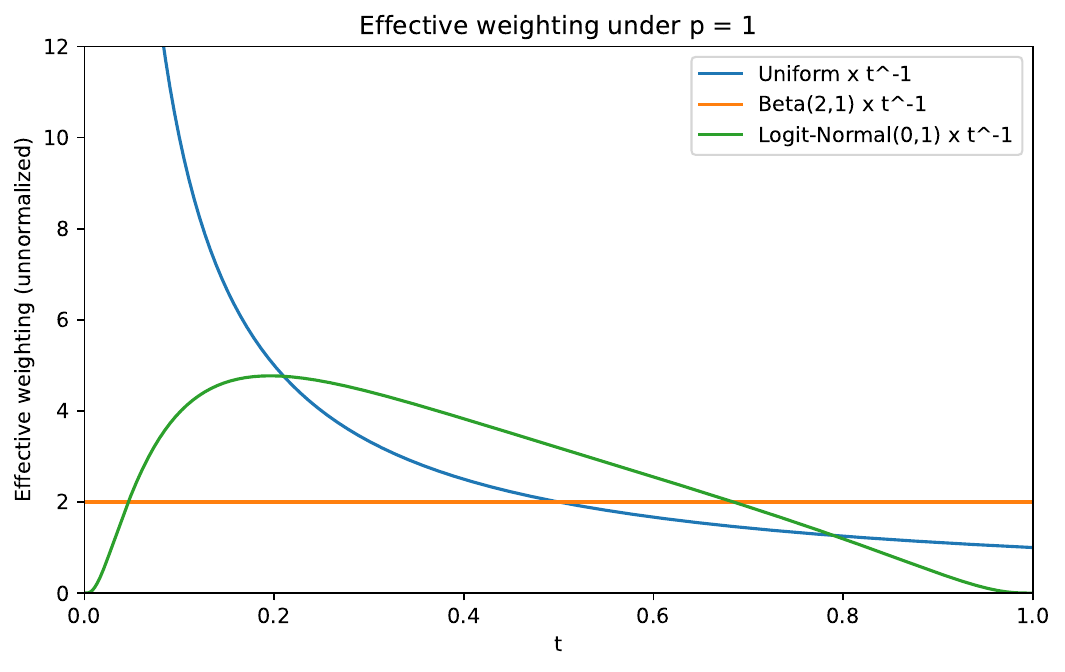} \\
        \small (a) Sampling densities &
        \small (b) Effective weighting under $p=1$
    \end{tabular}
    \caption{\textbf{Visualization of the proposed Beta-based $t$-sampling strategy.}
    (a) Comparison of the probability density functions used for sampling the interpolation parameter $t$, including uniform sampling, logit-normal$(0,1)$ sampling, and the proposed $\mathrm{Beta}(2,1)$ sampling used in the default $p=1$ setting of I2I-RFR. Compared with uniform sampling, $\mathrm{Beta}(2,1)$ places slightly more probability mass near the high-noise regime $t\approx 1$, while logit-normal$(0,1)$ concentrates more mass around intermediate noise levels.
    (b) Effective weighting induced by the training objective when $p=1$. With uniform sampling, the explicit $t^{-1}$ loss reweighting overemphasizes small $t$. In contrast, $\mathrm{Beta}(2,1)$ has density proportional to $t$, which approximately counterbalances the loss reweighting and yields a flatter effective weighting over $t$.}
    \label{fig:supp_beta_sampling}
\end{figure}

To clarify the role of our default Beta-based sampling strategy, Fig.~\ref{fig:supp_beta_sampling} visualizes both the sampling densities and the effective weighting induced by the training objective.
In the default setting of I2I-RFR, we use $p=1$, which leads to $t \sim \mathrm{Beta}(2,1)$.
Since the reformulated loss contains an explicit $t^{-1}$ factor, sampling $t$ uniformly would overemphasize small-$t$ regions and can increase gradient variance.
In contrast, $\mathrm{Beta}(2,1)$ has density proportional to $t$, which approximately counterbalances the loss reweighting and produces a flatter effective weighting over $t$.
Compared with logit-normal$(0,1)$ sampling, it also allocates slightly more probability mass to the high-noise regime ($t\approx 1$), which can be beneficial for robust refinement.

At the same time, this does not imply that Beta sampling is universally optimal in all settings.
Different backbones or tasks may favor somewhat different sampling distributions, and alternatives such as logit-normal sampling may be competitive or even preferable depending on the desired balance across noise levels.
We therefore view $\mathrm{Beta}(2,1)$ primarily as a simple and effective default
choice rather than a universally optimal one.
This interpretation is also consistent with the empirical comparison in Fig.~\ref{fig:sampling_strategy} of the main paper, where Beta sampling converges more smoothly and yields the highest validation PSNR among the tested strategies in our default setup.

\begin{table*}[t]
\centering
\caption{\textbf{Additional parameterization study.}
The main paper compares direct target prediction and velocity prediction on low-light blur enhancement and video restoration.
Here, we further extend the same comparison to underwater image enhancement and image deblurring.
Best values for each task are highlighted in \textcolor{red}{red}.}
\label{tab:parameterization_results}
\vspace{-3mm}
\setlength{\tabcolsep}{18pt}
\resizebox{\textwidth}{!}{
    \begin{tabular}{l|ccc}
        \toprule
        & PSNR$\uparrow$ & SSIM$\uparrow$ & LPIPS$\downarrow$ \\
        \midrule
        \multicolumn{4}{l}{Underwater image enhancement results on the LSUI test set.} \\
        \midrule
        SFGNet + I2I-RFR & \textcolor{red}{26.09} & \textcolor{red}{0.9190} & \textcolor{red}{0.1164} \\
        SFGNet + I2I-RFR ($v$-pred.) & 23.76 & 0.8817 & 0.1557 \\
        \midrule
        \multicolumn{4}{l}{Image deblurring results on the RealBlur-J test set.} \\
        \midrule
        Restormer + I2I-RFR & \textcolor{red}{29.04} & \textcolor{red}{0.8812} & \textcolor{red}{0.1408} \\
        Restormer + I2I-RFR ($v$-pred.) & 28.52 & 0.8241 & 0.1483 \\
        \bottomrule
    \end{tabular}
}
\vspace{-1mm}
\end{table*}

\subsection{Additional Parameterization Study}
In the main paper, we already compared direct target prediction and velocity prediction on low-light blur enhancement and video restoration, and observed that direct target prediction is consistently more effective in those settings.
Here, we further extend the same comparison to two additional tasks: underwater image enhancement and image deblurring.
As in the main paper, the default I2I-RFR formulation predicts the clean target image directly.
For comparison, we also train the same backbones in a velocity-prediction form by regressing the rectified-flow velocity, while keeping the model interface unchanged apart from the choice of prediction target.
The velocity-prediction variants are trained with the same data and optimization setup as the corresponding original I2I-RFR models, changing only the prediction target and the associated loss parameterization.

As shown in Table~\ref{tab:parameterization_results}, the same trend remains clear in both additional tasks: the velocity-prediction variants lead to substantial performance degradation in both settings.
This further supports the conclusion of the main paper that the choice of parameterization is crucial for I2I-RFR.
A plausible explanation is that, in paired image-to-image problems, direct target prediction defines a more stable learning objective because the target image is strongly correlated with the conditioning input, whereas velocity prediction is more sensitive to the noise level and must preserve the residual noise information in the intermediate state.
Moreover, many image-to-image backbones are originally designed for direct image regression rather than vector-field prediction, so replacing the regression target with velocity prediction can be mismatched with their original inductive bias.
Overall, these additional results reinforce the conclusion of the main paper that direct target prediction is the more effective parameterization for I2I-RFR on paired image-to-image tasks, even though velocity prediction is more commonly adopted in standard rectified-flow image generation.

\subsection{Additional Study on a Bridge-Style Formulation}
Beyond the prediction target, we also examine an alternative path design for I2I-RFR.
Our default formulation follows the standard rectified-flow choice of defining a straight path between Gaussian noise and the clean target image.
As an alternative, one can instead define the path between the degraded input image and the clean target image, following a bridge-style formulation explored in several recent image-to-image and diffusion-related works~\cite{tan2025visionbridgetransformerscale,xu2025fast,li2023bbdm,bansal2023cold,yue2023resshift}.

In this alternative formulation, the model predicts the clean target image from intermediate states interpolated between the input image and the target image, without explicitly concatenating the input image as a separate conditioning signal.
We evaluate this design under the same training setup as the original I2I-RFR-CL formulation.
As shown in Table~\ref{tab:bridge_results}, this bridge-style variant performs substantially worse than the original I2I-RFR formulation in this setting.

A plausible explanation is twofold. First, more generally, the bridge-style path is less flexible for tasks whose input and output domains differ, because it assumes that the refinement state and the prediction target live in compatible spaces.
Second, in the present low-light restoration setting, the model no longer receives the original input image explicitly at every refinement step, making it harder to exploit the conditioning information, especially in the high-noise regime where the intermediate state is dominated by noise. These results suggest that direct access to the conditioning input is important for effective I2I-RFR refinement, and that the interpolation state alone is insufficient to preserve the useful structure of the degraded input.

\begin{table}[t]
\centering
\caption{\textbf{Effect of an alternative bridge-style path design.}
We compare the default I2I-RFR formulations with a bridge-style variant that predicts the clean image from interpolation states between the input image and the target image.
Values in parentheses indicate the change relative to the original DarkIR baseline.}
\label{tab:bridge_results}
\vspace{-3mm}
\setlength{\tabcolsep}{20pt}
\resizebox{\textwidth}{!}{
    \begin{tabular}{l|ccc}
        \toprule
        & PSNR$\uparrow$ & SSIM$\uparrow$ &  LPIPS$\downarrow$ \\
        \midrule
        DarkIR~\cite{feijoo2025darkir} & 27.31 & 0.896 & 0.111 \\
        DarkIR + I2I-RFR & \textcolor{blue}{27.19} (-0.12) & \textcolor{blue}{0.897} \textbf{(+0.001)} & \textcolor{blue}{0.106} \textbf{(-0.005)} \\
        DarkIR + I2I-RFR-CL & \textcolor{red}{27.72} \textbf{(+0.41)} & \textcolor{red}{0.900} \textbf{(+0.004)} & \textcolor{red}{0.103} \textbf{(-0.008)} \\
        DarkIR + I2I-RFR-CL (bridge path) & 23.33 (-3.98) & 0.766 (-0.13)	& 0.201 (+0.09) \\
        \bottomrule
    \end{tabular}
}
\end{table}

\begin{table*}[t]
\centering
\caption{\textbf{Comparative study of training objectives and formulation variants.}
All models are trained on the LOLBlur benchmark.
Best values are highlighted in \textcolor{red}{red}, and second-best values in \textcolor{blue}{blue}.}
\label{tab:learning_objective_results}
\vspace{-3mm}
\setlength{\tabcolsep}{20pt}
\resizebox{\textwidth}{!}{
    \begin{tabular}{lccc}
        \toprule
        & PSNR$\uparrow$ & SSIM$\uparrow$ & LPIPS$\downarrow$ \\
        \midrule
        $\ell_1$ & 24.35 & 0.8070 & 0.2427 \\
        $\ell_1$ + Perceptual & \textcolor{blue}{24.69} & \textcolor{blue}{0.8110} & 0.2413 \\
        $\ell_1$ + Perceptual + Adversarial & 23.63 & 0.7949 & 0.1942 \\
        \midrule
        DDPM & 8.76 & 0.0957 & 0.7601 \\
        \midrule
        I2I-RFR ($v$-pred.) + CFG & 13.92 & 0.4220 & 0.4304 \\
        I2I-RFR ($v$-pred.) & 19.55 & 0.7749 & \textcolor{blue}{0.1704} \\
        \midrule
        I2I-RFR + CFG & 12.92 & 0.3558 & 0.4713 \\
        I2I-RFR & \textcolor{red}{26.26} & \textcolor{red}{0.8593} & \textcolor{red}{0.1585} \\
        \bottomrule
    \end{tabular}
}
\vspace{-4mm}
\end{table*}

\subsection{Comparative Study of Training Objectives and Formulation Variants}
\label{sec:objective_formulation}

To further compare I2I-RFR with several commonly used alternatives for perceptual-quality-oriented image-to-image training, we conduct an additional controlled study on a representative low-light blur enhancement benchmark.
Specifically, we compare standard $\ell_1$ regression, $\ell_1$ with perceptual loss, $\ell_1$ with perceptual and adversarial losses, DDPM training, and I2I-RFR variants with different parameterizations and with/without classifier-free guidance (CFG)~\cite{ho2022classifier}.
We use the LOLBlur dataset~\cite{zhou2022lednet}, train all models with a batch size of 8, and select the checkpoints with the lowest validation loss as the final models.

For perceptual supervision, we use pretrained VGG-16 features.
For adversarial training, we adopt a pix2pix-style discriminator~\cite{IsolaCVPR2017}.
The perceptual and adversarial loss weights are set to 0.01 and 0.05, respectively.
When CFG is used, the conditioning input is dropped with probability 0.1 during training, and a guidance scale of 4 is used at inference.
For sampling, DDPM-based models use DDIM with 50 steps, whereas I2I-RFR-based models use Euler inference with 3 steps.

The quantitative and qualitative results are presented in Table~\ref{tab:learning_objective_results} and Fig.~\ref{fig:learning_objective_results}, respectively.
Among all variants considered here, the default I2I-RFR formulation, namely direct target prediction without CFG, achieves the best overall performance.
Adding perceptual or adversarial losses to standard $\ell_1$ regression can improve individual metrics, but the gains remain limited and the overall performance still falls short of I2I-RFR.
The DDPM baseline performs much worse, which may reflect that learning a stochastic diffusion trajectory is considerably harder in this paired low-light enhancement setting and can lead to severe artifacts such as color degradation.
This is consistent with the observation in the main paper that the original DDPM version of Palette performs much worse than its I2I-RFR counterpart.

We also find that CFG causes large degradation for both direct-prediction and velocity-prediction variants of I2I-RFR.
A plausible explanation is that unconditional dropout makes the task unnecessarily difficult in strongly image-conditioned restoration settings, weakening the model's ability to learn a stable mapping from the noise-corrupted target state to the clean target image.
Finally, consistent with the parameterization study, the velocity-prediction variant again performs much worse than direct target prediction, even when using the standard Palette-style backbone.
Overall, these results support the default I2I-RFR formulation as the strongest and most stable choice among the alternatives considered here.

\newcommand{\jfigzoom}[2]{%
  \includegraphics[
    width=0.19\linewidth,
    trim=#1,clip
  ]{figs/results/Comp_learn/#2}%
}
\newcommand{\jfig}[1]{%
  \includegraphics[width=0.18\linewidth]{figs/results/Comp_learn/#1}%
}
\begin{figure}[t]
   \centering
   \setlength{\tabcolsep}{1pt}

   \begin{tabular}{ccccc}
   \jfigzoom{85mm 0mm 85mm 0mm}{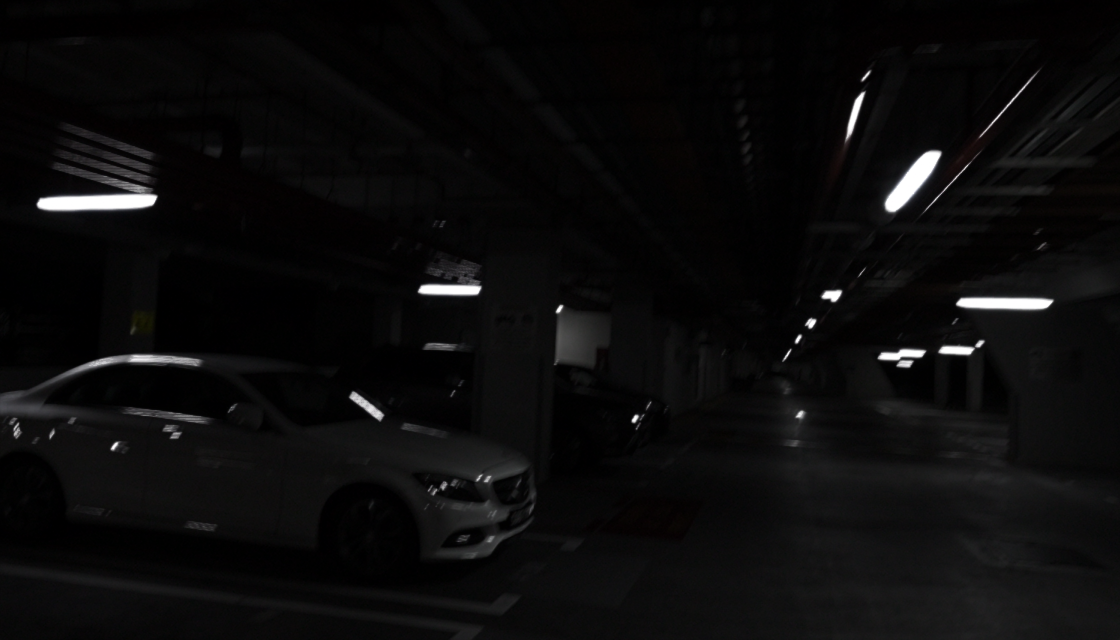} &
   \jfigzoom{85mm 0mm 85mm 0mm}{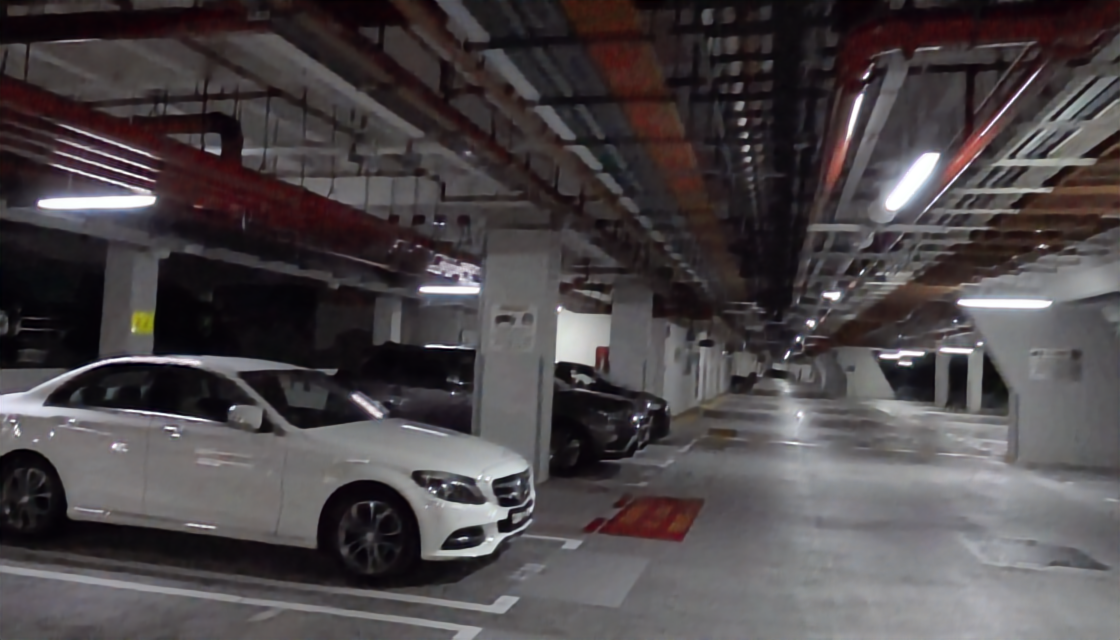} &
   \jfigzoom{85mm 0mm 85mm 0mm}{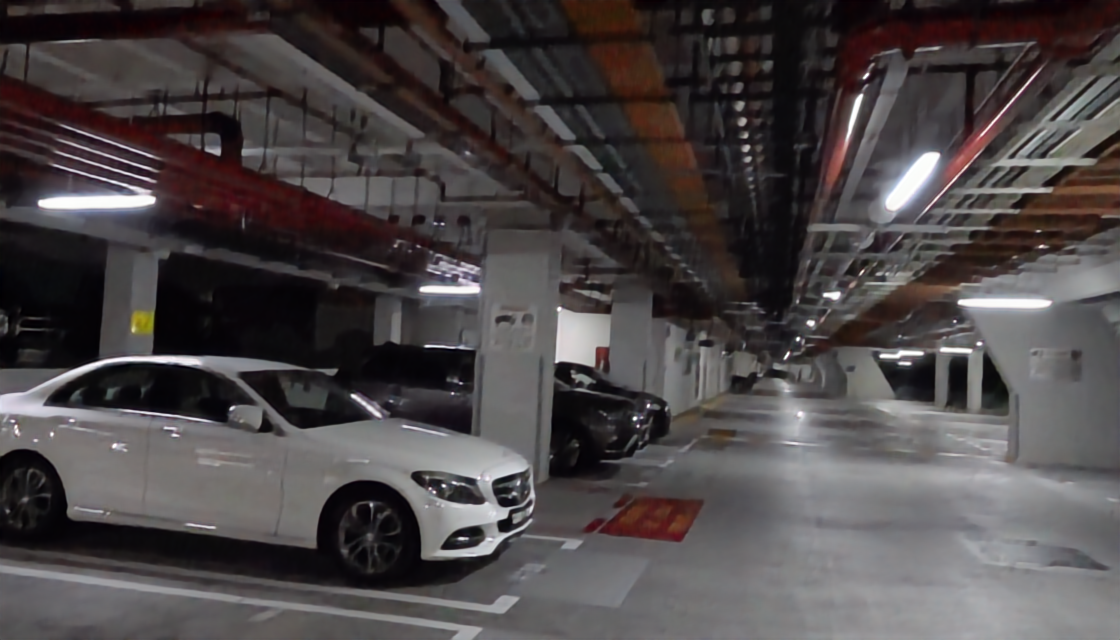} &
   \jfigzoom{85mm 0mm 85mm 0mm}{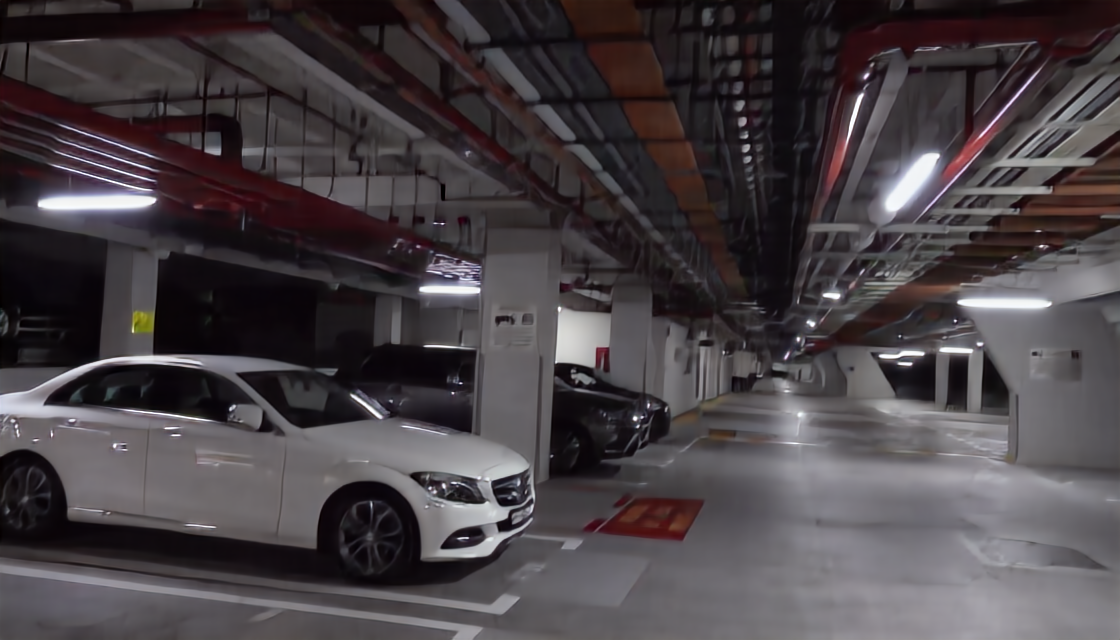} &
   \jfigzoom{85mm 0mm 85mm 0mm}{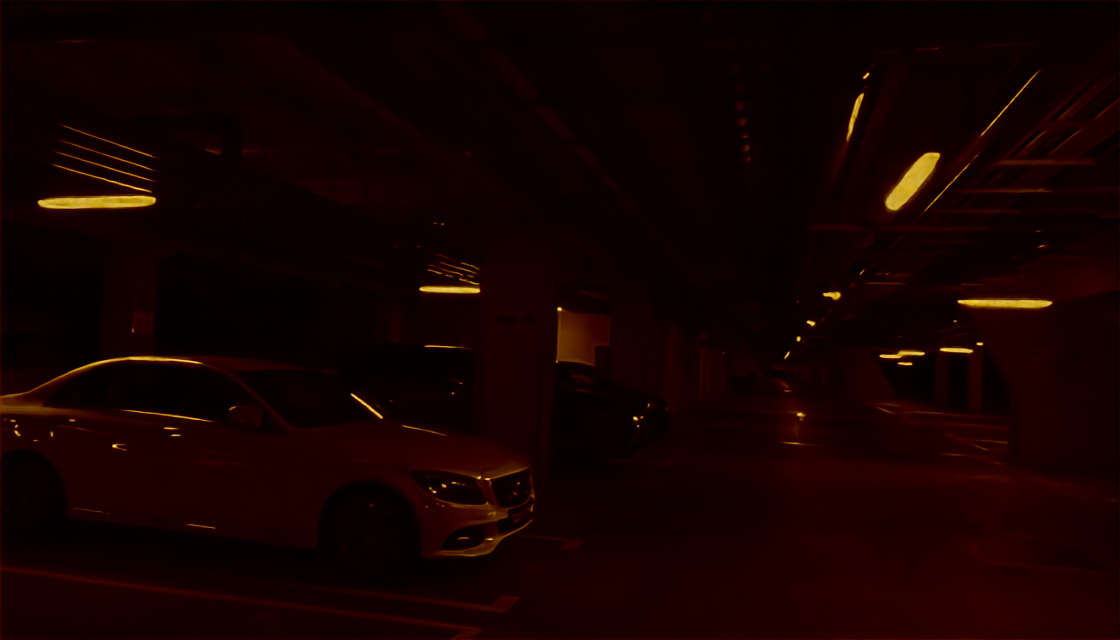} \\
   \jfigzoom{85mm 0mm 85mm 0mm}{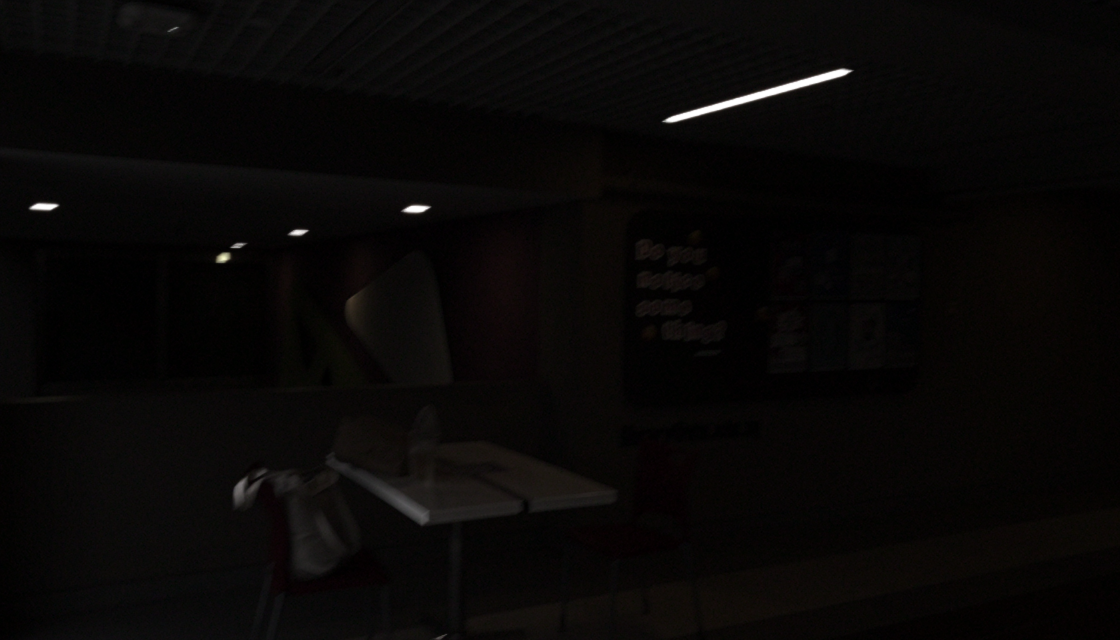} &
   \jfigzoom{85mm 0mm 85mm 0mm}{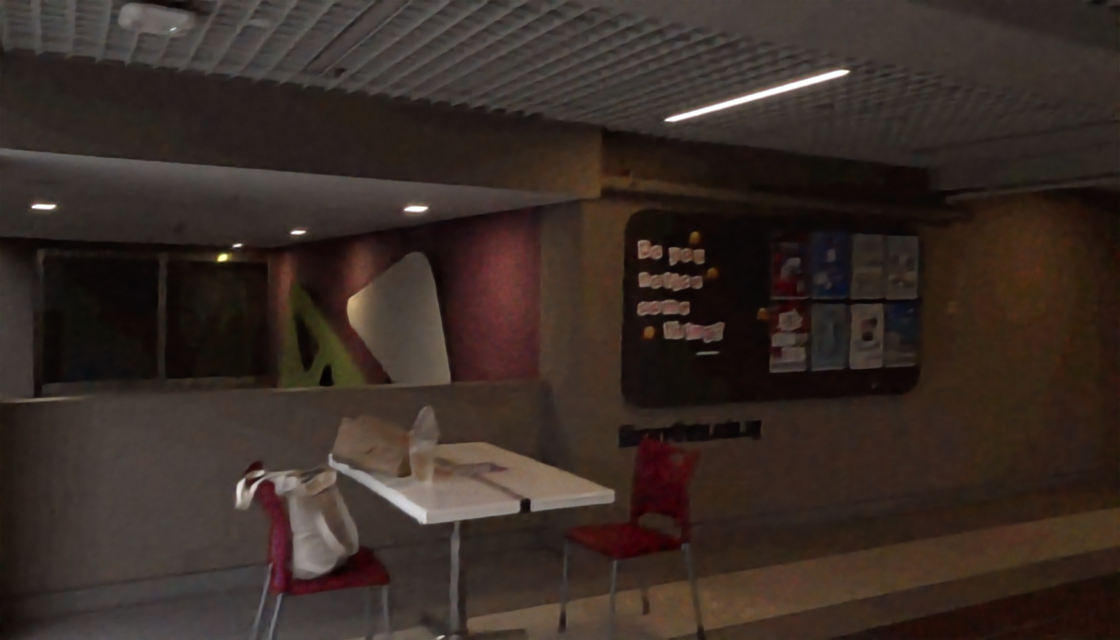} &
   \jfigzoom{85mm 0mm 85mm 0mm}{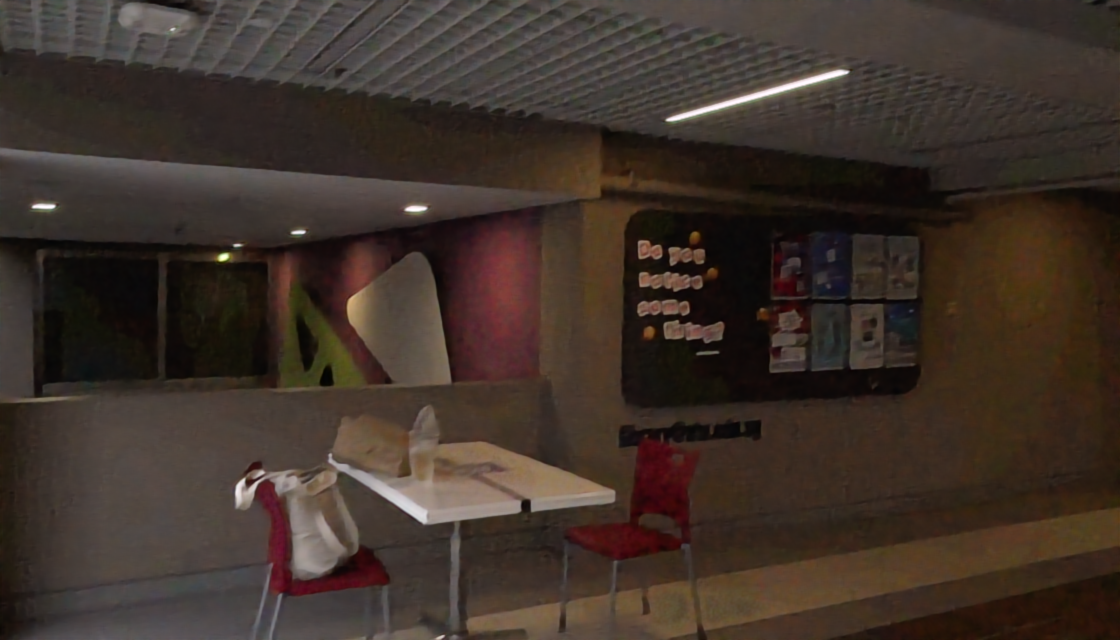} &
   \jfigzoom{85mm 0mm 85mm 0mm}{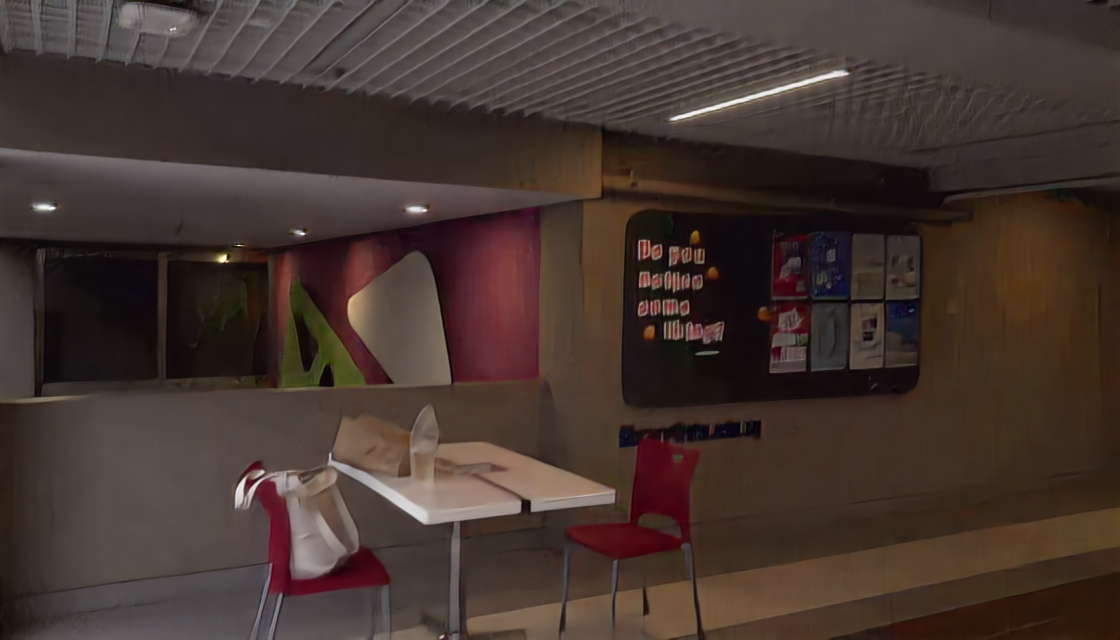} &
   \jfigzoom{85mm 0mm 85mm 0mm}{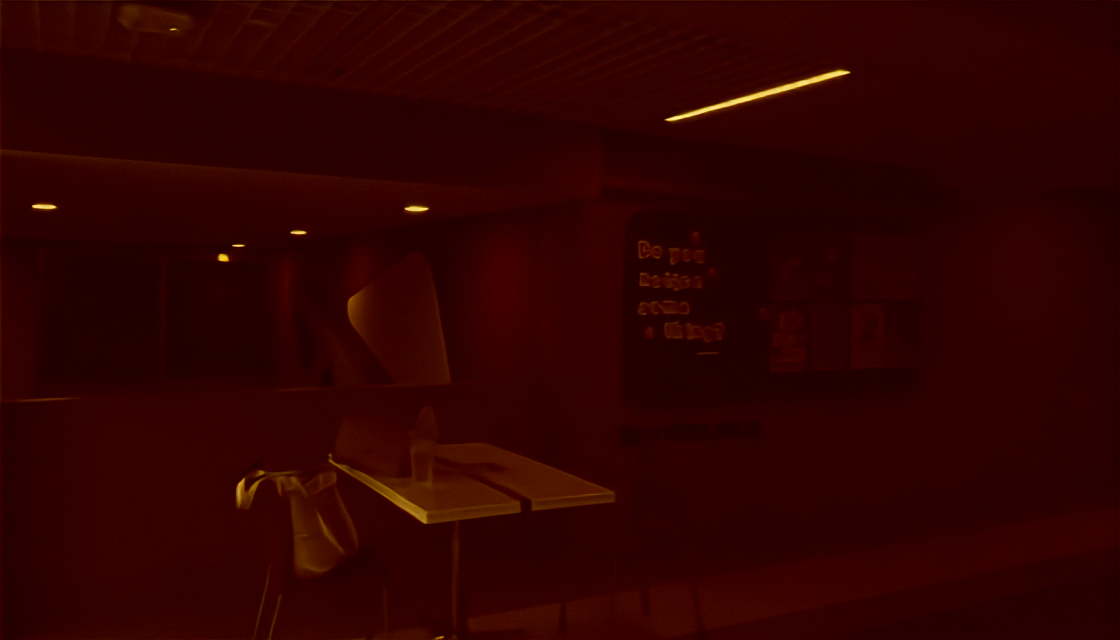} \\
   Input & $\ell_1$ & $\ell_1$+Per & $\ell_1$+Per+Adv & DDPM
   \end{tabular}

   \vspace{1mm}

   \begin{tabular}{ccccc}
    \jfigzoom{85mm 0mm 85mm 0mm}{01_input} &
   \jfigzoom{85mm 0mm 85mm 0mm}{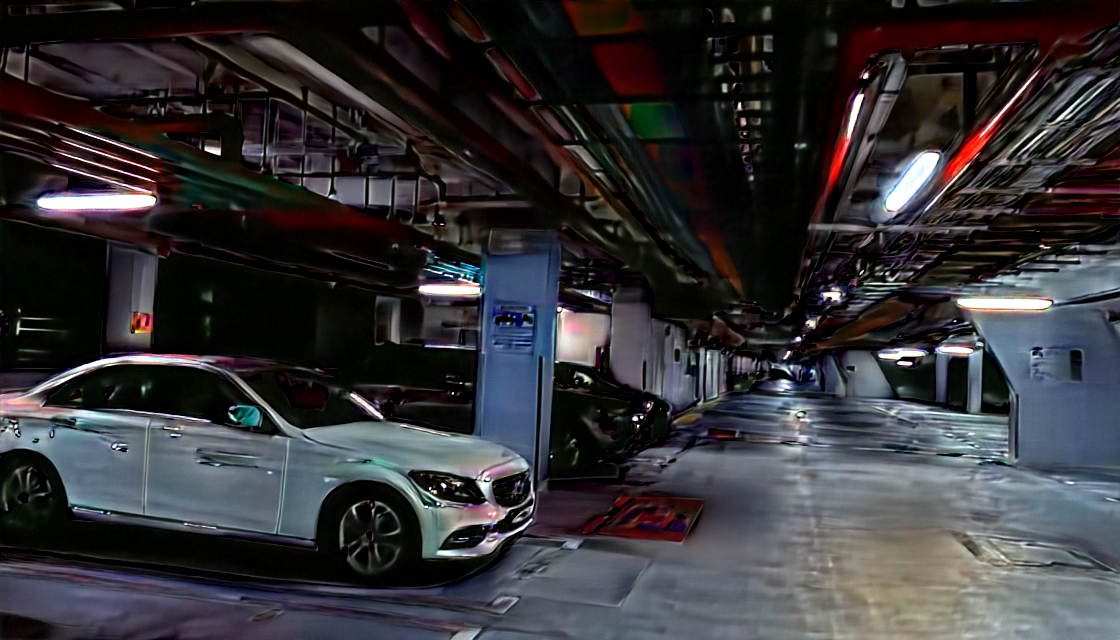} &
   \jfigzoom{85mm 0mm 85mm 0mm}{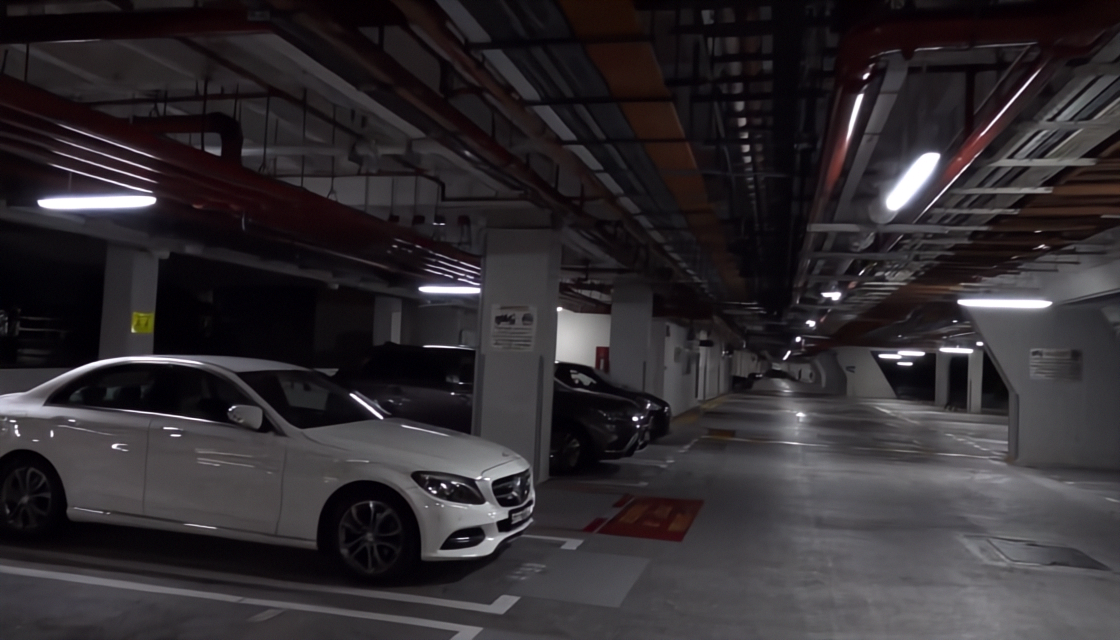} &
   \jfigzoom{85mm 0mm 85mm 0mm}{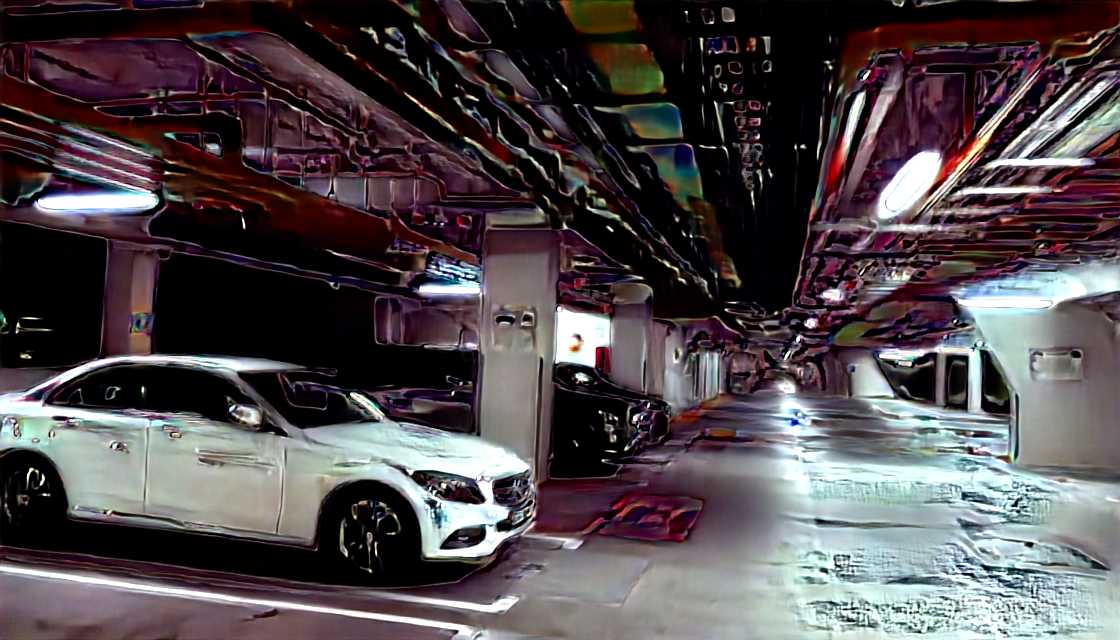} &
   \jfigzoom{85mm 0mm 85mm 0mm}{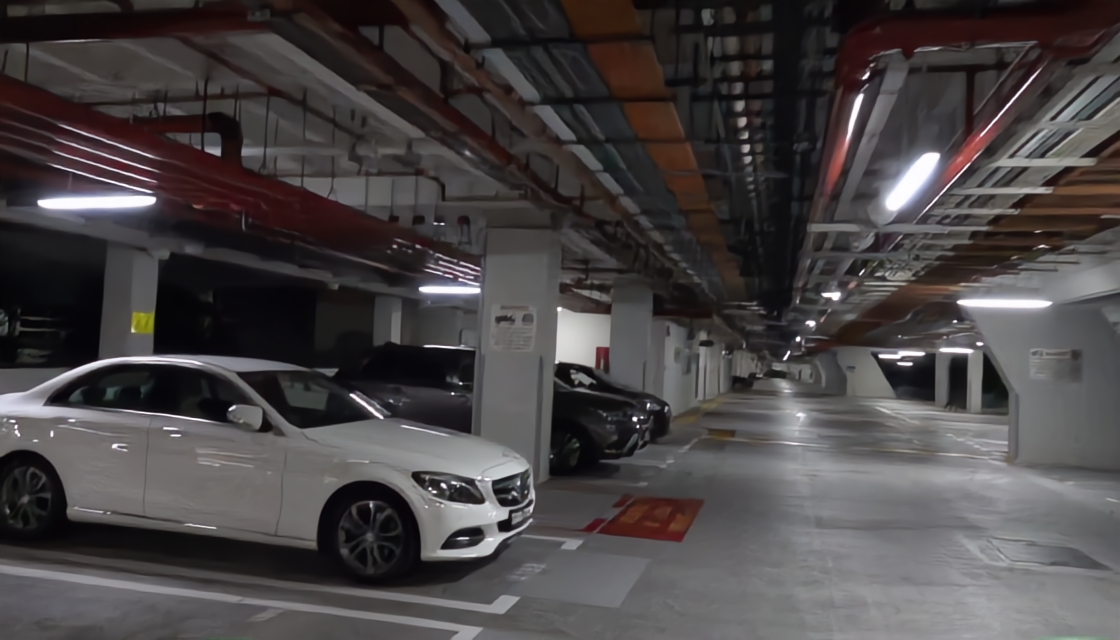} \\
   \jfigzoom{85mm 0mm 85mm 0mm}{02_input} &
   \jfigzoom{85mm 0mm 85mm 0mm}{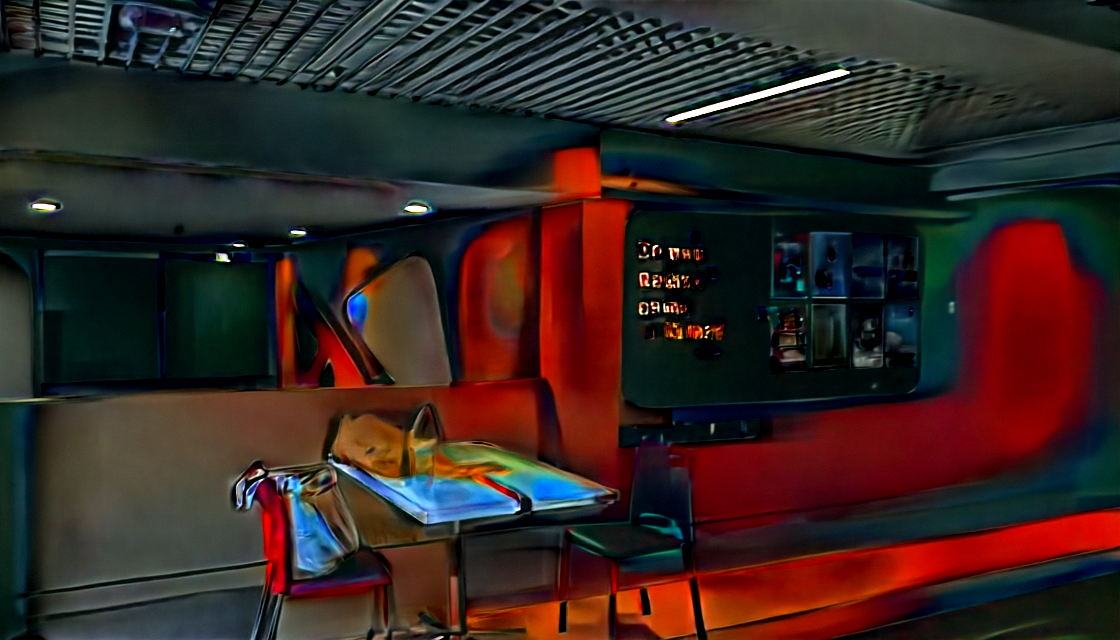} &
   \jfigzoom{85mm 0mm 85mm 0mm}{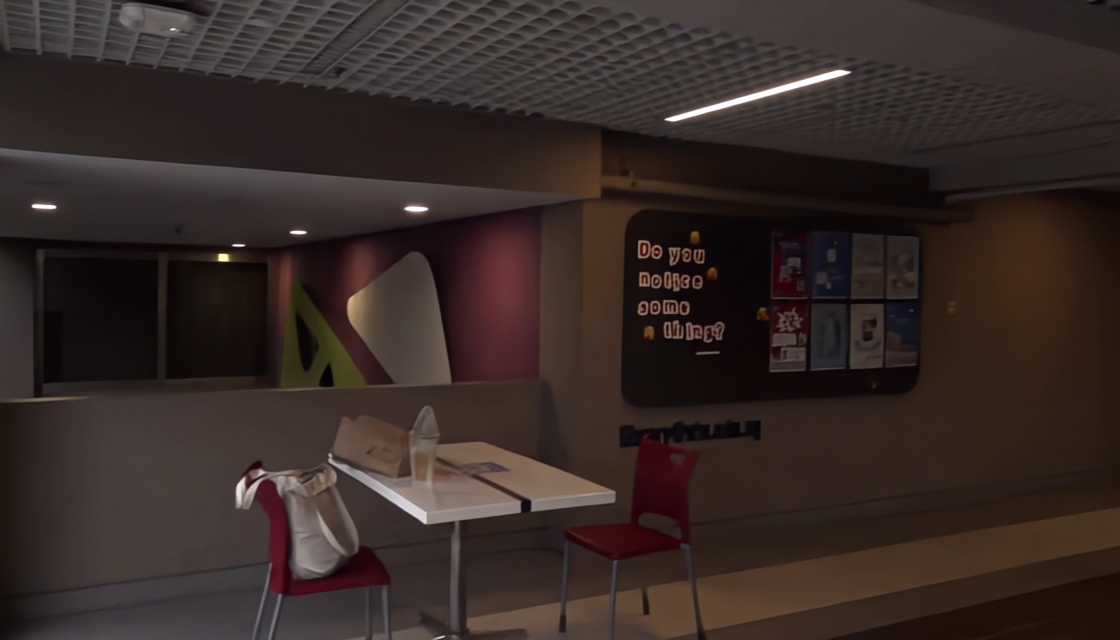} &
   \jfigzoom{85mm 0mm 85mm 0mm}{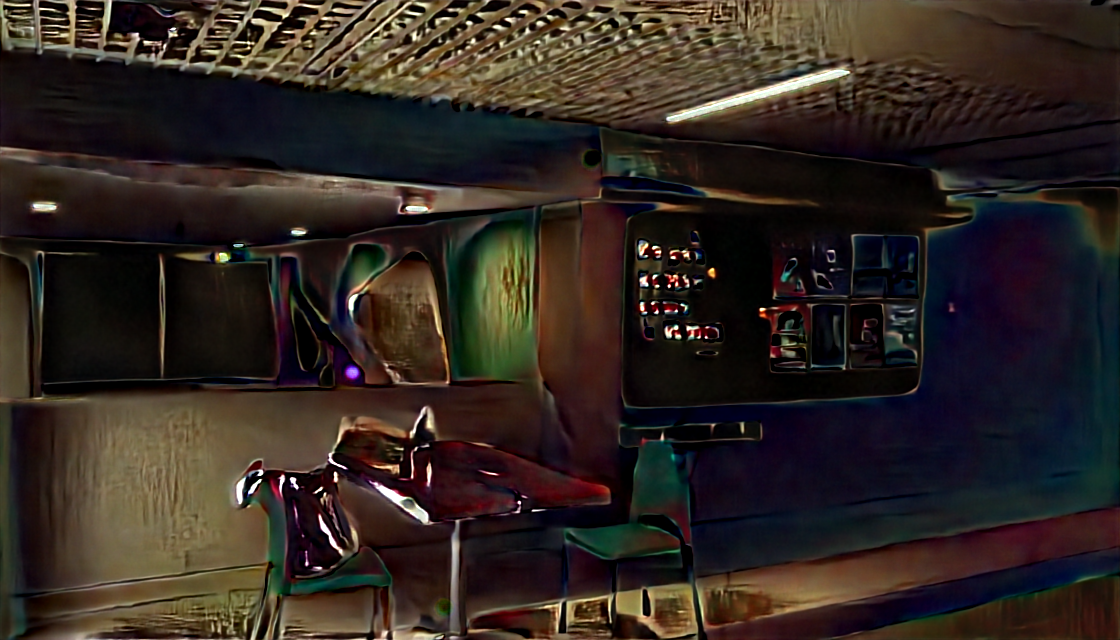} &
   \jfigzoom{85mm 0mm 85mm 0mm}{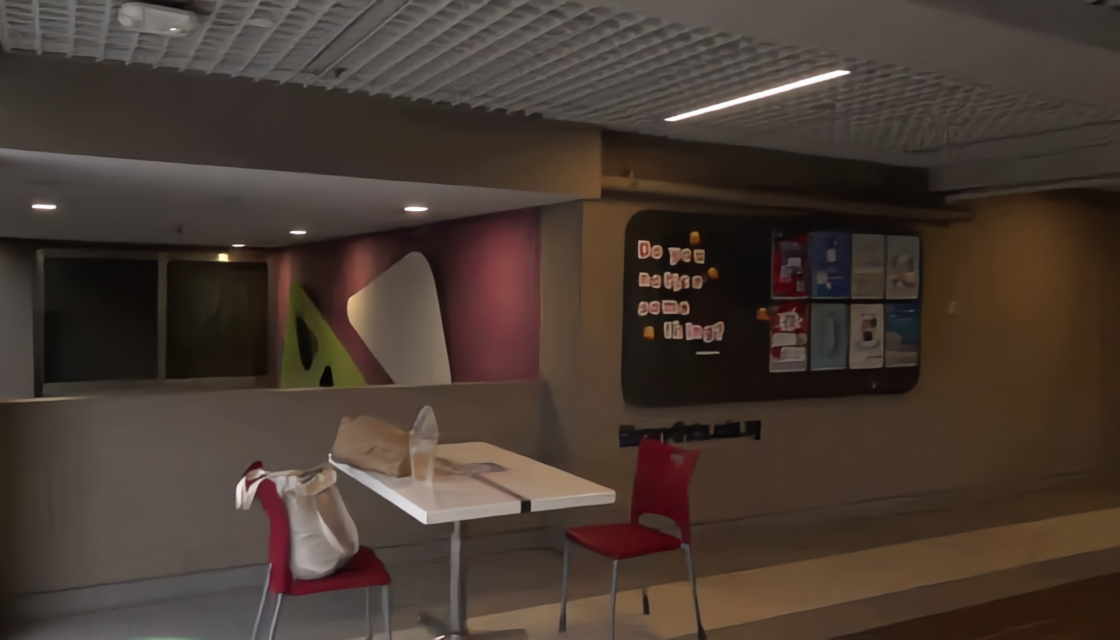} \\
   Input & RFR-$v$ + CFG & RFR-$v$ & RFR + CFG & RFR
   \end{tabular}

   \setlength{\tabcolsep}{6pt}
   \vspace{-3mm}
   \caption{\textbf{Visual comparison with different training objectives and formulation variants.}}
   \label{fig:learning_objective_results}
   \vspace{-3mm}
\end{figure}

\begin{table*}[t]
\centering
\caption{\textbf{Practical engineering overhead when integrating different perceptual-quality-oriented approaches into an existing supervised I2I system.}
This is not a formal computational complexity comparison. Instead, it summarizes the kinds of additional modules, design choices, balancing requirements, and tuning effort that are commonly encountered in practice.}
\label{tab:practical_overhead}
\vspace{-3mm}
\tiny
\setlength{\tabcolsep}{1.5pt}
\renewcommand{\arraystretch}{1.02}

\newcolumntype{Y}{>{\raggedright\arraybackslash}X}
\newcolumntype{L}[1]{>{\raggedright\arraybackslash}p{#1}}
\newcolumntype{C}[1]{>{\centering\arraybackslash}p{#1}}

\newcommand{\cellbullets}[1]{%
\begin{minipage}[t]{\linewidth}
\vspace{0pt}
\raggedright
#1
\end{minipage}
}

\newcommand{\bitem}{\par\noindent\hangindent=1.0em\hangafter=1\makebox[1.0em][l]{\textbullet}}

\begin{tabularx}{\columnwidth}{@{}L{1.15cm} L{1.55cm} Y C{1.10cm} C{1.05cm}@{}}
\toprule
\makecell[l]{\textbf{Method}\\\textbf{family}} &
\makecell[l]{\textbf{Extra}\\\textbf{modules}} &
\makecell[l]{\textbf{Typical integration work in practice}} &
\makecell[c]{\textbf{Inference}\\\textbf{steps}} &
\makecell[c]{\textbf{Tuning}\\\textbf{burden}} \\
\midrule

Perceptual loss
& Pretrained feature network
& \cellbullets{
\bitem Prepare a pretrained feature extractor (e.g., VGG).
\bitem Decide which feature layer(s) to use.
\bitem Tune the balance between perceptual loss and pixel loss.
}
& 1
& Low--Medium \\
\midrule

Adversarial loss
& Discriminator
& \cellbullets{
\bitem Design the discriminator (e.g., depth, PatchGAN vs.\ full-image).
\bitem Optionally choose normalization or stabilization strategies, such as spectral normalization.
\bitem Sometimes adjust the generator to maintain a reasonable generator--discriminator balance.
\bitem Tune the adversarial weight relative to other losses.
\bitem Possibly tune the discriminator learning rate and training schedule.
\bitem Training can be sensitive to these choices and may become unstable or introduce artifacts.
}
& 1
& High \\
\midrule

Diffusion-based I2I
& Diffusion-specific conditioning; sometimes latent encoder--decoder
& \cellbullets{
\bitem Choose the diffusion formulation, including parameterization, conditioning design, and noise schedule.
\bitem Choose the sampler and the denoising step count.
\bitem Latent variants additionally require an encoder--decoder (e.g., a VAE).
\bitem Accelerated variants may require a separate distillation stage and careful tuning.
\bitem Integration into existing task-specialized regression pipelines can be less straightforward.
}
& Many
& High \\
\midrule

\textbf{I2I-RFR (ours)}
& \textbf{No extra trainable module in the default form}
& \cellbullets{
\bitem Expand the backbone input channels (e.g., 3 $\rightarrow$ 6 for RGB I2I).
\bitem Concatenate a noise-corrupted target state with the input and use the $t$-reweighted pixel loss during training.
\bitem Use a small inference step count $N$ for ODE refinement (3 by default).
}
& \textbf{Few}
& \textbf{Low} \\
\bottomrule
\end{tabularx}
\end{table*}

\section{Practical Engineering Overhead}
\label{sec:practical_overhead}
Beyond quantitative performance, another practical consideration is how much engineering overhead is introduced when a perceptual-quality-oriented method is integrated into an existing supervised image-to-image (I2I) system.
Table~\ref{tab:practical_overhead} summarizes this aspect for several commonly used approach families. This comparison is not intended as a formal computational-complexity analysis. Rather, the goal is to highlight the kinds of additional modules, design choices, balancing requirements, and tuning effort that are often encountered in practice when extending a standard supervised I2I pipeline.

Perceptual-loss-based approaches typically preserve the basic single-stage supervised training pipeline, but they often require introducing a pretrained feature extractor and making task-dependent choices about which feature layers to use and how to balance feature-space supervision against the original pixel-space objective.
Adversarial objectives further increase practical overhead by introducing a discriminator together with additional design and optimization choices, such as discriminator architecture, stabilization strategies, adversarial loss weighting, and the generator--discriminator balance. In practice, these choices can make training more sensitive and may lead to instability or task-dependent artifacts.

Diffusion-based I2I approaches can produce strong perceptual quality, but they usually rely on diffusion-specific training and inference pipelines, including parameterization choices, conditioning mechanisms, noise schedules, iterative samplers, and often many denoising steps. Latent variants may additionally require an encoder--decoder such as a VAE, while accelerated variants may require a separate distillation stage together with further tuning. As a result, integrating diffusion-style pipelines into existing task-specialized regression systems can be less straightforward than extending a standard supervised regression baseline.

In contrast, I2I-RFR is designed as a practical plug-in reformulation of standard regression backbones. In the default setup, it requires only expanding the backbone input channels, concatenating a noise-corrupted target state during training, replacing the standard pixel objective with the reformulated $t$-reweighted pixel loss, and choosing a small inference step count for ODE
refinement. 
In our experiments, we use a fixed default recipe with Beta-based $t$-sampling, no explicit time embeddings, and $N=3$ Euler steps unless stated otherwise. This keeps the method close to the original supervised regression interface while introducing relatively little additional tuning burden.

This practical difference is especially important in task-specialized I2I settings, where both architectures and training recipes are already carefully engineered, and introducing new modules or substantially altering the pipeline can incur a non-trivial development cost.

\section{Additional I2I Applications}
\label{sec:I2Iapp}
To further illustrate the breadth of I2I-RFR beyond the main benchmarks, we provide additional qualitative results on image inpainting and colorization (Fig.~\ref{fig:additional_applications_1}), as well as label-to-image translation (Fig.~\ref{fig:additional_applications_2}). These examples are intended as application showcases rather than exhaustive benchmark studies, and they also highlight that the proposed formulation extends beyond same-domain RGB restoration to tasks with different input/output representations.

For all additional tasks, we use Palette~\cite{saharia2022palette} as the backbone and train it under the I2I-RFR formulation.
For image inpainting, we train the model on the Places2 dataset~\cite{ZhouARXIV2016} following the masking setup used in LaMa~\cite{suvorov2022resolution}. 
The loss is computed only on the masked region for efficiency. During inference, we follow RePaint~\cite{lugmayr2022repaint} and replace the non-masked region of the intermediate state with the input image at the corresponding noise level at each refinement step, 
and we finally replace the non-masked region of the output with the input image to ensure pixel-level consistency in the known region. 
Since image inpainting is relatively complex and multimodal, we use an explicit Euler solver with 15 steps for this task. 
We show qualitative results on the Places2 validation set at a resolution of $256\times256$.
For image colorization, we train the model on ImageNet~\cite{DengCVPR2009}. 
The input is the luminance channel $L$ in the Lab color space, 
and the model predicts only the chrominance channels $ab$. 
We show qualitative results on the ImageNet validation set at a resolution of $256\times256$.
Moreover, following pix2pix~\cite{IsolaCVPR2017}, we train the model on the Cityscapes dataset~\cite{cordts2016cityscapes} for labels-to-street-scene translation and on the CMP Facades dataset~\cite{tylevcek2013spatial} for labels-to-facade translation. 
We present qualitative results on the corresponding test sets.

\newcommand{\repfigzoomapp}[3]{%
  \includegraphics[
    width=0.16\linewidth,
    trim=#1,clip
  ]{figs/application/#2/#3}%
}
\newcommand{\repfigapp}[2]{\includegraphics[width=0.24\linewidth]{figs/application/#1/#2}}
\begin{figure*}[t]
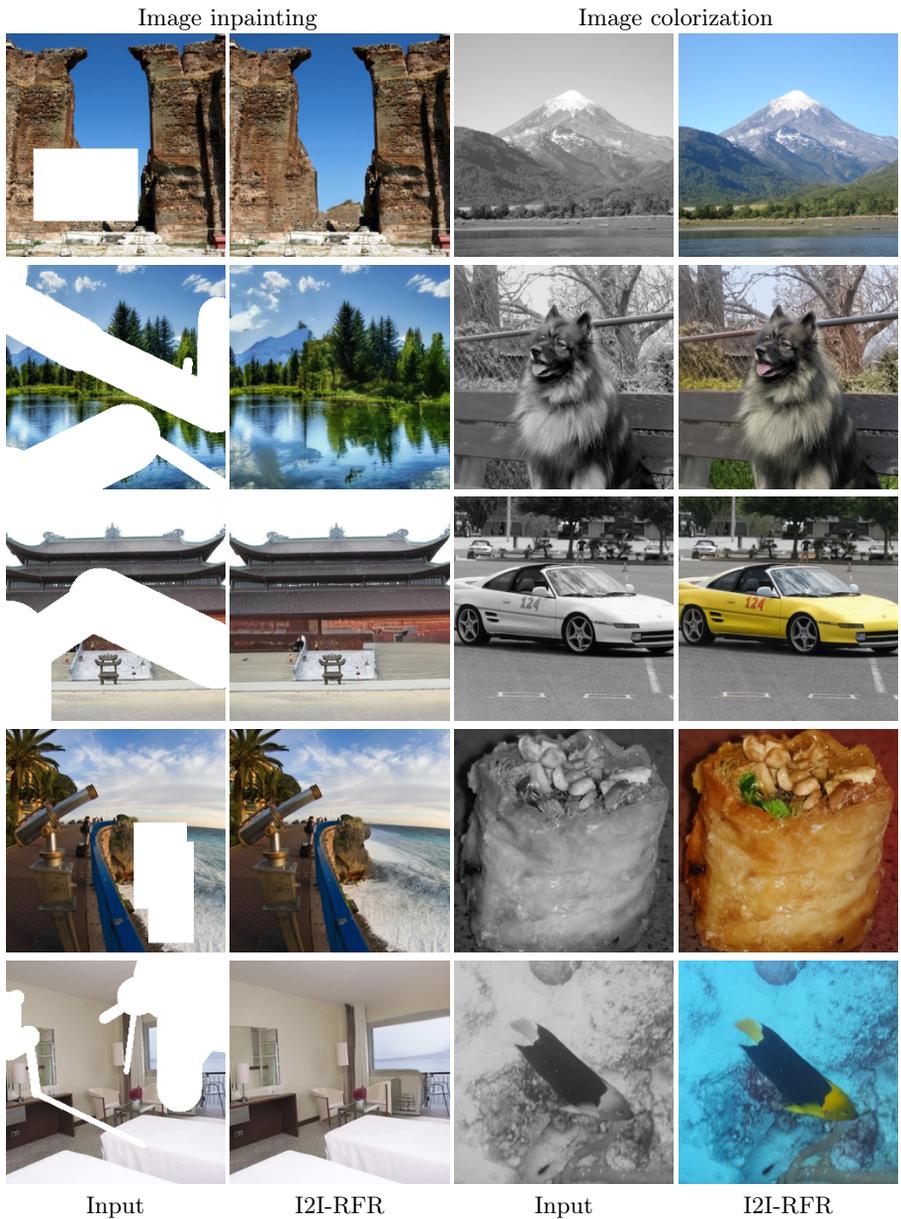

   \centering
   \setlength{\tabcolsep}{1pt}
   \begin{tabular}{cc cc}
   \multicolumn{2}{c}{Image inpainting} &
   \multicolumn{2}{c}{Image colorization} \\
    \repfigapp{inpaint}{img_0037_input.png} &
    \repfigapp{inpaint}{img_0037_pred.png} &
    \repfigapp{color}{img_0002_input.png} &
    \repfigapp{color}{img_0002_pred.png} \\
    \repfigapp{inpaint}{img_0305_input.png} &
    \repfigapp{inpaint}{img_0305_pred.png} &
    \repfigapp{color}{img_0040_input.png} &
    \repfigapp{color}{img_0040_pred.png} \\
    \repfigapp{inpaint}{img_0049_input.png} &
    \repfigapp{inpaint}{img_0049_pred.png} &
    \repfigapp{color}{img_0019_input.png} &
    \repfigapp{color}{img_0019_pred.png} \\
    \repfigapp{inpaint}{img_0052_input.png} &
    \repfigapp{inpaint}{img_0052_pred.png} &
    \repfigapp{color}{img_0095_input.png} &
    \repfigapp{color}{img_0095_pred.png} \\
    \repfigapp{inpaint}{img_0311_input.png} &
    \repfigapp{inpaint}{img_0311_pred.png} &
    \repfigapp{color}{img_0235_input.png} &
    \repfigapp{color}{img_0235_pred.png} \\
   Input &  I2I-RFR & Input &  I2I-RFR \\  
   \end{tabular}
   \setlength{\tabcolsep}{6pt}
    \caption{\textbf{Additional qualitative applications of I2I-RFR: image inpainting and image colorization.} }
   \label{fig:additional_applications_1}
\end{figure*}

\begin{figure*}[t]
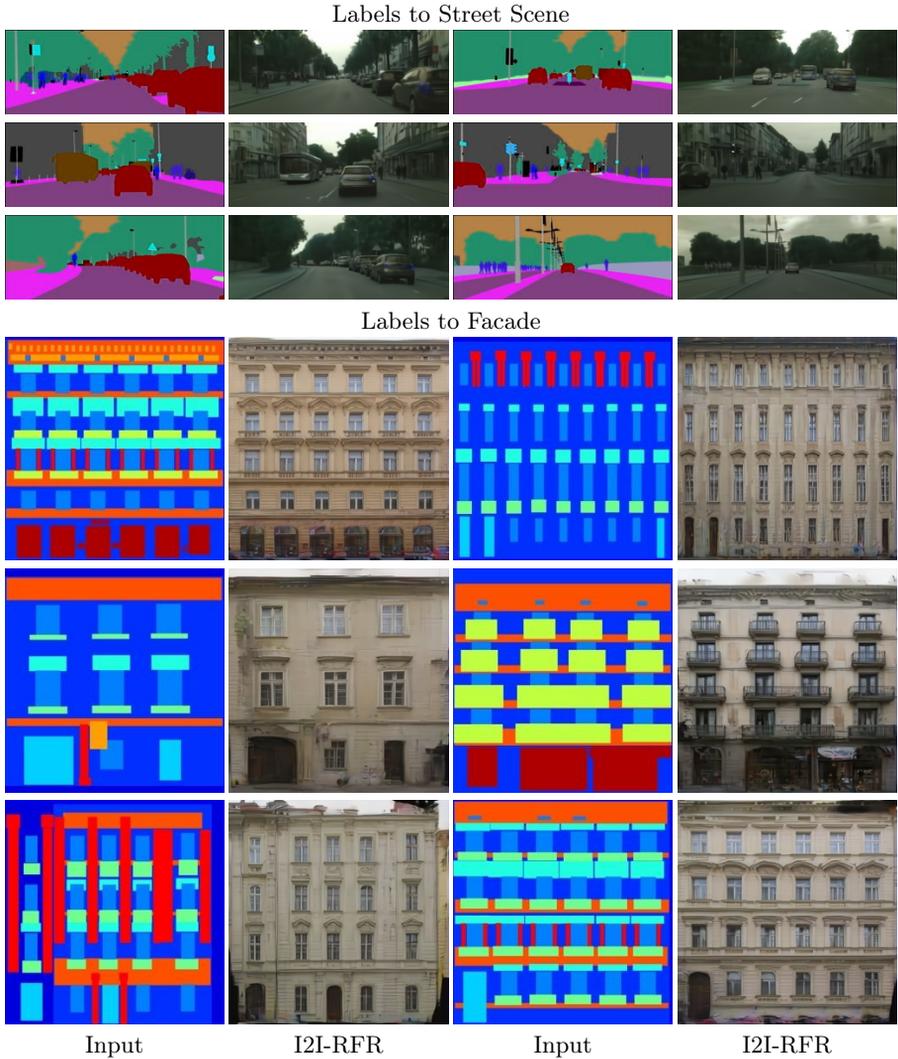

   \centering
   \setlength{\tabcolsep}{1pt}
   \begin{tabular}{cccc}
   \multicolumn{4}{c}{Labels to Street Scene} \\
    \repfigapp{city}{img_0342_input.png} &
    \repfigapp{city}{img_0342_pred.png} &
    \repfigapp{city}{img_0499_input.png} &
    \repfigapp{city}{img_0499_pred.png} \\
    \repfigapp{city}{img_0017_input.png} &
    \repfigapp{city}{img_0017_pred.png} &
    \repfigapp{city}{img_0020_input.png} &
    \repfigapp{city}{img_0020_pred.png} \\
    \repfigapp{city}{img_0025_input.png} &
    \repfigapp{city}{img_0025_pred.png} &
    \repfigapp{city}{img_0044_input.png} &
    \repfigapp{city}{img_0044_pred.png} \\
    \multicolumn{4}{c}{Labels to Facade} \\
    \repfigapp{facade}{img_0001_input.png} &
    \repfigapp{facade}{img_0001_pred.png} &
    \repfigapp{facade}{img_0025_input.png} &
    \repfigapp{facade}{img_0025_pred.png} \\
    \repfigapp{facade}{img_0047_input.png} &
    \repfigapp{facade}{img_0047_pred.png} &
    \repfigapp{facade}{img_0067_input.png} &
    \repfigapp{facade}{img_0067_pred.png} \\
    \repfigapp{facade}{img_0063_input.png} &
    \repfigapp{facade}{img_0063_pred.png} &
    \repfigapp{facade}{img_0102_input.png} &
    \repfigapp{facade}{img_0102_pred.png} \\
   Input &  I2I-RFR & Input &  I2I-RFR \\
   \end{tabular}
   \setlength{\tabcolsep}{6pt}
    \caption{\textbf{Additional qualitative applications of I2I-RFR: label-to-image translation.} We show representative results on street-scene and facade generation to further illustrate the flexibility of the proposed formulation.}
   \label{fig:additional_applications_2}
\end{figure*}

\section{Additional Qualitative Results}
\label{sec:qual}
We provide additional qualitative comparisons for the tasks studied in the main paper, including super-resolution (Fig.~\ref{fig:additional_super_resolution}), deblurring (Fig.~\ref{fig:additional_deblurring}), low-light blur enhancement (Fig.~\ref{fig:additional_low_light}), underwater image enhancement (Fig.~\ref{fig:additional_underwater}), and video restoration (Fig.~\ref{fig:additional_video_restoration}).
These examples are intended to complement the quantitative results and to highlight the perceptual behavior of I2I-RFR under different settings.

\newcommand{\repfigquali}[2]{\includegraphics[width=0.31\linewidth]{figs/results/#1/#2}}
\begin{figure*}[t]
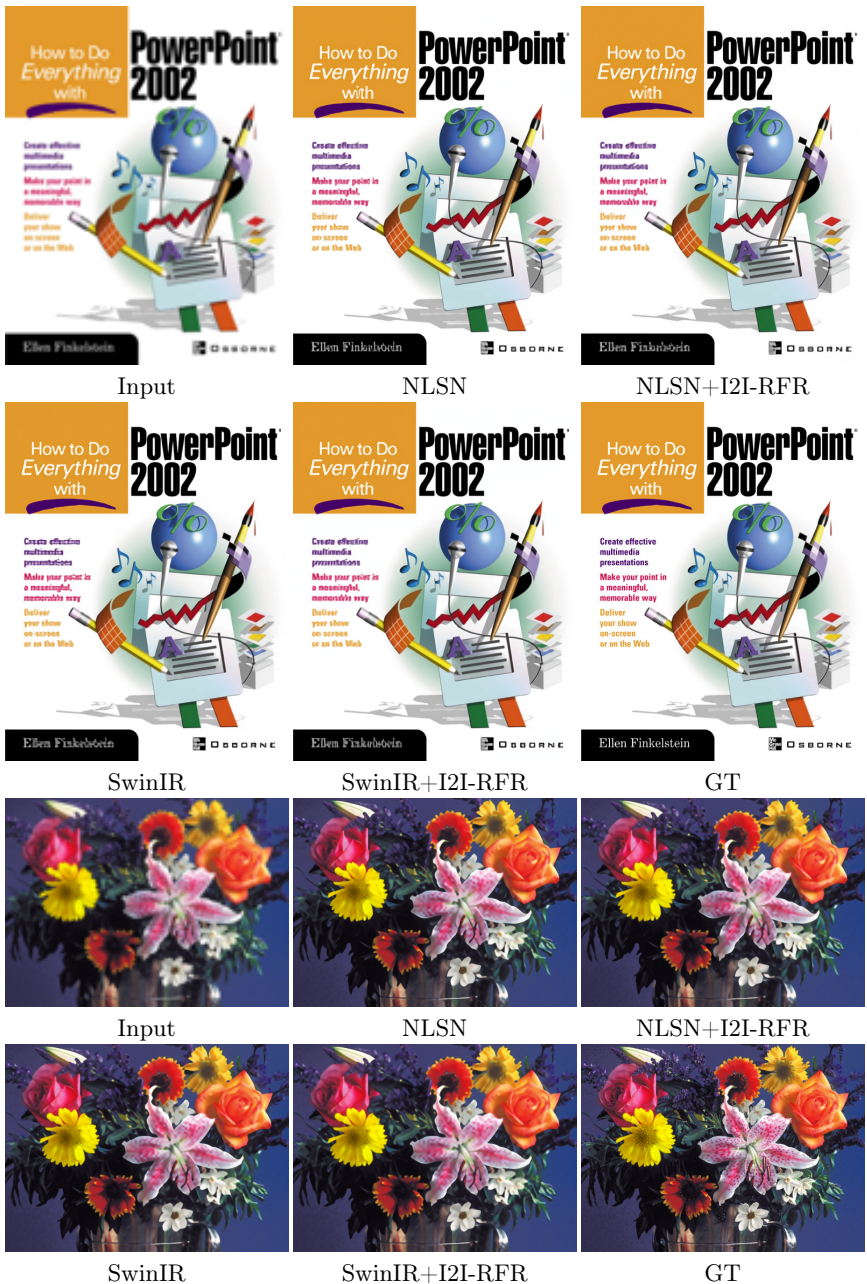

   \centering
   \setlength{\tabcolsep}{1pt}
   \begin{tabular}{ccc}
    \repfigquali{Comp_Supres}{3_input.png} &
    \repfigquali{Comp_Supres}{3_nlsn_i2i.png} &
    \repfigquali{Comp_Supres}{3_nlsn.png} \\
    Input & NLSN & NLSN+I2I-RFR \\
    \repfigquali{Comp_Supres}{3_swin_i2i.png} &
    \repfigquali{Comp_Supres}{3_swin.png} &
    \repfigquali{Comp_Supres}{3_gt.png} \\
    SwinIR & SwinIR+I2I-RFR & GT \\
        \repfigquali{Comp_Supres}{1_input.png} &
    \repfigquali{Comp_Supres}{1_nlsn_i2i.png} &
    \repfigquali{Comp_Supres}{1_nlsn.png} \\
    Input & NLSN & NLSN+I2I-RFR \\
    \repfigquali{Comp_Supres}{1_swin_i2i.png} &
    \repfigquali{Comp_Supres}{1_swin.png} &
    \repfigquali{Comp_Supres}{1_gt.png} \\
    SwinIR & SwinIR+I2I-RFR & GT \\
   \end{tabular}
   \setlength{\tabcolsep}{6pt}
    \caption{\textbf{Additional qualitative results on image super-resolution.} We show representative examples to further illustrate the perceptual behavior of I2I-RFR in this setting.}
   \label{fig:additional_super_resolution}
\end{figure*}

\newcommand{\repfigqualib}[2]{\includegraphics[width=0.30\linewidth]{figs/results/#1/#2}}
\begin{figure*}[t]
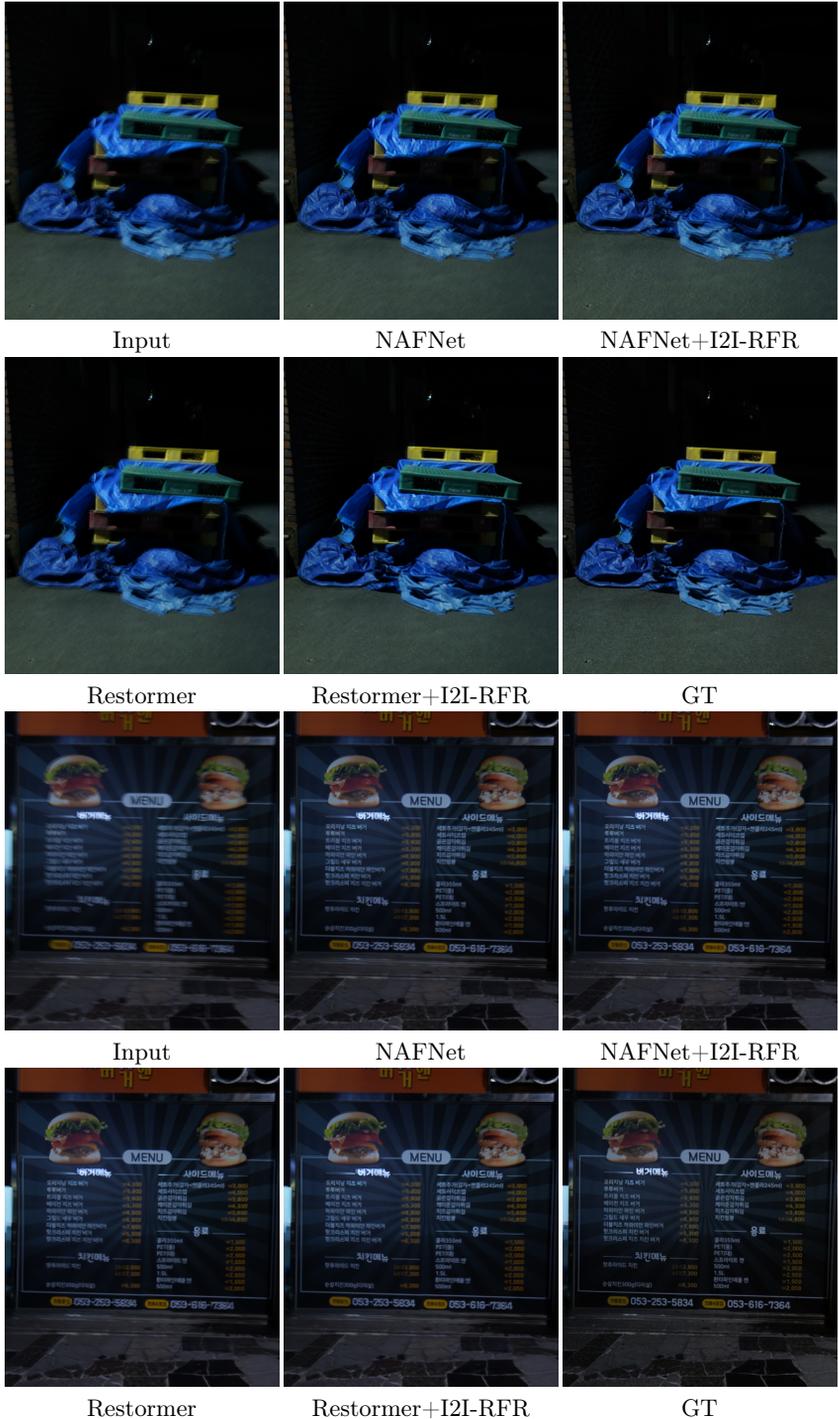

   \centering
   \setlength{\tabcolsep}{1pt}
   \begin{tabular}{ccc}
    \repfigqualib{Comp_Deblur}{1_input.png} &
    \repfigqualib{Comp_Deblur}{1_nafnet_i2i.png} &
    \repfigqualib{Comp_Deblur}{1_nafnet.png} \\
    Input & NAFNet & NAFNet+I2I-RFR \\
    \repfigqualib{Comp_Deblur}{1_restormer_i2i.png} &
    \repfigqualib{Comp_Deblur}{1_restormer.png} &
    \repfigqualib{Comp_Deblur}{1_gt.png} \\
    Restormer & Restormer+I2I-RFR & GT \\   
    \repfigqualib{Comp_Deblur}{0_input.png} &
    \repfigqualib{Comp_Deblur}{0_nafnet_i2i.png} &
    \repfigqualib{Comp_Deblur}{0_nafnet.png} \\
        Input & NAFNet & NAFNet+I2I-RFR \\
    \repfigqualib{Comp_Deblur}{0_restormer_i2i.png} &
    \repfigqualib{Comp_Deblur}{0_restormer.png} &
    \repfigqualib{Comp_Deblur}{0_gt.png} \\
    Restormer & Restormer+I2I-RFR & GT \\
   \end{tabular}
   \setlength{\tabcolsep}{6pt}
   \vspace{-2mm}
    \caption{\textbf{Additional qualitative results on image deblurring.} We show representative examples to further illustrate the perceptual behavior of I2I-RFR in this setting.}
   \label{fig:additional_deblurring}
\end{figure*}

\begin{figure*}[t]
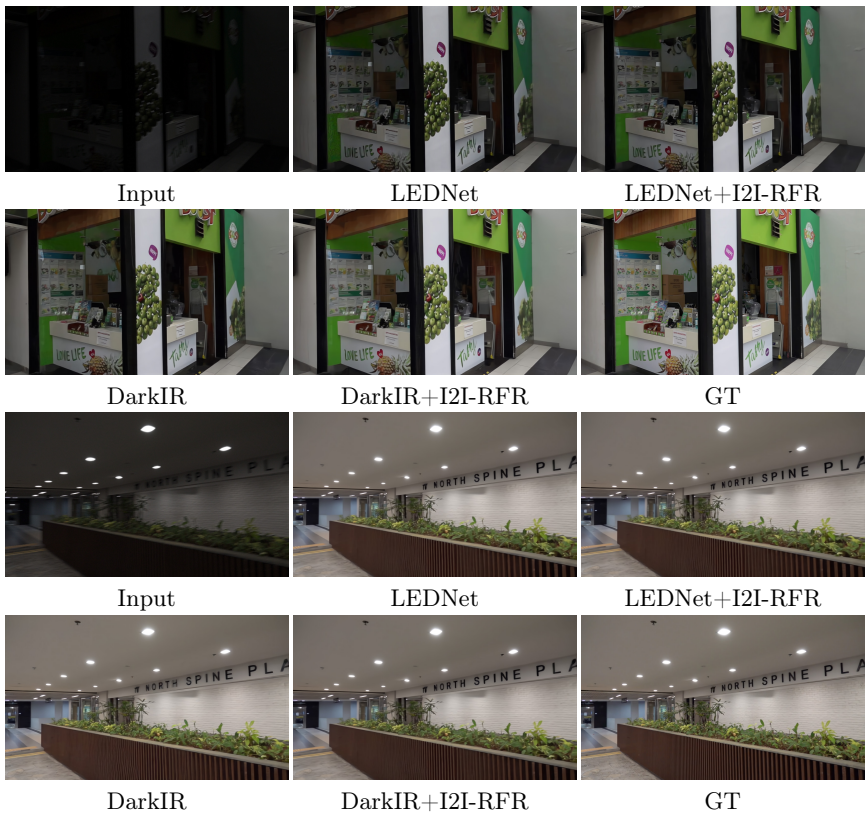

   \centering
   \setlength{\tabcolsep}{1pt}
   \begin{tabular}{ccc}
    \repfigquali{Comp_Dark}{0_input} &
    \repfigquali{Comp_Dark}{0_lednet_i2i} &
    \repfigquali{Comp_Dark}{0_lednet} \\
    Input & LEDNet & LEDNet+I2I-RFR \\
    \repfigquali{Comp_Dark}{0_darkir_i2i} &
    \repfigquali{Comp_Dark}{0_darkir} &
    \repfigquali{Comp_Dark}{0_gt} \\
    DarkIR & DarkIR+I2I-RFR & GT \\
    \repfigquali{Comp_Dark}{1_input} &
    \repfigquali{Comp_Dark}{1_lednet_i2i} &
    \repfigquali{Comp_Dark}{1_lednet} \\
    Input & LEDNet & LEDNet+I2I-RFR \\
    \repfigquali{Comp_Dark}{1_darkir_i2i} &
    \repfigquali{Comp_Dark}{1_darkir} &
    \repfigquali{Comp_Dark}{1_gt} \\
    DarkIR & DarkIR+I2I-RFR & GT \\
   \end{tabular}
   \setlength{\tabcolsep}{6pt}
    \caption{\textbf{Additional qualitative results on low-light blur enhancement.} We show representative examples to further illustrate the perceptual behavior of I2I-RFR in this setting.}
   \label{fig:additional_low_light}
\end{figure*}

\begin{figure*}[t]
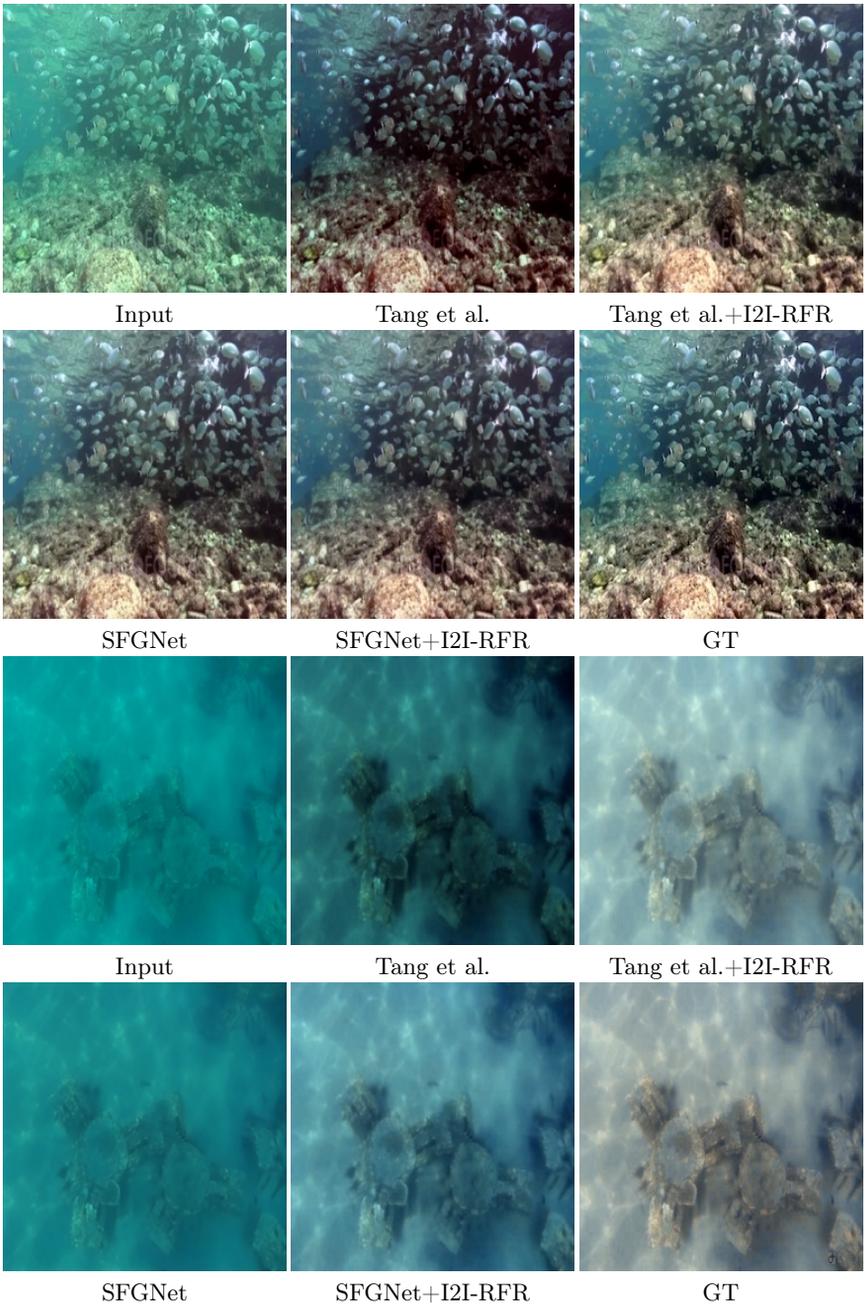

   \centering
   \setlength{\tabcolsep}{1pt}
   \begin{tabular}{ccc}
    \repfigquali{Comp_water}{0_input.png} &
    \repfigquali{Comp_water}{0_tang_i2i.png} &
    \repfigquali{Comp_water}{0_tang.png} \\
    Input & Tang et al. & Tang et al.+I2I-RFR \\
    \repfigquali{Comp_water}{0_sfg_i2i.png} &
    \repfigquali{Comp_water}{0_sfg.png} &
    \repfigquali{Comp_water}{0_gt.png} \\
    SFGNet & SFGNet+I2I-RFR & GT \\
    \repfigquali{Comp_water}{1_input.png} &
    \repfigquali{Comp_water}{1_tang_i2i.png} &
    \repfigquali{Comp_water}{1_tang.png} \\
    Input & Tang et al. & Tang et al.+I2I-RFR \\
    \repfigquali{Comp_water}{1_sfg_i2i.png} &
    \repfigquali{Comp_water}{1_sfg.png} &
    \repfigquali{Comp_water}{1_gt.png} \\
    SFGNet & SFGNet+I2I-RFR & GT \\
   \end{tabular}
   \setlength{\tabcolsep}{6pt}
    \caption{\textbf{Additional qualitative results on underwater image enhancement.} We show representative examples to further illustrate the perceptual behavior of I2I-RFR in this setting.}
   \label{fig:additional_underwater}
\end{figure*}

\begin{figure*}[t]
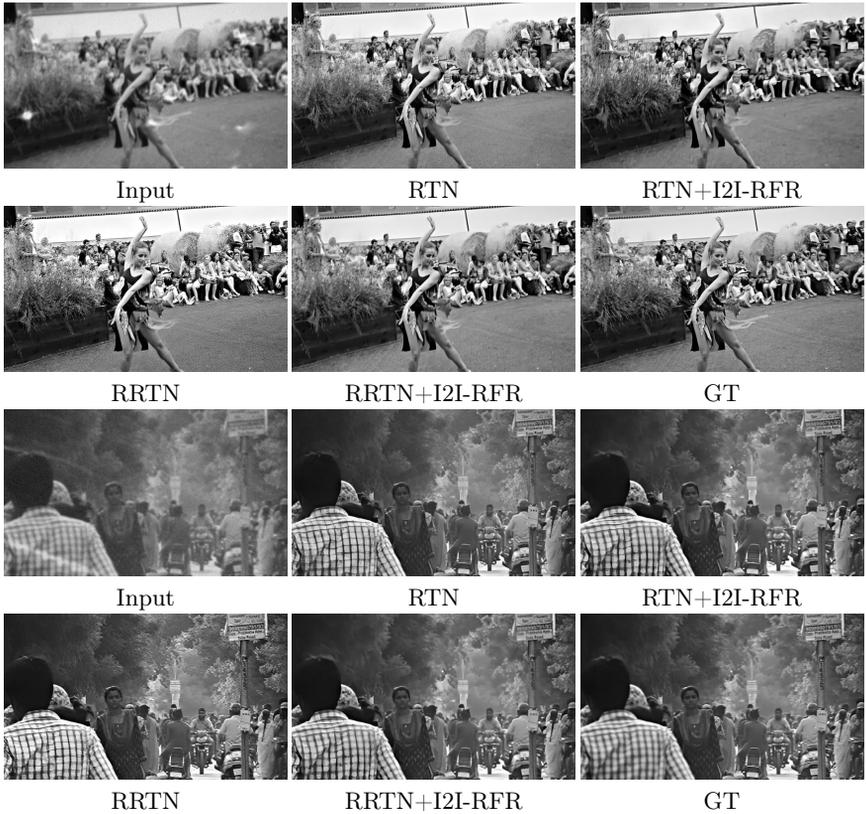

   \centering
   \setlength{\tabcolsep}{1pt}
   \begin{tabular}{ccc}
    \repfigquali{Comp_Remaster}{0_input} &
    \repfigquali{Comp_Remaster}{0_rtn_i2i} &
    \repfigquali{Comp_Remaster}{0_rtn} \\
    Input & RTN & RTN+I2I-RFR \\
    \repfigquali{Comp_Remaster}{0_rrtn_i2i} &
    \repfigquali{Comp_Remaster}{0_rrtn} &
    \repfigquali{Comp_Remaster}{0_gt} \\
    RRTN & RRTN+I2I-RFR & GT \\
    \repfigquali{Comp_Remaster}{1_input} &
    \repfigquali{Comp_Remaster}{1_rtn_i2i} &
    \repfigquali{Comp_Remaster}{1_rtn} \\
    Input & RTN & RTN+I2I-RFR \\
    \repfigquali{Comp_Remaster}{1_rrtn_i2i} &
    \repfigquali{Comp_Remaster}{1_rrtn} &
    \repfigquali{Comp_Remaster}{1_gt} \\
    RRTN & RRTN+I2I-RFR & GT \\
   \end{tabular}
   \setlength{\tabcolsep}{6pt}
    \caption{\textbf{Additional qualitative results on video restoration.} We show representative examples to further illustrate the perceptual behavior of I2I-RFR in this setting.}
   \label{fig:additional_video_restoration}
\end{figure*}


\end{document}